\documentclass[twoside,11pt]{article}
\usepackage{jmlr2e}

\usepackage[dvipsnames]{xcolor}
\usepackage{tikz}
\usepackage{style}
\usepackage{graphicx}
\usepackage[inline, shortlabels]{enumitem}
\usepackage{url}

\usepackage{todonotes}
\usepackage[most]{tcolorbox}

\usepackage{mopcommands}

\newtcolorbox{examplebox}
{
  breakable, enhanced,
  colframe = black!25!white,
  colback  = black!5!white
}
\newenvironment{Example}{\begin{examplebox}\begin{Exa}}%
                     {\end{Exa}\end{examplebox}}

\title{A Probabilistic Framework for Learnable Optimization Algorithms}

\author{%
\name Peter Ochs \email ochs@math.uni-saarland.de \\
       \addr Department of Mathematics and Computer Science\\
        Saarland University\\
        Saarbr\"ucken, Germany
\AND
\name Michael Sucker \email michael.sucker@math.uni-tuebingen.de \\
       \addr Department of Mathematics \\
        University of T\"ubingen\\
        T\"ubingen, Germany
}

\editor{}

\usepackage{lastpage}
\ShortHeadings{Probabilistic LOA}{Ochs and Sucker}
\firstpageno{1}

\begin{document}

\maketitle

\begin{abstract}%
We propose a statistical-learning framework for optimization algorithms. The framework is based on probability distributions over optimization trajectories induced by a distribution of optimization problems and a learnable optimization algorithm. Within this setting, optimization performance is represented through measurable performance functionals, including stopping times, contraction factors, and trajectory-level properties. The resulting framework provides a statistical perspective on optimization algorithms, allows their performance to be studied at the population level, and naturally accommodates the learning of optimization algorithms from data. Moreover, it enables the application of statistical-learning techniques to optimization, including PAC-Bayesian generalization guarantees for bounded optimization-specific performance measures. Experiments on convex, non-smooth, non-convex, and stochastic optimization problems illustrate the framework and demonstrate the statistical nature of optimization performance. 
\end{abstract}


%\tableofcontents

\section{Introduction}

Optimization algorithms are among the most fundamental tools in scientific computing, machine learning, and data analysis. Their performance is traditionally studied through convergence guarantees, complexity estimates, and asymptotic rates. Such analyses typically focus on a fixed optimization algorithm applied to a fixed optimization problem or on worst-case guarantees over a prescribed class of problems. This viewpoint has been extraordinarily successful and forms the foundation of modern optimization theory.

Many contemporary applications, however, involve distributions of related optimization problems rather than isolated problem instances. Examples range from inverse problems arising from a common measurement process to machine learning tasks generated by similar data distributions. This observation has motivated a growing interest in data-driven approaches for algorithm design, where optimization algorithms themselves are trained from representative optimization problems. Such approaches are commonly studied under the umbrella of \emph{learning to optimize} (L2O) and have demonstrated promising empirical performance across a broad range of applications.

Despite this progress, a fundamental question remains largely open:
\begin{center}
    \emph{What does it actually mean for optimization algorithms to be learnable?}
\end{center}

\noindent At first sight, one might argue that an optimization algorithm is learnable whenever its parameters can be adjusted from data. However, this viewpoint alone does not explain how optimization performance should be measured, compared, or generalized across a distribution of optimization problems. While optimization theory studies convergence properties of algorithms and statistical learning theory studies generalization from data, a framework that combines optimization-specific notions of performance with statistical reasoning is still largely missing.

The starting point of this work is the observation that a distribution of optimization problems, together with a learnable optimization algorithm, naturally induces a distribution of optimization trajectories. From this perspective, optimization performance is no longer associated with a single deterministic trajectory, but becomes a random quantity induced by the underlying problem distribution. Consequently, we argue that
\begin{center}
    \emph{optimization performance is inherently a statistical quantity.}
\end{center}

Rather than studying a single optimization problem or a worst-case instance, one may therefore investigate statistical properties of the induced trajectory distribution. Examples include stopping times, finite-horizon objective errors, distances to solution sets, contraction factors, and probabilities of satisfying a prescribed solution criterion. Such quantities capture complementary aspects of algorithmic behavior and naturally give rise to notions of average, typical, and high-probability performance.

Motivated by this viewpoint, we develop a statistical-learning framework for optimization algorithms. The central object of the framework is a probability distribution on optimization trajectories. Optimization-specific notions of performance are represented through measurable performance functionals on trajectory space, allowing the performance of an optimization algorithm to be studied through its induced probability distribution. This perspective separates the definition of optimization performance from the subsequent statistical analysis and provides a common language for a broad range of optimization-specific performance criteria.

Building on this probabilistic foundation, we formulate the learning of optimization algorithms as a statistical learning problem. This yields corresponding notions of empirical and population performance and allows optimization algorithms to be analyzed through the lens of generalization. As one application of the proposed framework, we derive PAC-Bayesian guarantees for bounded optimization-specific performance measures. Importantly, the resulting guarantees are not tied to a particular optimization problem, algorithmic architecture, or performance criterion, but apply to a broad class of measurable performance functionals.

To illustrate the proposed framework, we study several classes of optimization problems, including strongly convex quadratic optimization, image restoration, sparse recovery, deterministic neural network training, and stochastic empirical risk minimization. These experiments demonstrate that optimization performance often exhibits substantial variability across problem instances and that quantities such as means, medians, quantiles, stopping times, and contraction factors reveal different aspects of algorithmic behavior. Moreover, they illustrate how the proposed framework can be used to study the performance of optimization algorithms beyond classical worst-case analysis.

The main contributions of this work are summarized as follows:
\begin{enumerate}
    \item We introduce a probabilistic framework for optimization algorithms based on distributions of optimization trajectories.

    \item We formalize optimization-specific notions of performance through measurable performance functionals, including stopping times, contraction factors, finite-horizon quantities, and trajectory-level properties.

    \item We formulate the learning of optimization algorithms as a statistical learning problem and introduce corresponding notions of empirical and population performance.

    \item We derive PAC-Bayesian generalization guarantees for bounded optimization-specific performance measures.

    \item We provide empirical illustrations on convex, non-smooth, non-convex, and stochastic optimization problems, highlighting the statistical nature of optimization performance.
\end{enumerate}

\section{Related Work}

The literature on learning-to-optimize (L2O) has grown rapidly during the last decade and now encompasses a broad range of approaches, including algorithm unrolling, learned proximal mappings, neural optimizers, meta-learning strategies, and automated algorithm design. Comprehensive overviews of the area are provided by \citet{Chen_Chen_Chen_Heaton_Liu_Wang_Yin_2022} and \citet{Tang_Yao_2024}. While much of this literature focuses on the construction and empirical evaluation of learned optimization algorithms, comparatively less attention has been devoted to the foundational question of how optimization algorithms themselves should be treated as learnable objects. Since this question lies at the heart of the present work, we organize the discussion around learning-to-optimize, learnable optimization algorithms, generalization, and PAC-Bayesian learning.

\subsection{Learning-to-Optimize}

Learning-to-optimize seeks to replace or augment manually designed optimization algorithms through learning-based procedures. Instead of constructing algorithmic components exclusively through mathematical analysis, L2O approaches exploit data from previously solved optimization problems in order to improve future optimization performance.

The broader L2O landscape intersects with several neighboring research communities. Meta-learning, or ``learning-to-learn'', studies how experience from previously observed tasks can be exploited to improve future learning performance \citep{Vilalta_Drissi_2002, Hospedales_Antoniou_Micaelli_Storkey_2021}. AutoML aims at automating the entire machine learning pipeline, including model selection, hyperparameter optimization, and algorithm configuration \citep{Yao_Wang_Chen_Dai_Li_Tu_Yang_Yu_2018, Hutter_Kotthoff_Vanschoren_2019, He_Zhao_Chu_2021}. In contrast, the primary object of interest in learning-to-optimize is the optimization algorithm itself.

Existing L2O methods range from optimizer parameter tuning to fully learned update rules. Representative examples include algorithm unrolling \citep{Gregor_LeCun_2010}, learned iterative shrinkage schemes \citep{Chen_Liu_Wang_Yin_2018, Xin_Wang_Gao_Wipf_Wang_2016}, neural optimizers \citep{Wichrowska_Maheswaranathan_Hoffman_Colmenarejo_Denil_DeFreitas_SohlDickstein_2017, Metz_Maheswaranathan_Nixon_Freeman_SohlDickstein_2019, Metz_Freeman_Harrison_Maheswaranathan_SohlDickstein_2022}, and more recent attempts to incorporate optimization-theoretic principles into the learning process \citep{Liu_Chen_Wang_Yin_Cai_2023,Castera_Ochs_2025}.

Most existing approaches are evaluated through their performance on a prescribed family of optimization problems. This naturally raises questions concerning transferability, robustness, and out-of-distribution performance. More fundamentally, however, it raises the question: \emph{what does it actually mean for an optimization algorithm to be learnable?} While optimization algorithms are routinely trained from data, a general statistical framework describing their performance, learning, and generalization behavior is still largely missing. 

\subsection{Learnable Optimization Algorithms}

A recurring theme in the L2O literature is the tension between algorithmic flexibility and mathematical guarantees. Highly expressive neural optimization architectures often achieve impressive empirical performance, yet may be difficult to analyze theoretically. Conversely, methods with rigorous guarantees frequently rely on substantial structural assumptions.

One prominent strategy is to constrain learned components such that they remain compatible with established optimization theory. For example, \citet{Moeller_Moellenhoff_Cremers_2019} restrict learned updates to descent directions and establish convergence guarantees. Related approaches interpret learned operators as proximal mappings or impose conditions that recover classical optimization structures \citep{Sreehari_Venkatakrishnan_Wohlberg_Buzzard_Drummy_Simmons_Bouman_2016, Chan_Wang_Elgendy_2016, Teodoro_BioucasDias_Figueiredo_2017, Tirer_Giryes_2018, Buzzard_Chan_Sreehari_Bouman_2018, Ryu_Liu_Wang_Chen_Wang_Yin_2019, Sun_Wohlberg_Kamilov_2019, Terris_Repetti_Pesquet_Wiaux_2021, CEM2021}.

More recently, several authors have advocated embedding optimization principles directly into the learning process. \citet{Liu_Chen_Wang_Yin_Cai_2023} argue that optimization-theoretic structure should be enforced by design, while \citet{Castera_Ochs_2025} identify geometric and algorithmic properties shared by successful handcrafted algorithms and use them as design principles for learnable optimization algorithms.  Along similar lines, \citet{MMF25} provide a complete characterization of linearly convergent algorithms for classes of composite optimization problems and show how trainable algorithmic components can be incorporated while preserving worst-case linear convergence guarantees.  \citet{Xie_Yin_Wen_2024} investigate a continuous-time approach to control the learned optimization dynamics and derives convergence properties through the stability of a discretized dynamical system. The resulting guarantees are asymptotic and worst-case in nature.

The present work is complementary to these developments. Rather than proposing a new optimizer architecture or a new class of convergence guarantees, we investigate the statistical learning problem underlying learnable optimization algorithms. Our objective is to provide a framework in which optimization algorithms, trajectory distributions, performance measures, and generalization guarantees can be studied independently of a particular optimizer design.

\subsection{Generalization of Optimization Algorithms}

Generalization is a central concept in statistical learning theory, yet its role in learning-to-optimize remains comparatively underdeveloped. For unrolled optimization algorithms, the finite number of iterations often shifts the focus from asymptotic convergence towards the performance of a learned architecture on previously unseen optimization problems.

Only a limited number of works have investigated generalization questions directly. Existing approaches include analyses based on Rademacher complexity \citep{Chen_Zhang_Reisinger_Song_2020} and algorithmic stability \citep{KEKP2020}, which is closely connected to generalization \citep{Bousquet_Elisseeff_2000, Bousquet_Elisseeff_2002, ShalevShwartz_Shamir_Srebro_Sridharan_2010}.

Our perspective differs in two important ways. First, we formulate generalization directly in terms of optimization performance rather than prediction accuracy. Second, optimization performance is not tied to a single scalar quantity. Depending on the application, one may be interested in stopping times, objective reductions, distances to solution sets, contraction factors, oracle complexity measures, or probabilities of satisfying a prescribed solution criterion. Consequently, generalization should be formulated relative to an arbitrary optimization-specific performance measure rather than a single notion of error.

More fundamentally, we view optimization performance as a statistical quantity induced by a distribution of optimization problems and the resulting distribution of optimization trajectories. This perspective allows optimization performance itself to become the object of generalization analysis.

\subsection{PAC-Bayesian Generalization}

PAC-Bayesian theory provides high-probability guarantees relating empirical and population performance. Since its introduction, an extensive body of work has developed PAC-Bayesian bounds under a variety of assumptions and divergence measures \citep{Mc2003_1, Mc2003_2, Seeger_2002, Langford_ShaweTaylor_2002, Catoni_2004, Catoni_2007, Germain_Lacasse_Laviolette_Marchand_2009, Honorio_Jaakkola_2014, Begin_Germain_Laviolette_Roy_2016, London_2017, Alquier_Guedj_2018, Ohnishi_Honorio_2021, Amit_Epstein_Moran_Meir_2022, Haddouche_Guedj_2023}.

Most PAC-Bayesian analyses focus on prediction models. In contrast, we use PAC-Bayesian theory to study optimization performance and distributions over algorithmic parameters. The resulting guarantees are formulated in terms of optimization-specific performance measures rather than prediction losses. In this sense, PAC-Bayesian theory serves as a tool for studying learnable optimization algorithms rather than predictive models.

\subsection{Learning by Minimizing PAC-Bayesian Bounds}

PAC-Bayesian bounds are frequently used not only for analysis but also as learning objectives. Early examples include the work of \citet{Langford_Caruana_2001}, while more recent approaches directly optimize PAC-Bayesian upper bounds during training \citep{Dziugaite_Roy_2017, PRSS2021, Thiemann_Igel_Wintenberger_Seldin_2017}.

We follow this line of work and formulate learning as the minimization of a PAC-Bayesian upper bound. Similar to our prior work \citep{SFO25}, we employ data-dependent priors in order to improve both optimization performance and generalization.

\subsection{Relation to Stochastic Optimization}

Stochastic optimization studies the behavior of specific optimization algorithms under assumptions on the objective function, the stochastic oracle, or the noise model \citep{Bottou_Curtis_Nocedal_2018, DavisDrusvyatskiy2019, DavisDrusvyatskiy2022, Bianchi_Hachem_Schechtman_2022, DefossezBottouBachUsunier2022, KavisLevyCevher2022, SSSS2009_SCO}. The objective is typically to characterize convergence, stationarity, or complexity properties of a fixed algorithm.

The present work addresses a different problem. We do not study a particular optimization algorithm under structural assumptions such as convexity, smoothness, bounded gradients, or bounded stochastic noise. Instead, we view optimization algorithms themselves as statistical objects that can be learned, evaluated, and generalized. The central questions are therefore not convergence properties of a fixed algorithm, but the learning, performance, and generalization behavior of families of optimization algorithms over distributions of optimization problems.

\subsection{Relation to our Prior Work} 

The present work differs substantially from our earlier paper \emph{Learning-to-Optimize with PAC-Bayesian Guarantees: Theoretical Considerations and Practical Implementation} \citep{SFO25}. In the earlier work, the primary focus was the derivation of PAC-Bayesian guarantees for a specific class of learnable optimization algorithms and convergence-related performance measures. In contrast, the present paper develops a general statistical-learning framework for optimization algorithms. PAC-Bayesian analysis appears only as one application of this broader framework. Moreover, the framework accommodates a substantially wider range of optimization problems, performance measures, and algorithmic architectures, including stopping times, contraction factors, oracle-complexity measures, and trajectory-level properties. Thus, while both works employ PAC-Bayesian techniques, the main contribution of the present paper is the development of a general statistical perspective on optimization algorithms rather than the analysis of a particular learnable algorithm.

%%
%%\section{Preliminaries and Notation}\label{Sec:Preliminaries}
%%
%%We will endow every topological space $X$ with its Borel-$\sigma$-algebra $\mathcal{B}(X)$, and we assume it to be a Polish space, that is, it is separable and admits a complete metrization. Given two spaces $X$ and $Y$, we write their product as $X \times Y$, and, for products with a generic number of terms, we use $\prod_{n=1}^N X_n$. Similarly, the product-$\sigma$-algebra of $\mathcal{B}(X)$ and $\mathcal{B}(Y)$ on $X \times Y$ is denoted by $\mathcal{B}(X) \otimes \mathcal{B}(Y)$. Further, for a generic number of terms, we use $\bigotimes_{n=1}^N \mathcal{B}(X_n)$ to denote the product-$\sigma$-algebra, and, if $X_n \equiv X$ for all $n = 1, ..., N$, also $X^N$ and $\mathcal{B}(X)^{\otimes N}$. 
%%
%%We use the same notation for measures: Given two measures $\nu_1$ and $\nu_2$ on $X$ and $Y$, the corresponding product measure on $X \times Y$ is denoted by $\nu_1 \otimes \nu_2$. If the product involves a generic number of measures, this is denoted by $\bigotimes_{n=1}^N \nu_n$, and, if $\nu_n \equiv \nu$ for all $n = 1, ..., N$, we also use $\nu^{\otimes N}$. Further, if necessary, we will write the integral of a function $f: X \to \R$ w.r.t. $\nu$ in the operator notation, that is, $\nu[f] := \int_X \nu(dx) \ f(x) := \int_X f(x) \ \nu(dx)$. 
%%
%%\todo[inline]{Define the underlying probability space.}
%%

\section{A Probabilistic Framework for Optimization Algorithms}\label{sec:prob-framework}

Classical optimization studies the behavior of algorithms on individual optimization problems or classes of problems through worst-case guarantees. Learnable Optimization Algorithms (LOAs), in contrast, are designed and evaluated relative to a distribution of optimization problems. This shift from individual instances to populations of optimization problems raises new questions concerning performance, learning, and generalization that are not naturally addressed within the classical optimization paradigm.

The purpose of this section is to introduce a probabilistic framework for studying optimization algorithms from this perspective. The framework is built around four central concepts: \emph{probabilistic classes of optimization problems} (Section~\ref{sec:prob-class-problem}), \emph{learnable optimization algorithms} (Section~\ref{sec:intro-LOA}), \emph{probability distributions on optimization trajectories} (Section~\ref{sec:cond-dist-traj}), and \emph{statistical measures of algorithmic performance} (Section~\ref{sec:perf-measures}). Together, these concepts provide a common language for average-case analysis, generalization, and the statistical learning of optimization algorithms.

Our emphasis in this section is conceptual rather than technical. The goal is to identify the relevant probabilistic objects and explain their role in the study of learnable optimization algorithms. In particular, we deliberately suppress measure-theoretic details in order to make the underlying ideas more transparent. Whenever possible, we work directly with random variables, trajectory distributions, and performance functionals without discussing their explicit construction.

The mathematical foundations underlying these objects are developed separately. In particular, the construction of trajectory distributions, the associated probability kernels, and the required measurability results are deferred to the appendix. This allows us to first develop the conceptual framework and its connection to statistical learning, while maintaining full mathematical rigor in the background.

Throughout the section, we illustrate the framework using a simple running example that will accompany the theoretical developments and serve as a bridge between classical optimization, average-case analysis, and learnable optimization algorithms.

\subsection{A Probabilistic Class of Optimization Problems} \label{sec:prob-class-problem}

In analogy to the deterministic setting of optimization, a probabilistic theory of optimization should be built upon clearly specified mathematical objects. In order to compare optimization algorithms in a principled manner, we first introduce a probabilistic class of optimization problems. It consists of the following components: 
\begin{enumerate} 
    \item \textbf{Distribution over optimization problems.} Let $\spX$ be the optimization space and $(\ell(\cdot,\realPar))_{\realPar\in\spacePar}$ be a family of loss functions $\ell(\cdot,\realPar)\colon \spX\to[0,\infty]$ and let $\rvPar$ be a $\spacePar$-valued random variable with probability distribution $\distPar$. A realization $\realPar$ represents a concrete optimization problem, usually,
    \[
        \min_{\vx\in\spX}\, \ell(\vx,\realPar) \,,
    \]
    while the distribution $\distPar$ describes the probabilistic class of problems under consideration. The fundamental object is therefore not a single optimization problem but a probability distribution over optimization problems. 
    \item \textbf{Oracle.} We consider a local black-box oracle that provides information about a realization $\realPar$ of the optimization problem at a query point $\vx$. Typical examples include the objective value $\ell(\vx,\realPar)$, first-order information $\nabla \ell(\vx,\realPar)$, or higher-order information whenever it exists. It may also allow for stochastic estimates of the objective or its derivatives. The oracle specifies the information available to an optimization algorithm and therefore determines the admissible computational model. 
    \item The \textbf{family of solution criteria} is a measurable set $\solutionCrit\subset \spX\times\spacePar$ such that, for each realization $\realPar\in\spacePar$ of the optimization problem, we have a solution criterion on the optimization space $\solutionCrit_{\realPar}\subset\spX$ that defines what points in $\spX$ constitute a successful solution to problem $\realPar$. This formulation includes finite-time complexity measures (e.g. $\solutionCrit_{\realPar}=\set{\vx\in\spX\setsep \ell(\vx,\realPar) - \inf_{\vx\in\spX}\ell(\vx,\realPar)\leq \eps}$ for some $\eps>0$) as well as asymptotic optimization goals (e.g. convergence to a global minimizer $\solutionCrit_{\realPar}=\argmin_{\vx\in\spX} \ell(\vx,\realPar)$). For more details, see Section~\ref{appdx:examples-solution-crit}.
\end{enumerate}
\begin{Example}\label{ex:running-ex-toy-class}
    We illustrate the definitions on a simple example with $\spX=\R$.
    \begin{enumerate}
        \item  The \emph{distribution over optimization problems} is specified through the following quantities. Let $\rvPar$ be uniformly distributed on $\spacePar:=[m,L]$ with $0<m<L$ and define $\ell(x,\realPar) = \frac{\realPar}{2}x^2-x$. 
        \item Since all objectives in this problem distribution are differentiable, we consider a \emph{first-order oracle} that computes $\ell'(x,\realPar)$ at a query point $x$ for problem $\realPar$. 
        \item In theory, we can easily determine the minimizer of $\ell(\vx,\realPar)$ as $\vx^*_{\realPar}=\frac{1}{\realPar}$. Therefore, a suitable \emph{solution criterion} is defined, for $\eps>0$ and any $\realPar\in \spacePar$, as  
        \[
            \solutionCrit_{\realPar} = \set{\vx\in\spX\setsep \abs{\vx-\vx^*_\realPar}\leq \eps} \,.
        \]
    \end{enumerate}
\end{Example}

\subsection{Learnable Optimization Algorithms} \label{sec:intro-LOA}

A probabilistic class of optimization problems specifies the environment in which optimization algorithms operate. We now turn to the central object of this work, namely iterative optimization algorithms whose behavior can be adapted to a distribution of optimization problems. Let $\realPar\in\spacePar$ represent an optimization problem from the distribution $\distPar$. \\

We introduce a \emph{state space} $\spaceState$ that contains the optimization space $\spX$ as a subspace and consider iterative optimization algorithms $\LOA$ of the form 
\begin{equation}\label{eq:LOA-update}
    \iter{\realState}{t+1} = \LOA(\realHyp,\realPar,\iter{\realState}{t},\iter{\realRand}{t+1})\,,
\end{equation}
where $\realHyp$ represents the algorithmic parameters,
and $\iter{\realRand}{t}$ models internal stochasticity.  The algorithmic parameter $\realHyp\in\spaceHyp$ controls the behavior of the optimization algorithm and basically defines the collection of admissible algorithms. Classical optimization methods such as gradient descent, heavy-ball methods, or stochastic gradient methods correspond to particular choices of $\LOA$ and $\realHyp$, where $\LOA$ formalizes the update rule and $\realHyp$ is, for example, the step size that is used in the update rule (see Example~\ref{ex:running-ex-toy-alg}). Since, together they determine the algorithm, we introduce the abbreviation:
\[
    \LOA_{\realHyp}(\realPar,\iter{\realState}{t},\iter{\realRand}{t+1}) :=  \LOA(\realHyp,\realPar,\iter{\realState}{t},\iter{\realRand}{t+1})\,.
\]
In contrast to classic optimization, the central premise of Learnable Optimization Algorithms (LOAs) is that these parameters may themselves be adapted to the distribution of optimization problems and therefore constitute the \emph{learnable parameters} $\realHyp$. The aspect of learning in our framework is discussed thoroughly in Section~\ref{sec:statistical-learning}.
Defining the update mapping \eqref{eq:LOA-update} on a state space instead of the optimization space itself is classical and opens the door for a wealth of algorithms that require a memory of previous configurations. In the classic optimization setting, this extension allows for momentum or quasi-Newton updates. The iterates on the optimization space can be recovered by a simple projection $\proj{\spX}$ onto that subspace $\spX$.\\

The resulting optimization process is influenced by several sources of randomness: 
\begin{enumerate*}[series = tobecont, itemjoin = \ , label=(\roman*)]
\item \emph{Problem randomness.} The optimization problem is sampled from the random variable $\rvPar$ that obeys the problem distribution $\distPar$. 
\item \emph{Initialization randomness.} The initial state $\realState^{(0)}$ may be a sample from a random variable $\iter{\rvState}{0}$ that obeys $\distInit$. 
\item \emph{Internal randomness.} The update rule may involve a random variable $\iter{\rvRand}{t+1}$ that allows us to model stochastic algorithms.
\item \emph{Algorithm randomness.} The algorithm parameters may themselves be sampled from random variable $\rvHyp$, for example, when obtained through a learning procedure (that depends on a (random) dataset). 
\end{enumerate*}
As a consequence, the trajectory of iterates that is generated by the algorithm is understood as a sequence of random variables $(\iter{\rvState}{t})_{t\in\N}$ generated recursively by
\begin{equation} \label{eq:LOA-update-rv}
    \iter{\rvState}{t+1} = \LOA(\rvHyp,\rvPar,\iter{\rvState}{t},\iter{\rvRand}{t+1})\,.
\end{equation}
Optimization algorithms become probabilistic objects. The objective of the present framework is to analyze such algorithms statistically. We denote the learnable algorithm by $\LOA_{\rvHyp}$. Examples that illustrate the modeling flexibility of algorithms are developed in Section~\ref{appdx:examples-LOA}.

In Section~\ref{sec:perf-measures} we introduce and discuss how to measure the performance of a learned algorithm $\LOA_{\realHyp}$, where a \emph{learned algorithm} means that $\realHyp$ was fixed through a learning procedure. In contrast, Section~\ref{sec:statistical-learning} casts the learning of algorithms as a statistical learning problem, in which \emph{learnable algorithms} take the role of a predictor (learning function, hypothesis function). This will emphasize the role of \emph{generalization} in our framework. 

\begin{Example}\label{ex:running-ex-toy-alg}
    We continue Example~\ref{ex:running-ex-toy-class}. 
    As a \emph{class of admissible optimization algorithms}, we take inspiration from Gradient Descent and, for a given problem $\realPar$, define a learnable hyperparameter $\realHyp=\alpha\geq 0$ together with the update mapping 
    \[
        \iter{x}{t+1} 
        = \iter{x}{t} - \alpha \ell'(\iter{x}{t}, \realPar)
        = \iter{x}{t} - \alpha (\realPar x - 1)
        = (1-\alpha\realPar)\iter{x}{t} + \alpha \,,
    \]
    starting at $\iter{x}0 = 0$. This algorithm $\LOA_{\realHyp}$ directly operates in the optimization space $\spX=\R$ and is deterministic without internal randomness. 
\end{Example}
\begin{Rem}
    An important aspect that is deliberately left open in the present framework is the complexity of a class of learnable optimization algorithms. While admissibility ensures that an algorithm induces well-defined trajectory distributions and performance measures, admissibility alone does not characterize the expressive power of the underlying algorithm class. For example, a classical optimization method may be parameterized by only a few scalar quantities, whereas a modern learnable optimization algorithm may involve millions of trainable parameters. Yet the complexity of an optimization algorithm need not be fully reflected by the dimension of its parameter space: distinct parameterizations may induce similar optimization trajectories, while small architectural changes can drastically alter algorithmic behavior. This suggests that meaningful notions of algorithmic complexity may have to be defined in terms of the families of trajectory distributions or performance measures that an algorithm class can realize. Understanding the relationship between algorithmic expressiveness, optimization performance, computational efficiency, and statistical generalization constitutes an intriguing direction for future research.
\end{Rem}

\subsection{The Conditional Distribution of Trajectories} \label{sec:cond-dist-traj}

A central premise of the present work is that optimization algorithms are not viewed as fixed procedures but as objects that can be adapted to a distribution of optimization problems. Consequently, understanding the behavior of a learnable optimization algorithm requires more than analyzing individual iterates. Instead, one must describe the complete optimization process generated by the interaction of 
\begin{enumerate*}[series = tobecont, itemjoin = \ , label=(\roman*)]
\item the optimization problem $\realPar$, 
\item the algorithmic parameters $\realHyp$, 
\item the initialization $\realInit$, and 
\item possible sources of internal randomness $(\iter{\realRand}{t})_{t\in\N}$. 
\end{enumerate*} 
The outcome of this interaction is an optimization trajectory $(\iter{\realState}{t})_{t\in\N}$, which we regard as a trajectory-valued (i.e. $\spaceTraj$-valued) random variable $\rvTraj$. Samples from the distribution of trajectories are denoted by $\realTraj$.
Indeed, as listed above, the trajectory depends on the initialization and on possible sources of internal randomness. Consequently, even for fixed problem parameters $\realPar$ and algorithmic parameters $\realHyp$, the algorithm $\LOA_{\realHyp}$ generally generates a distribution of trajectories rather than a single deterministic trajectory.\\

Therefore, the basic object on which our subsequent notions of performance and generalization are built is the distribution of optimization trajectories induced by an optimization algorithm. We introduce a probabilistic description of optimization trajectories on the trajectory space and study how the resulting distributions depend on the problem parameter $\realPar$ and the algorithmic parameters $\realHyp$.  The trajectory distribution provides a probabilistic description of the behavior of an optimization algorithm that is independent of any particular performance measure. As will be seen in the next section, performance measures such as stopping times, convergence rates, and objective errors can all be expressed in terms of this distribution. To describe optimization trajectories probabilistically, we consider the joint distribution of algorithmic parameters, optimization problems, and trajectories:
\[
    \left( \distHyp \otimes \distPar\right) \otimes \psi \,,
\] 
where $\psi$ is a probability kernel from $\spaceHyp\times\spacePar$ to $\spaceTraj$, whose explicit construction is deferred to Section~\ref{appdx:dist-of-traj}. For fixed algorithmic parameters $\realHyp$ and a problem $\realPar$, the measure $\psi(\realHyp,\realPar,\cdot)$  
describes the probability distribution of the trajectory generated by $\LOA_{\realHyp}$ on $\realPar$. Equivalently, $\psi$ may be viewed as a regular conditional distribution of $\rvTraj$ given $(\rvHyp,\rvPar)$. Examples of algorithms from this viewpoint are shown in Section~\ref{appdx:examples-LOA-kernels}.

In the special case of a deterministic optimization algorithm, the conditional trajectory distribution reduces to a Dirac measure of the induced trajectory. In other words, in this case, every pair $(\realHyp,\realPar)$ generates a unique trajectory and 
\[ 
    \psi(\realHyp,\realPar,\cdot) = \delta_{\traj(\realHyp,\realPar)}\,, \quad \text{where}\ \traj(\realHyp,\realPar) := (\iter{\realState}{t})_{t\in\N}
\] 
Stochastic algorithms, random initializations, and randomized algorithmic parameters lead to genuinely non-degenerate trajectory distributions.

\begin{Example}\label{ex:running-ex-toy-prob-traj}
    We continue Example~\ref{ex:running-ex-toy-alg}. For fixed parameters $(\realHyp,\realPar)$ and initialization $\vx^{(0)}=0$, the optimization trajectory is deterministic and hence uniquely determined. Consequently, the corresponding conditional trajectory distribution is given by the Dirac measure $\psi(\realHyp,\realPar,\cdot) = \delta_{\traj(\realHyp,\realPar)}$ of the trajectory generated by gradient descent with step size $\realHyp$ for problem $\realPar$. Although the optimization algorithm itself is deterministic, the uniformly distributed random problem parameter $P$ in $[m,L]$ induces a probability distribution on the collection of trajectories. Moreover, different choices of the step size $\realHyp$ lead to different distributions of optimization trajectories. This simple example illustrates the central viewpoint of the present framework: optimization algorithms induce probability laws on trajectory space through the interaction of a problem distribution and an algorithmic update rule.
\end{Example}

\subsection{Statistical Performance of an Algorithm}\label{sec:perf-measures}

A probabilistic class of optimization problems specifies what constitutes a successful solution through a collection of solution criteria. Such criteria provide a binary notion of success: for a given optimization problem and trajectory, the criterion is either satisfied or violated. They therefore answer the question of whether an optimization problem has been solved.
\begin{Def}
    Let $\realPar\in\spacePar$ and let $\traj=(\iter{\realState}{t})_{t\in\N}\in\spaceTraj$ be a trajectory. The trajectory $\traj$ is said to \emph{solve} the optimization problem $\realPar$ if there exists $t\in\N$ such that 
    \[ 
    \bigl( \proj{\spaceState}(\iter{\realState}{t}), \realPar \bigr) \in \solutionCrit \,.
    \]
\end{Def}

Solving a problem and assessing the quality of a solution process, however, are fundamentally different notions. Two optimization algorithms may both solve the same problem while exhibiting very different behavior in terms of speed, robustness, computational cost, or solution quality. To capture such aspects, we introduce performance functionals. Unlike solution criteria, which merely distinguish success from failure, performance functionals quantify specific properties of optimization trajectories, such as stopping times, objective values, distances to solution sets, contraction factors, or oracle complexity measures.
\begin{Def}[Performance Functional]
A \emph{performance functional} is a measurable mapping
\[
    \map{\performFunc}{\spacePar \times \spaceTraj}{\eR}\,,
\]
where $\eR:=[-\infty,\infty]$.
\end{Def}
A performance functional assigns a numerical value to a problem-trajectory pair and is, therefore, an instance-wise quantity. Statistical notions of optimization performance arise by aggregating them over a distribution of optimization problems.\\

\begin{figure}[t!]
    \centering
    \tikz{
      \node (A) {\includegraphics[width=\textwidth]{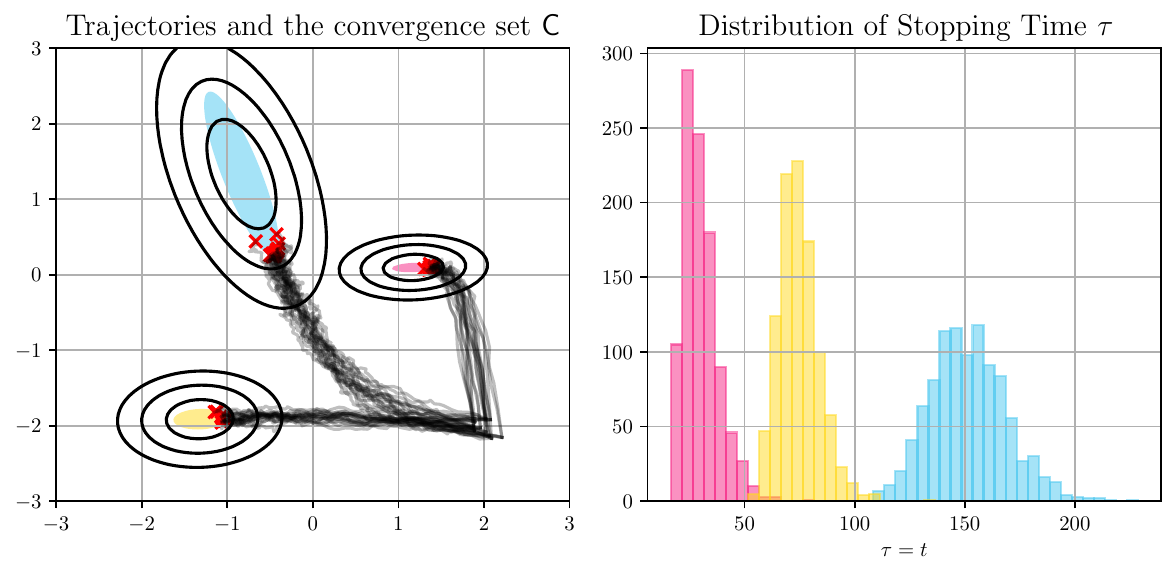}};
      \draw (A.north west) ++ (0.8,-0.1) coordinate (A1);
      \draw (A1) ++ (6.7,-0.6) coordinate (A2);
      \draw[white,fill=white] (A1) rectangle (A2);  
      \draw (A1) node[below right=0mm and 2mm] {Trajectories and solution criterion $\solutionCrit$};

      \draw (A1) ++ (7.7,0) coordinate (B1);
      \draw (B1) ++ (6.7,-0.6) coordinate (B2);
      \draw[white,fill=white] (B1) rectangle (B2);  
      \draw (B1) node[below right=0mm and 5mm] {Distribution of stopping time $\tau$};
    }
    \caption{Visualization of the stopping time $\tau$: In the left plot, the solution criterion for three problems is shown as shaded regions. As soon as a trajectory enters this region (red crosses), the algorithm is stopped. This yields the distribution of $\tau$ depending on the problem parameter $\realPar$, as shown in the right plot.}
    \label{fig:stopping_times}
\end{figure}

The framework accommodates a broad range of performance functionals. The verification of measurability for the following examples is deferred to Section~\ref{appdx:perform-func}. Many important examples can be expressed in terms of a measurable \emph{optimality-gap function} $\map{\lyap}{\spaceState\times\spacePar}{[0,\infty]}$, typically chosen as a Lyapunov function, for example, the objective value gap $\lyap(\realState,\realPar) = \ell(\vx,\realPar) - \min_{\vx\in\spX}\ell(\vx,\realPar)$, where $\vx=\proj{\spX}(\realState)$, the distance to a solution set $\lyap(\realState,\realPar) = \operatorname{dist} \bigl( \vx, \spX^\star_{\realPar} \bigr)$, where $\spX^\star_{\realPar} := \argmin_{\vx\in\spX} \ell(\vx,\realPar) $, or other problem-specific measures of progress. We include finite-horizon quantities, such as the optimality-gap after a prescribed number of iterations $k\in\N$, e.g.,
\begin{enumerate}
\item \textbf{$k$-step optimality-gap:}
\begin{equation}\label{eq:perf-func-k-step-gap}
\performFunc_k(\realPar,\traj) := \lyap(\iter{\realState}{k},\realPar)  \,.
\end{equation}
\end{enumerate}
Moreover, we have trajectory-level quantities, such as, for example, the following:
\begin{enumerate}\addtocounter{enumi}{1}
\item \textbf{Stopping time:} The minimal number of iterations to satisfy the solution criterion (see Figure~\ref{fig:stopping_times} for a schematic illustration)
\begin{equation} \label{eq:def-stopping-time-func}
    \tau(\realPar,\traj) := \inf\set{ t\in\N \setsep \iter{\vx}{t} \in \solutionCrit_{\realPar} } \,,\quad\text{where}\ \iter{\vx}{t}=\proj{\spaceState}(\iter{\realState}{t})\,.
\end{equation}
\item \textbf{Convergence rate:} Estimates the linear convergence rate of the optimality-gap (the rate is only linear, if this quantity is less than $1$), if the denominator is strictly positive:
\begin{equation}\label{eq:def-convergence-rate-perf}
    \contrate(\realPar,\traj) = \sup_{t\geq 0 } \frac{\lyap(\iter{\realState}{t+1},\realPar)}{\lyap(\iter{\realState}{t},\realPar)} \,.
\end{equation}
\item \textbf{Contraction Factor:} If the stopping time is finite, we can estimate a finite-time contraction factor (assuming $\lyap(\iter{\realState}{0},\realPar)>0$) as follows (and set $\contrate_{\tau}(\realPar,\traj)=0$ for $\tau=0$):
\begin{equation}\label{eq:def-contraction-factor}
    \contrate_{\tau}(\realPar,\traj) := \left( \frac{ \lyap(\iter{\realState}{\tau},\realPar) }{ \lyap(\iter{\realState}{0},\realPar) } \right)^{1/\tau}\,.% \mathds{1}_{\{\tau\ge1\}}(\tau)\,.
\end{equation}
For example, the stopping time may remain finite for an $\eps$-optimal solution criterion with $\eps>0$. Also note that \eqref{eq:def-contraction-factor} can be generalized to include stopping times $\tau=\infty$. 
\item \textbf{Indicators of trajectory properties:} For a measurable set $\setA\subset\spacePar\times\spaceTraj$, where $(\realPar,\traj)\in\setA$ indicates that the trajectory $\traj$ generated for problem $\realPar$ possesses a desired property, a corresponding performance functional is simply the indicator function 
\begin{equation}\label{eq:def-convergence-probability-perf}
    \performFunc(\realPar,\traj) = \mathds{1}_{\setA}(\realPar,\traj) 
    :=\begin{cases}
        1\,,&\ \text{if}\ (\realPar,\traj)\in\setA\,; \\
        0\,,&\ \text{otherwise}\,.
    \end{cases}
\end{equation}
\item \textbf{Oracle complexity measures}, defined for example as the number of oracle evaluations required until a solution is found.
\end{enumerate}
While a performance functional measures behavior on a single problem-trajectory pair, the goal of learnable optimization is to assess performance over an entire distribution of optimization problems. This motivates the introduction of statistical performance measures.

Recall that the probabilistic framework provides a joint probability measure $(\distHyp\otimes\distPar)\otimes\psi$ of algorithmic parameters, optimization problems, and trajectories, where $\psi$ denotes the conditional trajectory distribution from Section~\ref{sec:cond-dist-traj}. For fixed algorithmic parameters $\realHyp$, the problem-trajectory random variable $(\rvPar,\rvTraj)$ obeys the distribution of problem-trajectory pairs that is induced by the problem distribution $\distPar$ and the probability kernel $\psi$ (conditioned on the algorithmic parameter $\rvHyp=\realHyp$).
\begin{Def}[Statistical Performance Measure] \label{def:performance-measure}
Let $\performFunc$ be a performance functional and let 
$
\prob_{(\rvPar,\rvTraj)\condto\rvHyp=\realHyp} = \distPar \otimes \psi(\realHyp,\rvPar,\cdot)
$
denote the conditional distribution of
problem-trajectory pairs. A \emph{statistical performance measure} is a mapping %$\performMeas$
\[
    \performMeas \Bigl( \performFunc, \prob_{(\rvPar,\rvTraj)\condto\rvHyp=\realHyp} \Bigr) \in\R. \]
\end{Def}
This abstract definition allows one to model various notions of (statistical) algorithmic performance measures.
\begin{enumerate}
    \item \textbf{Expected Performance.} The most fundamental statistical performance measure is the expected performance
    \begin{equation}
    \label{eq:def-expected-performance}
    \performMeas_{\mathrm{exp}} = \Exp\set{ \performFunc(\rvPar,\rvTraj) \condto \rvHyp=\realHyp } =
        \intgr{\spacePar}{}{ \intgr{\spaceTraj}{}{\performFunc(\realPar,\traj)}{\psi(\realHyp,\realPar,\df\traj)}}{\distPar(\df\realPar)}\,.
    \end{equation}
    Examples include expected stopping times, expected objective gaps, expected oracle complexity, and expected convergence rates.
    \item \textbf{Tail Probabilities.} For a fixed threshold $\alpha\in\R$, one may consider 
    \[
        \performMeas_{\mathrm{tail}} = \prob\set{ \performFunc(\rvPar,\rvTraj) \leq \alpha \condto \rvHyp=\realHyp } \,.
    \]
    This includes, for example, solution probabilities and high-probability performance guarantees.
    \item \textbf{Quantiles.} A robust alternative to expectations is provided by quantiles. For $\beta\in(0,1)$, the $\beta$-quantile of the performance functional is defined by
    \[
        \Phi_{q_\beta} = \inf\set{ \alpha\in\R \setsep \prob\set{ \performFunc(\rvPar,\rvTraj) \leq \alpha \condto \rvHyp=\realHyp } \geq \beta }\,.
    \]
    Quantiles naturally describe the performance achieved on a prescribed
    fraction of optimization problems and allow one to distinguish typical
    behavior from rare atypical instances.
    \item \textbf{Risk Measures.} More generally, one may consider risk measures such as conditional value-at-risk (CVaR), which interpolate between average-case and worst-case notions of performance by placing increased emphasis on the most difficult problems.
\end{enumerate}

The probabilistic viewpoint allows one to quantify trade-offs that are invisible from a purely worst-case perspective. An algorithm may perform substantially better on the overwhelming majority of problems while accepting a small probability of poor performance on rare or atypical instances. Similar trade-offs are central in statistical learning theory, where one is often willing to accept a small probability of prediction error in exchange for significantly improved average performance.

Importantly, the ultimate objective remains instance-wise. Even though performance is evaluated statistically over a distribution of optimization problems, the learned optimization algorithm is eventually applied to individual realizations. The role of the distribution is therefore not to replace individual optimization problems by an average problem. Rather, it serves as a source of statistical information from which optimization algorithms can learn. In this sense, optimization performance generalizes across a distribution of optimization problems in much the same way that prediction performance generalizes across a distribution of data in statistical learning, leading to the statistical learning aspect of our framework in the upcoming section.\\

We conclude this section by illustrating three conceptually distinct statistical performance measures using the running example. Example~\ref{ex:running-example-stopping-time} considers the expected stopping time and highlights the connection between solution criteria and optimization performance. Example~\ref{ex:running-ex-toy-performance} shows that different performance functionals induce different notions of optimality. Finally, Example~\ref{ex:toy-success-prob-ourlier-case} demonstrates how high-probability performance measures naturally capture trade-offs between performance and robustness in the presence of rare outlier problems.
\begin{Example}[stopping time] \label{ex:running-example-stopping-time}
    We continue Example~\ref{ex:running-ex-toy-prob-traj} and define, for some $\eps>0$, the solution criterion 
    \[ 
        \solutionCrit_{\realPar,\varepsilon} = \set{ x : (x-x^*_{\realPar})^2 \leq \eps } \,.
    \] 
    As a particularly important performance measure for optimization, we consider the expected stopping time for the Gradient Descent-like algorithm from Example~\ref{ex:running-ex-toy-alg}, i.e., we measure the expected performance \eqref{eq:def-expected-performance} with respect to the stopping time functional in \eqref{eq:def-stopping-time-func}. 

    For the analysis, we abbreviate $\iter{e_\realPar}{t+1}:=(\iter{x}{t+1}-x_\realPar^*)$ and observe that
    \begin{equation}\label{eq:ex-toy-stopping-time-error-def}
        (\iter{e_\realPar}{t+1})^2
        = (1-\alpha\realPar)^2 (\iter{e_\realPar}{t})^2
        = (1-\alpha\realPar)^{2t} (\iter{e_\realPar}{0})^2 \,,
    \end{equation}
    which yields $\tau_\eps(\realPar,\traj) = \ceil{\log(\eps (\iter{e_\realPar}{0})^{-2})/(2\log(1-\alpha\realPar))}$, assuming that $0<\alpha\realPar<1$. Consequently, the expected stopping-time (expected iteration-complexity) is given by
    \[
        \Exp\set{\ceil{\log(\eps (\iter{e_\rvPar}{0})^{-2})/(2\log(1-\rvHyp\rvPar))} \condto \rvHyp =\alpha} \,.
    \]
    This quantity does not admit a simple closed-form minimizer with respect to the step size. Nevertheless, it provides a direct measure of the computational effort required to solve a typical optimization problem drawn from the distribution $\distPar$. 
    
    Stopping times are particularly important because they connect the solution criteria defining the optimization problem with statistical performance measures. %They will also play a central role in the PAC-Bayesian generalization guarantees developed later in this work. 
\end{Example}
\begin{Example}[expected squared and expected absolute error vs. worst case] \label{ex:running-ex-toy-performance}
    We continue Example~\ref{ex:running-ex-toy-prob-traj}. Consider the expected squared error after $t$ steps as performance functional
    $%$\[
        \performFunc_t^{\mathrm{sq}}(\realPar, (\iter{\vx}{t})_{t\in\N}) = (\iter{\vx}{t} - \vx_\realPar^*)^2% \,.
    $. %\]
    Since the Gradient Descent-like algorithm from Example~\ref{ex:running-ex-toy-alg} is deterministic, the conditional trajectory distribution reduces to a Dirac measure, and thus the performance measure is
    \[
        \Exp\set{ \performFunc_t^{\mathrm{sq}}(\rvPar, \rvTraj)\condto \rvHyp=\alpha }  
        %= \performFunc(\rvPar, \traj(\rvPar,\rvHyp))
        = \Exp\set{(\iter{\rvX}{t} - \vx^*_{\rvPar})^2\condto \rvHyp=\alpha} \,.
    \]
%%    For the analysis, we abbreviate $\iter{e_\realPar}{t+1}:=(\iter{x}{t+1}-x^*)$ and observe that
%%    \[
%%        (\iter{e_\realPar}{t+1})^2
%%        = (1-\alpha\realPar)^2 (\iter{e_\realPar}{t})^2
%%        = (1-\alpha\realPar)^{2t} (\iter{e_\realPar}{0})^2 \,,
%%    \]
%%    which yields the expected performance 
    Using $\iter{e_\realPar}{t+1}:=(\iter{x}{t+1}-x^*)$ from Example~\ref{ex:running-example-stopping-time}, we obtain the expected performance
    \begin{equation} \label{ex:running-ex-toy-performance-expected-performance}
        \Exp\set{(\iter{e_\rvPar}{t+1})^2\condto \rvHyp=\alpha} 
    %    = \frac{(\iter{e_\realPar}{0})^2}{L-m}\intgr{m}{L}{(1-\alpha \realPar)^{2t}}{\df\realPar}
        = \frac{(\iter{e_\realPar}{0})^2\Big((1-\alpha m)^{2t}-(1-\alpha L)^{2t}\Big)}{(2t+1)\alpha(L-m)} \,.
    \end{equation}
    In order to find the algorithm with the best expected performance, this expression needs to be minimized. %, e.g., via the optimality condition $\Exp\set{\rvPar (1-\alpha\rvPar)^{2t-1}} = 0$, which has a closed from solution 
    In the case $t=1$, there is a closed-form solution:
    \[
        \alpha_{\text{exp-sq}} = \frac{3(m+L)}{2(m^2 + mL + L^2)}\,.
    \]
    In contrast, the classical worst-case step size (for the class of $m$-strongly convex and $L$-smooth functions, which naturally embeds our problem distribution) is known to be
    \[
        \alpha_{\text{wc}} = \frac{2}{L+m} \,.
    \]
    Thus, average-case and worst-case performance lead to different notions of optimality. Moreover, using convergence of $L^p$-norms to the supremum norm, one obtains
    \[ 
        \lim_{t\to\infty} \Big( \Exp[(1-\alpha\rvPar)^{2t}] \Big)^{1/(2t)} = \sup_{\realPar\in[m,L]} \abs{1-\alpha\realPar}\,,
    \] 
    showing that long-horizon expected performance converges to a worst-case criterion. On an intuitive level, for $t\to\infty$, the cost of expected performance is dominated by small instabilities and problems for which the algorithm diverges, hence for having a low expected performance those cases must be eliminated, leading to the worst-case. \\

    Taking the absolute error instead of the squared error yields, for the case $t=1$, %the optimal step size 
    \[
        \alpha_{\text{exp-abs}} = \sqrt{\frac{2}{m^2 + L^2}}\,,
    \]
    which, again, for $t\to\infty$, converges to the worst-case performance. 
    Again for $t\to\infty$, the expected absolute error cost converges to the worst-case performance. 
\end{Example}
\begin{Example} \label{ex:toy-success-prob-ourlier-case}
    We consider a slight modification of our running example in Example~\ref{ex:running-ex-toy-class}, \ref{ex:running-ex-toy-alg}, and \ref{ex:running-ex-toy-prob-traj}. The distribution of optimization problems is generated as $\ell(\vx,\realPar) = \frac{\realPar_L}{2}\vx^2 - \realPar_b\vx$ with $\realPar=(\realPar_L,\realPar_b)$, where $\realPar_b$ is uniformly distributed on $[0,1]$ and $\realPar_L$ follows the mixture distribution 
    \[ 
        \prob(\realPar_L=m)=0.99\,, \qquad \text{and}\quad
        \prob(\realPar_L=L_{\rm outlier})=0.01\,,
    \] 
    with $L_{\rm outlier}\gg m$.
    %To emphasize the effect, we consider the case $m=L$. 
    Then $99\%$ of problems have the same curvature constant $m=L$, while the remaining $1\%$ consist of outlier instances with $L_{\rm outlier}$. 
     
    The Gradient Descent-like algorithm from Example~\ref{ex:running-ex-toy-alg} with $\alpha=\frac{1}{L}$ converges for the case $\realPar_L=m$ in a single step. However, in worst-case scenario, the optimal step size for $m$-strongly convex and $L_{\text{outlier}}$-smooth functions is $\alpha_{\text{wc}}=\frac2{m+L_{\text{outlier}}}$ and we only obtain a linear rate of convergence for all problems in this distribution. 

    Using the solution probability after one iteration as a statistical performance measure, 
    \[ 
        \performMeas_{\rm succ} = \prob\!\left( \rvX^{(1)}=\rvX^\star_{\rvPar} \condto \rvHyp=\alpha \right)\,,
    \] 
    the aggressive choice $\alpha=1/L$ achieves a performance of $99\%$, whereas the worst-case step size achieves a substantially smaller solution probability. 
    
    This example illustrates a key feature of the probabilistic framework. A worst-case analysis is governed entirely by the rare outlier instances and therefore seeks to eliminate failure for every possible problem. In contrast, a high-probability performance measure allows one to trade a small probability of failure for significantly improved performance on the overwhelming majority of optimization problems.
\end{Example}
\begin{Rem}
    The optimal choice of step size in Example~\ref{ex:toy-success-prob-ourlier-case} would be problem-adaptive (or instance-adaptive), which we discuss in Section~\ref{sec:problem-adaptive-learning}, in general.
\end{Rem}

\subsection{Beyond Distribution-Adaptive Learning} \label{sec:problem-adaptive-learning}

The running examples suggest another possible form of adaptation. In the quadratic example, the optimal step size depends directly on the problem parameter. If the problem parameter were known, one could choose the step size specifically for the realized optimization problem rather than for the entire distribution of problems.
The present work focuses on \emph{distribution-adaptive} learning. The algorithmic parameters are learned from a collection of optimization problems and subsequently applied to previously unseen instances drawn from the same distribution. Consequently, the algorithmic parameters are chosen independently of the particular problem realization. 

A natural extension consists of learning algorithmic parameters as a function of the observed optimization problem. In this setting, the algorithmic parameters and the optimization problem become statistically dependent, yielding a more expressive form of adaptation. Conceptually, this shifts the learning problem from selecting algorithmic parameters to learning a mapping from optimization problems to algorithmic parameters. 
The probabilistic framework developed in this work naturally accommodates such problem-adaptive optimization algorithms through the induced trajectory distributions and performance measures. However, the PAC-Bayesian theory developed in the following sections is tailored to the distribution-adaptive setting and studies distributions over algorithmic parameters rather than distributions over problem-dependent mappings. Extending the generalization theory to the latter setting is an interesting direction for future research.

\section{Statistical Learning of Learnable Optimization Algorithms (LOA)} \label{sec:statistical-learning}

The previous sections introduced probabilistic classes of optimization problems, learnable optimization algorithms, trajectory distributions, and performance measures. We now turn to the central question underlying the present work:
\begin{center}
What does it mean to learn an optimization algorithm?
\end{center}
At first sight, the proposed framework appears closely related to statistical learning theory. Indeed, optimization algorithms are trained on samples of optimization problems and subsequently evaluated on previously unseen problems drawn from the same distribution. Consequently, notions such as training, generalization, overfitting, sample complexity, and distribution shift arise naturally.

However, there is a fundamental difference. In statistical learning, the learned object is typically a predictor
%\[
%h:\mathcal X \to \mathcal Y,
%\]
whose purpose is to infer an unknown relationship between inputs and outputs. Performance is measured through prediction errors on unseen examples.
In the present setting, the learned object is not a predictor but an optimization algorithm. More precisely, we consider a parameterized family of algorithms as in \eqref{eq:LOA-update}, which are repeatedly applied to optimization problems drawn from a probability distribution. The objective is not to infer an unknown input-output map. Instead, the objective is to improve the computational performance of an optimization procedure by exploiting statistical information about the distribution of optimization problems introduced in Section~\ref{sec:perf-measures}.
This distinction is important. For a given optimization problem, the desired solution is already specified by the optimization objective itself and can in principle be approximated using classical optimization methods. Learning is therefore not used to recover an unknown target function. Rather, learning is used to construct optimization algorithms that achieve improved performance on future optimization problems.

\subsection{The Learning Problem}\label{sec:learning-problem}

Let $(\ell(\cdot,\realPar))_{\realPar\in\spacePar}$ be a probabilistic class of optimization problems and let $\rvPar\sim\distPar$ denote a random optimization problem. Assume that we are given a dataset $(\realPar_1,\ldots,\realPar_{\sizeData})$ consisting of $\sizeData$ optimization problems, which are realizations of $(\rvPar_1,\ldots,\rvPar_{\sizeData})$ that are independent and identically distributed to $\distPar$.
The goal of learning a Learnable Optimization Algorithm (LOA) $\LOA_{\realHyp}$ of the form \eqref{eq:LOA-update} is to determine algorithmic parameters $\realHyp\in\spaceHyp$ that achieve the best performance on future optimization problems drawn from the same distribution. In contrast to classical optimization, the algorithmic parameters are not chosen a priori but are adapted to statistical information contained in the dataset.

Given a performance functional $\map{\performFunc}{\spacePar\times \spaceTraj}{\R}$, the quality of an optimization algorithm is assessed through its statistical performance. For a fixed choice of algorithmic parameters $\realHyp$, the \emph{population performance} is defined by
\[
\risk(\LOA_{\realHyp}) := \Exp\set{ \performFunc(\rvPar,\rvTraj) \condto \rvHyp=\realHyp },
\]
which describes the expected performance of the optimizer on previously unseen optimization problems. Recall that this quantity coincides with the expected performance introduced in \eqref{eq:def-expected-performance}.
Since the distribution $\distPar$ is not directly accessible, the population performance cannot be evaluated exactly. Instead, one only has access to the finite dataset $(\realPar_1,\ldots,\realPar_{\sizeData})$  and therefore considers the \emph{empirical performance}
\[
\empRisk_{\sizeData}(\LOA_{\realHyp}) := \frac{1}{\sizeData} \sum_{n=1}^{\sizeData} \Exp\set{ \performFunc(\rvPar_n,\rvTraj_n) \condto \rvHyp=\realHyp,\, \rvPar_n=\realPar_n } \,.
\]
A learned optimization algorithm is obtained by selecting algorithmic parameters that achieve favorable empirical performance while aiming to simultaneously minimize the population performance. In this sense, the learning problem amounts to minimizing the empirical performance and hoping that the resulting optimizer also performs well with respect to the population performance.

At a conceptual level, this resembles a standard statistical learning problem. Optimization algorithms are trained on a finite sample and subsequently evaluated on previously unseen instances drawn from the same distribution. Consequently, familiar notions such as training, generalization, overfitting, sample complexity, and robustness arise naturally in the present setting. There is, however, an important conceptual difference. In statistical learning, the goal is to infer an unknown input-output relationship from data. The learned object is typically a predictor and performance is measured through prediction accuracy on unseen examples. In the present framework, the learned object is not a predictor but an optimization algorithm.
Moreover, the objective is not to recover an unknown solution map. For every realization $\realPar$, the optimization problem is already specified by the objective function $\ell(\cdot,\realPar)$ and can, at least in principle, be solved using a classical optimization algorithm. Learning is therefore not used to infer the solution itself. Instead, learning is used to improve computational performance by exploiting statistical regularities present in a distribution of optimization problems. Consequently, generalization is not sought with respect to a prediction error but with respect to a chosen performance measure. Depending on the application, the performance functional $\performFunc$ may describe stopping times, convergence rates, objective errors, solution probabilities, robustness measures, or oracle complexity. Different performance functionals generally induce different notions of optimality and may therefore lead to different learned optimization algorithms.

This gives rise to a performance-generalization trade-off that is analogous to the accuracy-generalization trade-off in statistical learning. By exploiting statistical structure present in the problem distribution, a learned optimizer may achieve substantially improved performance on typical optimization problems. Such gains may come at the cost of weaker guarantees on rare or atypical instances. In particular, a small loss in generalization may permit a significant improvement in computational performance.

To study this trade-off rigorously, the dataset itself is modeled as a random variable $\rvPar_{[\sizeData]}:=(\rvPar_1,\ldots,\rvPar_{\sizeData})$. Consequently, the learned optimization algorithm depends on the random training dataset. Generalization guarantees therefore quantify the discrepancy between empirical and population performance with high probability over the draw of $\rvPar_{[\sizeData]}$. The rigorous development of the (empirical) distribution of trajectories over a dataset is subject of Section~\ref{appdx:learning-from-data}.

PAC bounds provide such guarantees for fixed algorithmic parameters $\realHyp$. PAC-Bayesian bounds extend this perspective to randomized optimization algorithms and distributions over algorithmic parameters. Accordingly, the PAC-Bayesian framework models the algorithm as in \eqref{eq:LOA-update-rv}, in which algorithmic parameters are also treated as a random variable $\rvHyp$, and studies the generalization behavior of distributions over optimization algorithms rather than individual parameter choices.

\subsection{Generalization of Optimization Algorithms}

The central question is no longer whether a particular optimization algorithm converges. Instead, the central question becomes whether the performance of a learned optimization algorithm generalizes. Recall the convergence is only a particular example that can be modeled using a performance measure. 
 In supervised learning, generalization refers to predictive performance on previously unseen inputs. In contrast, generalization in the present framework refers to the behavior of an optimization algorithm on previously unseen optimization problems.
Consequently, generalization is always defined relative to a chosen performance measure. Different performance measures generally lead to different notions of generalization and may therefore induce different learned optimization algorithms. This dependence on the performance measure is not a weakness but a natural feature of the framework. Optimization algorithms are designed for different purposes and should therefore be evaluated according to criteria that reflect their intended use.

\subsection{Performance versus Generalization}

As in classical statistical learning, learning optimization algorithms
naturally introduces a trade-off between performance and generalization. An optimization algorithm that is strongly specialized to a narrow family of optimization problems may achieve excellent empirical performance while generalizing poorly outside the observed training sample. Conversely, highly robust algorithms may generalize well but fail to exploit statistical structure present in the problem distribution. A particularly important aspect of this trade-off is the possibility of accepting a controlled probability of failure in exchange for significantly improved performance. To illustrate this phenomenon, consider the running Example~\ref{ex:toy-success-prob-ourlier-case} in which $99\%$ of the optimization problems possess curvature parameter $m$, while the remaining $1\%$ correspond to rare outlier problems with curvature $L_{\rm out}\gg m$.
 A classical worst-case analysis is governed by the outliers and therefore requires a step size that remains stable for all possible realizations. As a result, convergence on the typical problems may be unnecessarily slow.

In contrast, a distribution-aware optimization algorithm may deliberately select more aggressive parameters that solve the vast majority of optimization problems substantially faster while accepting a small probability of failure on rare atypical instances. High-probability performance guarantees naturally capture this trade-off.
From this perspective, learnable optimization algorithms play a role analogous to learned models in statistical learning theory. Both seek to exploit statistical regularities present in a population of tasks. However, whereas statistical learning aims to improve predictive accuracy, learnable optimization aims to improve computational performance.

\subsection{A Statistical View of Optimization Algorithms}

The preceding discussion suggests a shift in perspective. Classical optimization treats algorithms as fixed objects and studies their behavior under various assumptions on the optimization problem. The framework developed in this work instead treats optimization algorithms as statistical objects that can be trained, generalized, and evaluated on
distributions of optimization problems. The probabilistic class of optimization problems provides the environment from which optimization problems are sampled. Performance functionals specify what aspects of optimization behavior are of interest. Learning consists of adapting algorithmic parameters to the observed optimization problems, while generalization quantifies how well the resulting optimizer performs on future problems.
This perspective naturally gives rise to questions of sample complexity, overfitting, robustness, transferability, and distribution shift for optimization algorithms. In the remainder of this work, we focus on PAC-Bayesian generalization guarantees as a first step towards a statistical learning theory of optimization algorithms.

\section{PAC-Bayesian Generalization Guarantees}\label{sec:PAC-Bayesian-guarantees}

The preceding discussion establishes learnable optimization algorithms as statistical objects and identifies generalization as a central theoretical question. Various approaches from statistical learning theory may be employed to address this question. In the present work, we focus on a PAC-Bayesian approach, which is particularly well suited to randomized optimization algorithms and distributional notions of performance. In Subsection~\ref{subsec:PAC-performance-functionals}, we establish a PAC-Bayesian generalization result for general performance functionals, which are then specialized to stopping times and convergence rates. Subsection~\ref{subsec:PAC-success-probability} develops a general statement for generalizing the solution probability.

PAC-Bayesian theory provides high-probability bounds relating population performance to empirical performance while simultaneously quantifying the complexity of the learned model through its deviation from a prior distribution. In contrast to classical PAC bounds, which typically study a single hypothesis, PAC-Bayesian bounds naturally accommodate randomized hypotheses and distributions over hypotheses. This makes them particularly well suited to the present setting, where the learned object is not a predictor but a distribution over algorithmic parameters. 

Within the proposed framework, the role of the loss function is played by a performance functional defined on problem-trajectory pairs. Consequently, PAC-Bayesian theory allows us to derive generalization guarantees for a broad range of optimization-specific performance measures, including stopping times, convergence indicators, objective errors, and solution probabilities. In this sense, the following results provide guarantees for the generalization of optimization performance rather than prediction performance.

\paragraph{Notation.} In order to present the PAC-Bayesian framework, we need some notation. We denote the space of measures on $X$ by $\mathcal{M}(X)$ and for an absolutely continuous probability measure $\nu$ w.r.t. a reference measure $\mu \in \mathcal{M}(X)$, we use the standard notation $\nu\ll\mu$. Further, if necessary, we will write the integral of a function $f: X \to \R$ w.r.t. $\nu$ in the operator notation, that is, $\nu[f] := \int_X \nu(dx) \ f(x) := \int_X f(x) \ \nu(dx)$. 
In this context, the Kullback-Leiber divergence between two measures $\mu$ and $\nu$ is defined as:
$$
\divergence{\rm{KL}}{\nu}{\mu} = \begin{cases}
    \nu[\log(f)] = \int_\space{X} \log(f(x)) \ \nu(dx), &\text{$\nu \ll \mu$ with density $f$}\,, \\ +\infty, &\text{otherwise}\,.
\end{cases}
$$ 
For notational simplicity, we use $x_{[N]}:=(x_1,\ldots,x_N)$.

\subsection{Generalization of Abstract Performance Functionals} \label{subsec:PAC-performance-functionals}

%Having formulated the learning of optimization algorithms as a statistical learning problem, we now investigate the generalization of optimization performance. In particular, we seek to relate the population performance of a learned optimization algorithm to its empirical performance observed on a finite collection of training problems. To this end, we adopt a PAC-Bayesian perspective and study distributions over algorithmic parameters. The resulting bounds apply to arbitrary bounded performance functionals and therefore yield generalization guarantees for quantities such as stopping times, convergence rates, objective gaps, and solution probabilities.

We recall the conditional probability measure on problem-trajectory pairs $\prob_{(\rvPar,\rvTraj)\condto \rvHyp=\realHyp}$ given algorithmic parameters $\realHyp$ as in Definition~\ref{def:performance-measure}, which by construction yields a random variable on $\spaceHyp$ and thus can be integrated with respect to an arbitrary probability measure $\rho\in\mathcal{M}(\spaceHyp)$. 

Since PAC-Bayesian bounds guarantee generalization of distributions, they are often hard to read. Therefore, we first introduce the \emph{expected performance over a distribution $\rho\in\mathcal{M}(\spaceHyp)$}
\[
    \mathcal{R}(\rho) := \rho\left[\Exp\set{ \performFunc(\rvPar,\rvTraj) \condto \rvHyp} \right] %=
    %\intgr{\spaceHyp}{}{\intgr{\spacePar}{}{
    %    \intgr{\spaceTraj}{}{\performFunc(\realPar,\traj)}{\psi(\realHyp,\realPar,\df\traj)}}{\distPar(\df\realPar)}}{\distHyp(\df\realHyp)}\,.
\]
which averages/integrates the population performance (see \eqref{eq:def-expected-performance}) over algorithmic parameters distributed according to $\rho$. Moreover, for a (random) i.i.d. dataset $\rvPar_{[\sizeData]}:=(\rvPar_1,\ldots,\rvPar_{\sizeData})$, we define the \emph{empirical performance over a distribution $\rho\in\mathcal{M}(\spaceHyp)$} as 
\[
    \widehat{\mathcal{R}}_{\rvPar_{[\sizeData]}}(\rho) := 
    \frac{1}{N} \sum_{n=1}^N  \rho \left[\Exp\set{\performFunc(\rvPar_n,\rvTraj_n)\condto \rvHyp, \rvPar_n} \right] \,,
\]
which averages the expected performance of $\LOA_{\rvHyp}$ over problems $\rvPar_1,\ldots,\rvPar_{\sizeData}$ measured via the distribution $\rho$. Note that this is a measurable function with respect to the dataset $\rvPar_{[\sizeData]}$.

The last ingredient in a PAC-Bayesian bound is the deviation of the distribution $\rho$ from a so-called prior distribution, which in our case is $\distHyp$. 
\begin{Thm}[Generalization of Performance Functionals]\label{Thm:generalization_general_bounded_function}\label{thm:PAC-performance-func}
    Let $\map{\performFunc}{\spacePar\times\spaceTraj}{[0,\performFunc_{\max}]}$ be a bounded performance functional with $\performFunc_{\rm{max}} \in \R$. Then, for every $\lambda \in (0, \infty)$ and $\eps > 0$ it holds that: With probability at least $1-\eps$ over the draw of the dataset $\rvPar_{[\sizeData]}$, it holds simultaneously for all $\rho\ll \distHyp$ that
    \begin{align*}
         \mathcal{R}(\rho) \leq
        \widehat{\mathcal{R}}_{\rvPar_{[\sizeData]}}(\rho)
        + \frac{\divergence{\rm{KL}}{\rho}{\distHyp} + \frac{\lambda^2}{2N} \performFunc_{\rm{max}}^2 - \log(\eps)}{\lambda}  \,.
    \end{align*}
\end{Thm}
\begin{proof}
    The proof is deferred to Section~\ref{proof:Thm:generalization_general_bounded_function} in the appendix.
\end{proof}
The theorem shows that a distribution of algorithmic parameters with small empirical performance and small divergence from the prior distribution also enjoys good population performance. Consequently, optimization performance observed on a finite training set generalizes to previously unseen optimization problems drawn from the same distribution. Importantly, the result applies to arbitrary bounded performance functionals and is therefore independent of the specific notion of performance under consideration.
\begin{Rem}
    The boundedness assumption is convenient because it allows the use of Hoeffding-type concentration inequalities. Although PAC-Bayesian generalization bounds for unbounded losses are available, many performance functionals arising in optimization naturally admit finite bounds. Examples include finite stopping times, bounded objective errors, solution indicators, and finite-horizon convergence measures. 
\end{Rem}

\subsection{Guarantees for the Convergence Time}

As a special performance functional, we consider \emph{stopping time} \eqref{eq:def-stopping-time-func}. Stopping times are among the most natural performance measures in optimization, as they directly quantify the computational effort required to solve a problem. While no maximal iteration budget is required in the definition of the stopping time itself, the PAC-Bayesian bound in Theorem~\ref{thm:PAC-performance-func} assumes a bounded performance functional. Therefore, we consider a truncated stopping time (finite-horizon stopping time) within a maximal computational budget. This restriction is imposed solely to satisfy the boundedness assumption of Theorem~\ref{thm:PAC-performance-func} and is not a conceptual limitation of the framework.
\begin{Cor}[Generalization of Expected Stopping Times]\label{Cor:gen_conv_time}
   Let $\tau_{t_{\max}} := \min(\tau, t_{\max})$ be the stopping time truncated at a maximal computational budget $t_{\max}\in\N$. Using this stopping time as performance functional, we define the \emph{population stopping time over a distribution $\rho\in\mathcal{M}(\spaceHyp)$}
   \[ 
        \mathcal R_{\tau}(\rho) := \rho\left[ \Exp\set{ \tau_{t_{\max}}(\rvPar,\rvTraj) \condto \rvHyp } \right] 
    \]
   and the \emph{empirical stopping time over a distribution $\rho\in\mathcal{M}(\spaceHyp)$}
   \[ 
        \widehat{\mathcal R}_{\tau,\rvPar_{[\sizeData]}}(\rho) := \frac1N \sum_{n=1}^N \rho\left[ \Exp\set{ \tau_{t_{\max}}(\rvPar_n,\rvTraj_n) \condto \rvHyp,\rvPar_n } \right] \,.
    \] 
    Then, for every $\lambda \in (0, \infty)$ and $\eps > 0$ it holds that: With probability at least $1-\eps$ over the draw of the dataset $\rvPar_{[\sizeData]}$, it holds simultaneously for all $\rho\ll \distHyp$ that
    \begin{align*}
         \mathcal{R}_{\tau}(\rho) \leq \widehat{\mathcal{R}}_{\tau,\rvPar_{[\sizeData]}}(\rho) + \frac{\divergence{\rm{KL}}{\rho}{\distHyp} + \frac{\lambda^2}{2N} \performFunc_{\rm{max}}^2 - \log(\eps)}{\lambda}  \,.
    \end{align*}
\end{Cor}
\begin{proof}
    The result immediately follows from Theorem~\ref{thm:PAC-performance-func}.
\end{proof}
The corollary shows that stopping times generalize in exactly the same way as standard loss functions in statistical learning. More specifically, if a distribution of algorithmic parameters achieves a small empirical stopping time on a finite collection of training problems, then its expected stopping time on previously unseen optimization problems is controlled by the empirical stopping time and the complexity term measured through the Kullback--Leibler divergence from the prior distribution. Consequently, the number of iterations required to solve optimization problems can itself be treated as a learnable quantity and equipped with statistical generalization guarantees. This illustrates how the proposed framework extends PAC-Bayesian learning from prediction errors to optimization-specific notions of performance.

\subsection{Generalization of Contraction Factors}

Besides stopping times, optimization algorithms are often evaluated by how strongly they reduce a measure of suboptimality along the generated trajectory. While stopping times quantify the computational effort required to solve an optimization problem, convergence rates quantify the speed at which progress is achieved before reaching the solution criterion. Therefore, this subsection aims for generalization of the contraction factor $\contrate_{\tau}(\realPar,\traj)$ defined in \eqref{eq:def-contraction-factor}. The quantity $\contrate_{\tau}(\realPar,\traj)$ may be interpreted as the average per-iteration contraction factor required to reach the solution criterion. Smaller values correspond to faster convergence. The definition is inspired by classical asymptotic convergence rates while remaining meaningful in finite-time and stochastic settings.
\begin{Cor}[Generalization of Contraction Factors]
   Let $\contrate_{\contrate_{\max}} := \min(\contrate_{\tau},  \contrate_{\max})$
   be the contraction factor bounded by some $\contrate_{\max}>0$. Using this contraction factor as performance functional, we define the 
   \emph{population contraction factor over a distribution $\rho\in\mathcal{M}(\spaceHyp)$}
\[
    \mathcal R_\contrate(\rho) := \rho\left[ \Exp\set{ \contrate_{\contrate_{\max}}(\rvPar,\rvTraj) \condto \rvHyp } \right]
\]
and the \emph{empirical contraction factor over a distribution $\rho\in\mathcal{M}(\spaceHyp)$}
\[
    \widehat{\mathcal R}_{\contrate,\rvPar_{[\sizeData]}}(\rho) := \frac1N \sum_{n=1}^{N} \rho\left[ \Exp\set{ \contrate_{\contrate_{\max}}(\rvPar_n,\rvTraj_n) \condto \rvHyp,\rvPar_n } \right]\,.
\]

Then, for every $\lambda>0$ and $\eps>0$, with probability at least $1-\eps$ over the draw of the dataset $\rvPar_{[\sizeData]}$, it holds simultaneously for all $\rho\ll\distHyp$ that
\[
    \mathcal R_\contrate(\rho) \leq \widehat{\mathcal R}_{\contrate,\rvPar_{[\sizeData]}}(\rho) + \frac{ \divergence{\rm{KL}}{\rho}{\distHyp} + \frac{\lambda^2}{2N}\contrate_{\max}^2 - \log(\eps) }{\lambda} \,.
\]
\end{Cor}
\begin{proof}
    The result immediately follows from Theorem~\ref{thm:PAC-performance-func}.
\end{proof}
The corollary provides a PAC-Bayesian generalization guarantee for the average contraction achieved by an optimization algorithm before reaching a prescribed solution criterion. Consequently, contraction behavior observed on a finite collection of training problems generalizes to previously unseen optimization problems drawn from the same
distribution.

Unlike classical convergence analysis, which focuses on asymptotic rates and worst-case guarantees, the present result concerns finite-time statistical performance. In particular, it remains meaningful for stochastic optimization algorithms and illustrates once more that the proposed framework applies uniformly to optimization-specific notions of performance beyond prediction-style loss functions.

\subsection{Generalization of Trajectory Properties}
\label{subsec:PAC-success-probability}

The previous subsections considered stopping times and contraction factors as examples of optimization-specific performance functionals. More generally, one may be interested in whether a trajectory satisfies a prescribed property, which is achieved using the performance functional in \eqref{eq:def-convergence-probability-perf}. Examples include convergence to a stationary point, convergence within a prescribed tolerance, finite-time solution, or convergence with a certain contraction factor. While Theorem~\ref{thm:PAC-performance-func} could be applied directly to this bounded performance functional, the Bernoulli structure allows for sharper PAC-Bayesian bounds. The following result therefore provides a direct generalization guarantee for probabilities of trajectory properties.
\begin{Thm} \label{Thm:gen_property_trajectory}
Let $\setA\subset\spacePar\times\spaceTraj$ be measurable. Define the \emph{population probability over a distribution $\rho\in\mathcal{M}(\spaceHyp)$}
\[ 
    \mathcal P_{\setA}(\rho) := \rho\left[ \prob\set{ (\rvPar,\rvTraj)\in\setA \condto \rvHyp } \right] 
\] 
and the \emph{empirical probability over a distribution $\rho\in\mathcal{M}(\spaceHyp)$}
\[ 
    \widehat{\mathcal P}_{\setA,\rvPar_{[\sizeData]}}(\rho) := \frac1N \sum_{n=1}^{N} \rho\left[ \prob\set{ (\rvPar_n,\rvTraj_n)\in\setA \condto \rvHyp,\rvPar_n } \right]\,.
\] 
Then, for every $\lambda>0$ and $\eps>0$, with probability at least $1-\eps$ over the draw of the dataset $\rvPar_{[\sizeData]}$, it holds simultaneously for all $\rho\ll\distHyp$ that 
\[ 
    \mathcal P_{\setA}(\rho) \le \Phi_{\lambda/N}^{-1} \left( \widehat{\mathcal P}_{\setA,\rvPar_{[\sizeData]}}(\rho) + \frac{  \divergence{\rm{KL}}{\rho}{\distHyp} + \log(1/\eps) } {\lambda} \right)
\]
where
\[
    \Phi_a(p) := - \frac{1}{a} \log \left( 1 - \left[ 1 - \exp(-a) \right] p \right) \,.
\]
\end{Thm}
\begin{proof}
The proof is deferred to Appendix~\ref{Proof:Thm:gen_property_trajectory}.
\end{proof}
The theorem provides a PAC-Bayesian generalization guarantee for the probability that a trajectory satisfies a prescribed property. Consequently, convergence events, solution criteria, and other trajectory-level properties can be studied within the same framework as stopping times and contraction factors.

A concrete application of this viewpoint was developed in \cite{SO25}, where PAC-Bayesian bounds were used to study the generalization of convergence for learned optimization algorithms in non-smooth and nonconvex optimization. There, the event $\setA$ encoded convergence of the entire optimization trajectory to a critical point characterized through the limiting subdifferential achieved through the usage of the Kurdyka-{\L}ojasiewicz framework.

\section{Numerical Experiments}

The purpose of the numerical experiments is not to introduce a new state-of-the-art optimization algorithm, but to illustrate the statistical-learning perspective on optimization algorithms developed throughout this work. In particular, we investigate how optimization performance should be viewed as a random quantity induced by a distribution of optimization problems rather than as a property of a single problem instance. 

We focus on stopping times and contraction factors as a statistical performance measure. For each experiment, a learned optimization algorithm is evaluated on previously unseen test problems sampled from the underlying problem distribution. The reported histograms therefore visualize the induced distribution of stopping times and contraction factors and provide an empirical description of the optimization performance across the corresponding class of optimization problems. This viewpoint reveals several phenomena that are invisible from a purely worst-case perspective. In particular, the experiments expose the variability of optimization performance across the problem distribution and highlight differences between various statistical summaries such as means, medians, quantiles, and tail behavior. Consequently, they illustrate the motivation behind the performance measures introduced in Section~\ref{sec:perf-measures}. For several problem classes, one can observe that a small number of difficult instances substantially affects the average stopping time, while the typical performance remains considerably better. 

A second objective of the experiments is to demonstrate the breadth of the proposed framework. We therefore consider optimization problems of increasing complexity, ranging from strongly convex quadratic objectives to large-scale inverse problems, non-smooth optimization, nonconvex neural-network training, and stochastic empirical risk minimization. Together, these examples illustrate that the framework is largely independent of the specific structure of the underlying optimization problem and naturally accommodates convex and nonconvex, smooth and non-smooth, as well as deterministic and stochastic optimization algorithms. 

Finally, the experiments assess the practical relevance of the PAC-Bayesian guarantees developed in Section~\ref{sec:PAC-Bayesian-guarantees}. Alongside the observed stopping-time distributions, we report the corresponding PAC-Bayesian estimates for the expected stopping time and the contraction-factor performance. This allows us to compare the predicted guarantees with the empirically observed behavior across a variety of optimization settings. We consider the following five problem classes: strongly convex quadratic optimization, image restoration, sparse recovery via the LASSO problem, deterministic training of a neural network, and stochastic empirical risk minimization through mini-batch training of the same neural network. 

\paragraph{Experimental setup.} The training procedure largely follows \citet{SFO25}. Briefly, the learnable optimization algorithm is first trained on optimization problems sampled from the training distribution. The resulting parameter vector serves as a reference point for constructing a discrete prior distribution on the algorithmic parameter space. Subsequently, a closed-form PAC-Bayesian optimization step is performed, yielding a posterior distribution over algorithmic parameters. For simplicity, the final algorithm is chosen as the maximizer of the posterior probability mass. Complete architectural specifications, hyperparameter choices, and implementation details are available in \href{https://github.com/MichiSucker/Markovian-Model-for-L2O}{our public code repository}.

The only modification with respect to \citet{SFO25} concerns the training objective. Instead of optimizing a finite-horizon contraction objective, we employ a one-step contraction objective that is active only while the current iterate does not satisfy the solution criterion $\solutionCrit$:
$$
    \ell_{\mathrm{train}}(\realHyp, \realPar, \iter{\vx}{t}) = \mathds{1}\{ \ell(\iter{\vx}{t}, \realPar) > 0\} \frac{\ell(\iter{\vx}{t+1}, \realPar)}{\ell(\iter{\vx}{t}, \realPar)} \cdot \mathds{1}_{\solutionCrit^c}(\iter{\vx}{t}, \realPar)\,.
$$

\subsection{Distribution-Adaptive Optimization on Quadratic Problems}

We begin with a class of strongly convex quadratic optimization problems. This experiment serves as a controlled setting in which the statistical framework developed in this work can be illustrated on a well-understood family of optimization problems. Owing to their simple structure, quadratic problems allow the observed statistical behavior to be attributed primarily to the distribution of problem instances rather than model or data complexity.

Each problem is of the form
\[
    \min_{x\in\R^d} \frac12\|Ax-b\|^2,
\]
with optimization variable $x\in\R^d$, where $d=200$. A distribution of optimization problems is generated by sampling strong-convexity $m$ and smoothness constants $L$ from prescribed intervals $[m_-,m_+]$ and $[L_-,L_+]$, respectively, and constructing a diagonal matrix $A=\diag(a_{ii})_{i=1,\ldots,d}$ with entries $a_{ii}=\sqrt{m}+i\frac{\sqrt{L}-\sqrt{m}}d$, $i=1, \ldots,d$. The right-hand side $b$ is sampled independently. The solution criterion is defined by
\[
    \ell(x,\realPar)<10^{-8}
    \quad\text{or}\quad
    \|\nabla\ell(x,\realPar)\|<10^{-6}.
\]
The learned algorithm $\LOA$ performs an update of the form $\iter{x}{t+1} = \iter{x}{t} + \iter{\beta}{t} \cdot \iter{d}{t}$, where $\iter{\beta}{t}$ and $\iter{d}{t}$ are predicted by separate blocks of a neural network. For more details on the architecture we refer to Appendix~\ref{Appendix:architecture_quad}. 

\begin{figure}[t!]
    \centering
    \tikz{
      \node (A) {\includegraphics[width=0.95\textwidth]{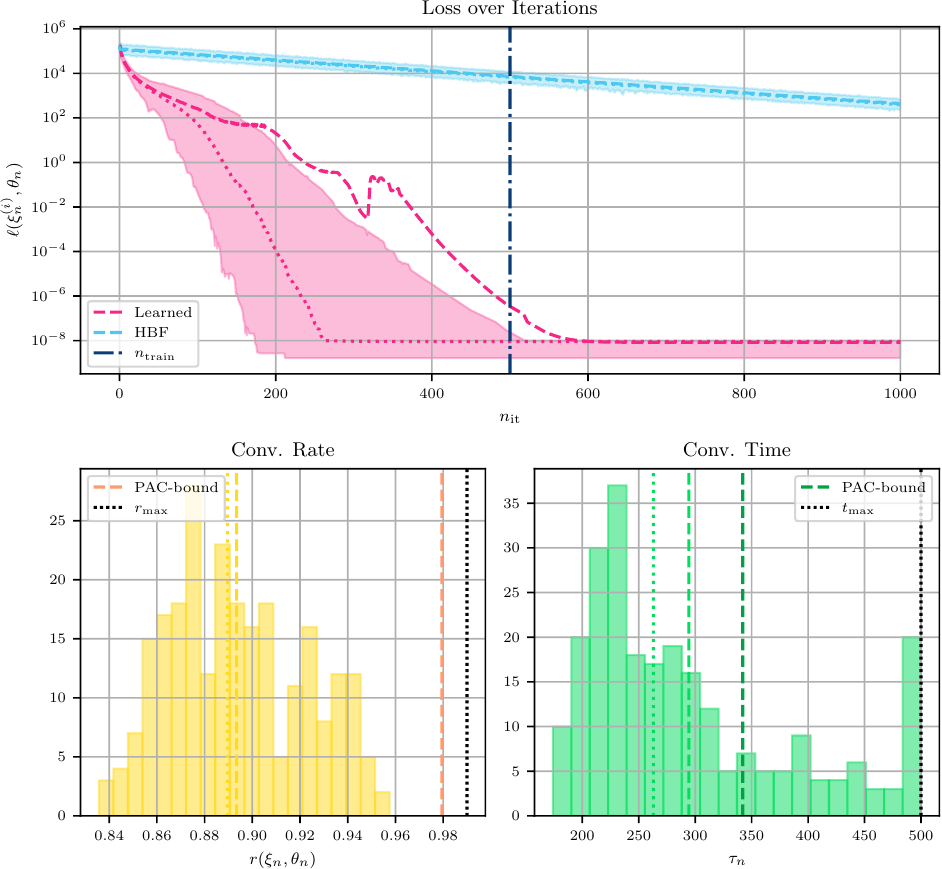}};

      \draw (A.north west) ++ (0,-2.5) coordinate (A1);
      \draw (A1) ++ (0.6,-1.4) coordinate (A2);
      \draw[white,fill=white] (A1) rectangle (A2);
      \draw (A1) node [below left, rotate=90,font=\small] {$\ell(\iter{\vx}{t}, \realPar)$};

      \draw (A) ++ (+0.3,0.6) coordinate (B1);
      \draw (B1) ++ (0.6,-0.6) coordinate (B2);
      \draw[white,fill=white] (B1) rectangle (B2);
      \draw (B1) node [below right,font=\small] {$t$};

      \draw (A) ++ (-4.5,+0.15) coordinate (C1);
      \draw (C1) ++ (3,-0.6) coordinate (C2);
      \draw[white, fill=white] (C1) rectangle (C2);
      \draw (C1) node [below right,font=\small] {Contraction factor};

      \draw (A) ++ (+2,+0.15) coordinate (D1);
      \draw (D1) ++ (4,-0.6) coordinate (D2);
      \draw[white, fill=white] (D1) rectangle (D2);
      \draw (D1) node [below right,font=\small] {Truncated stopping time};

      \draw (A) ++ (-3.5,-6.3) coordinate (E1);
      \draw (E1) ++ (1.3,-0.6) coordinate (E2);
      \draw[white,fill=white] (E1) rectangle (E2);
      \draw (E1) node [below right,font=\small] {$\contrate_{\contrate_{\max}}$};

      \draw (A) ++ (+3.5,-6.3) coordinate (F1);
      \draw (F1) ++ (1.3,-0.6) coordinate (F2);
      \draw[white,fill=white] (F1) rectangle (F2);
      \draw (F1) node [below right,font=\small] {$\tau_{t_{\max}}$};

      \draw (A) ++ (-5.3,-0.9) coordinate (F1);
      \draw (F1) ++ (1.3,-0.35) coordinate (F2);
      \draw[white,fill=white] (F1) rectangle (F2);
      \draw (F1) node [below right,font=\scriptsize] {$\contrate_{\max}$};

    }
    \caption{
    Strongly convex quadratic problems. Top: evolution of the objective value over previously unseen test problems. The shaded region contains $95\%$ of the observed trajectories, while dashed and dotted curves indicate the empirical mean and median, respectively. The learned optimizer exhibits substantially faster convergence than the Heavy-ball method (HBF) \cite{Polyak_1964}. Bottom left: empirical distribution of contraction factors together with the corresponding PAC-Bayesian estimate. Bottom right: empirical distribution of stopping times together with the corresponding PAC-Bayesian estimate. The discrepancy between mean and median stopping times illustrates that optimization performance is inherently statistical and cannot be fully characterized by a single problem instance.}
    \label{fig:quadratics}
\end{figure}

Figure~\ref{fig:quadratics} illustrates several statistical summaries of optimization performance on previously unseen test problems. The experiment highlights that optimization performance is inherently a statistical quantity. While most problem instances satisfy the solution criterion well before the prescribed budget $t_{\max}=500$, there remains substantial variability across instances. In particular, the mean stopping time is noticeably larger than the median, indicating that a relatively small number of difficult problems exert a disproportionate influence on the expected performance. Consequently, a single summary statistic does not fully characterize the behavior of the algorithm over the problem distribution.

The lower-right plot shows that the expected stopping time is below $300$ iterations and that the corresponding PAC-Bayesian estimate provides a reasonably accurate upper bound. Likewise, the lower-left plot indicates that the optimizer achieves an average contraction factor of approximately $0.9$ per iteration. Although the associated PAC-Bayesian estimate is less tight, it nevertheless remains non-vacuous and captures the correct magnitude of the expected performance.

Overall, the experiment illustrates the central thesis of the present work: optimization performance should be studied statistically over distributions of optimization problems rather than solely through individual instances or worst-case guarantees.

\subsection{Large-Scale Image Restoration}

To demonstrate that the proposed framework extends beyond synthetic quadratic examples, we consider a distribution of image restoration problems consisting of image deblurring and denoising tasks with smoothed total-variation regularization. In contrast to the quadratic setting, this experiment involves a large-scale inverse problem with real-world image data and a high-dimensional optimization space. Each problem takes the form
\[
    \min_{x\in\R^d} \frac12\|Ax-b\|^2 + \lambda \sum_{i,j} \sqrt{ (D_hx)_{i,j}^2 + (D_wx)_{i,j}^2 + \varepsilon^2 }\,,
\]
where $A$ models image blur and $D_h$, $D_w$ denote discrete image derivative operators. We use grayscale images of size $200\times150$, resulting in an optimization space of dimension $d=30000$. The problem parameters consist of the observed image and the regularization parameter $\lambda$, which is sampled uniformly from $[0.05,0.5]$. The solution criterion is given by
\[
    \|\nabla\ell(x,\realPar)\|<10^{-4}\,.
\]
The learned algorithm $\LOA$ performs an update of the form $\iter{x}{t+1} = \iter{x}{t} + \frac{\norm{\nabla_x \ell(\iter{x}{t}, \realPar) }{}}{L} \iter{d}{t}_1 + \norm{\iter{x}{t} - \iter{x}{t-1}}{} \iter{d}{t}_2 - \frac{1}{L} \nabla_x \ell(\iter{x}{t}, \realPar)$, where the directions $\iter{d}{t}_1$ and $\iter{d}{t}_2$ are predicted by a neural network, and $L$ is given by the largest eigenvalue of $A^T A + \frac{\lambda}{\varepsilon} D^T D$, where $D \in \R^{2d \times d}$ is given by \say{stacking} $D_h$ and $D_w$. For more details on the architecture we refer to Appendix~\ref{Appendix:architecture_img}. 

\begin{figure}[t!]
    \centering
    
        \tikz{
      \node (A) {\includegraphics[width=0.95\textwidth]{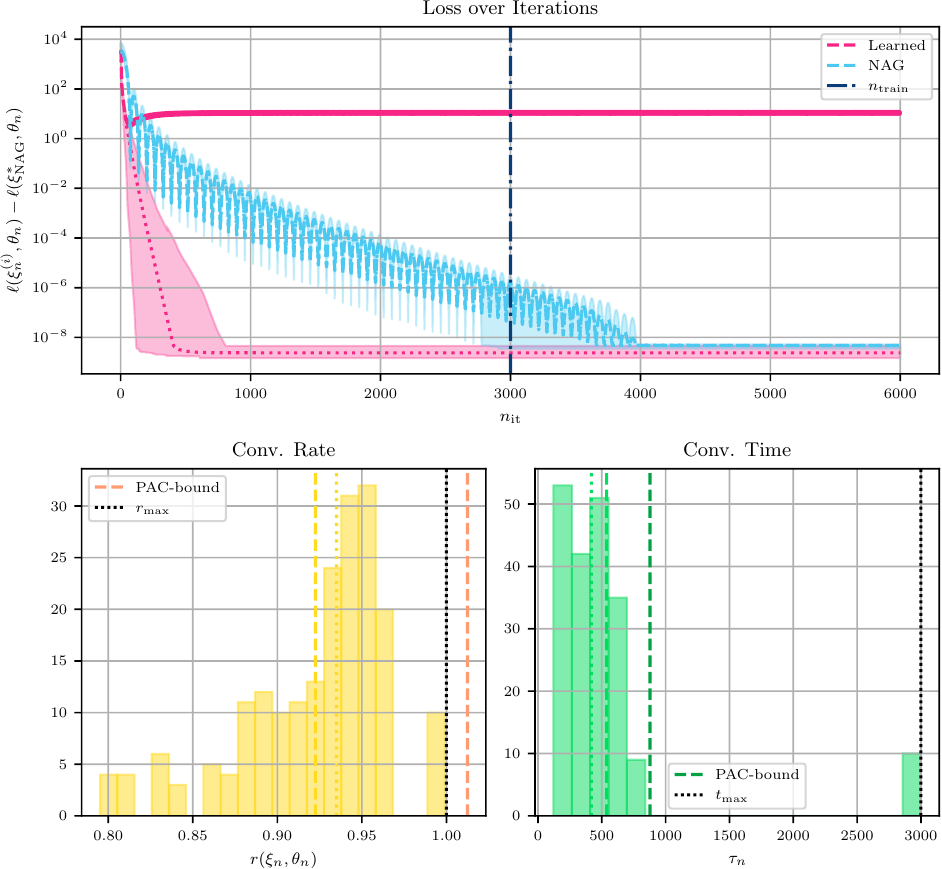}};

      \draw (A.north west) ++ (0,-1.5) coordinate (A1);
      \draw (A1) ++ (0.6,-4.4) coordinate (A2);
      \draw[white,fill=white] (A1) rectangle (A2);
      \draw (A1) node [below left, rotate=90,font=\small] {$\ell(\iter{\vx}{t}, \realPar)- \ell(\vx^*_{\text{NAG}},\realPar)$};

      \draw (A) ++ (+0.3,0.5) coordinate (B1);
      \draw (B1) ++ (0.6,-0.6) coordinate (B2);
      \draw[white,fill=white] (B1) rectangle (B2);
      \draw (B1) node [below right,font=\small] {$t$};

      \draw (A) ++ (-4.5,+0.15) coordinate (C1);
      \draw (C1) ++ (3,-0.6) coordinate (C2);
      \draw[white, fill=white] (C1) rectangle (C2);
      \draw (C1) node [below right,font=\small] {Contraction factor};

      \draw (A) ++ (+2,+0.15) coordinate (D1);
      \draw (D1) ++ (4,-0.6) coordinate (D2);
      \draw[white, fill=white] (D1) rectangle (D2);
      \draw (D1) node [below right,font=\small] {Truncated stopping time};

      \draw (A) ++ (-3.5,-6.3) coordinate (E1);
      \draw (E1) ++ (1.3,-0.6) coordinate (E2);
      \draw[white,fill=white] (E1) rectangle (E2);
      \draw (E1) node [below right,font=\small] {$\contrate_{\contrate_{\max}}$};

      \draw (A) ++ (+3.5,-6.3) coordinate (F1);
      \draw (F1) ++ (1.3,-0.6) coordinate (F2);
      \draw[white,fill=white] (F1) rectangle (F2);
      \draw (F1) node [below right,font=\small] {$\tau_{t_{\max}}$};

      \draw (A) ++ (-5.3,-0.9) coordinate (F1);
      \draw (F1) ++ (1.3,-0.35) coordinate (F2);
      \draw[white,fill=white] (F1) rectangle (F2);
      \draw (F1) node [below right,font=\scriptsize] {$\contrate_{\max}$};
    }
    \caption{
    Image restoration problems with smoothed total-variation regularization. Top: evolution of the objective value over previously unseen test problems. The shaded region contains $95\%$ of the observed trajectories, while dashed and dotted curves indicate the empirical mean and median, respectively. The pronounced spread across trajectories reflects substantial variability in optimization difficulty. As a comparison, the evolution for Nesterov's Accelerated Gradient method (NAG) \cite{Nesterov_1983} shown. Bottom left: empirical distribution of contraction factors together with the corresponding PAC-Bayesian estimate. In this experiment, the estimate becomes vacuous. Bottom right: empirical distribution of stopping times together with the corresponding PAC-Bayesian estimate. The estimate remains reasonably tight despite the heavier-tailed stopping-time distribution.}
    \label{fig:image_processing}
\end{figure}

Figure~\ref{fig:image_processing} shows the stopping-time and contraction-factor distributions on previously unseen image restoration problems. Compared to the quadratic setting, the stopping-time distribution exhibits a noticeably heavier tail. While most problems reach the solution criterion within a few hundred iterations, a small number of instances fail to do so within the prescribed computational budget. As a consequence, the gap between the mean and the typical performance is significantly larger than in the quadratic experiment.

The stopping-time PAC estimate remains reasonably accurate and captures the expected computational effort well. In contrast, the corresponding contraction-factor estimate becomes vacuous. This illustrates that different statistical performance measures may exhibit substantially different generalization behavior, even on the same class of optimization problems.

More broadly, the experiment demonstrates that the proposed framework naturally extends from low-dimensional synthetic benchmarks to large-scale inverse problems involving tens of thousands of optimization variables. The resulting statistical phenomena persist across both settings, suggesting that they are intrinsic features of optimization performance rather than artifacts of a particular problem class.

\subsection{Non-smooth Sparse Recovery}

To illustrate that the framework naturally extends beyond smooth optimization, we consider a distribution of LASSO problems \citep{Tibshirani_1996} of the form
\[
    \min_{x\in\R^d} \frac12\|Ax-b\|_2^2 + \lambda\|x\|_1 \,,
\]
where $A\in\R^{p\times d}$ with $p=35$ and $d=70$. The problem parameters are given by the right-hand side $b$ and the regularization parameter $\lambda$. The matrix $A$ is fixed throughout the experiment, while the regularization parameter is sampled uniformly from $[5,10]$ and the right-hand side is sampled from a multivariate normal distribution.

Since the objective is non-smooth, we split it into two terms, $h$ is the quadratic and $g$ the second summand. The solution criterion is defined via the proximal-gradient residual:
\[
    \Bigl\| x- \operatorname{prox}_{\beta g} \bigl( x-\beta\nabla h(x) \bigr) \Bigr\| < 10^{-6}\,,
\]
with $\beta=1/L$ and the largest eigenvalue of $A^\top A$ being $L$. 
The learned algorithm $\LOA$ performs an update of the form 
\[
    \iter{x}{t+1} = \mathrm{prox}_{\beta g}\left(\iter{x}{t} + \frac{1}{L} \left(\iter{d}{t}_1 - \nabla h(\iter{x}{t}) + \norm{\iter{x}{t} - \iter{x}{t-1}}{} \cdot \iter{d}{t}_2 \right)\right)\,,
\]
where $\iter{d}{t}_1$ and $\iter{d}{t}_2$ are predicted by a neural network. For more details on the architecture we refer to Appendix~\ref{Appendix:architecture_lasso}. 

\begin{figure}[t!]
    \centering
    
        \tikz{
      \node (A) {\includegraphics[width=0.95\textwidth]{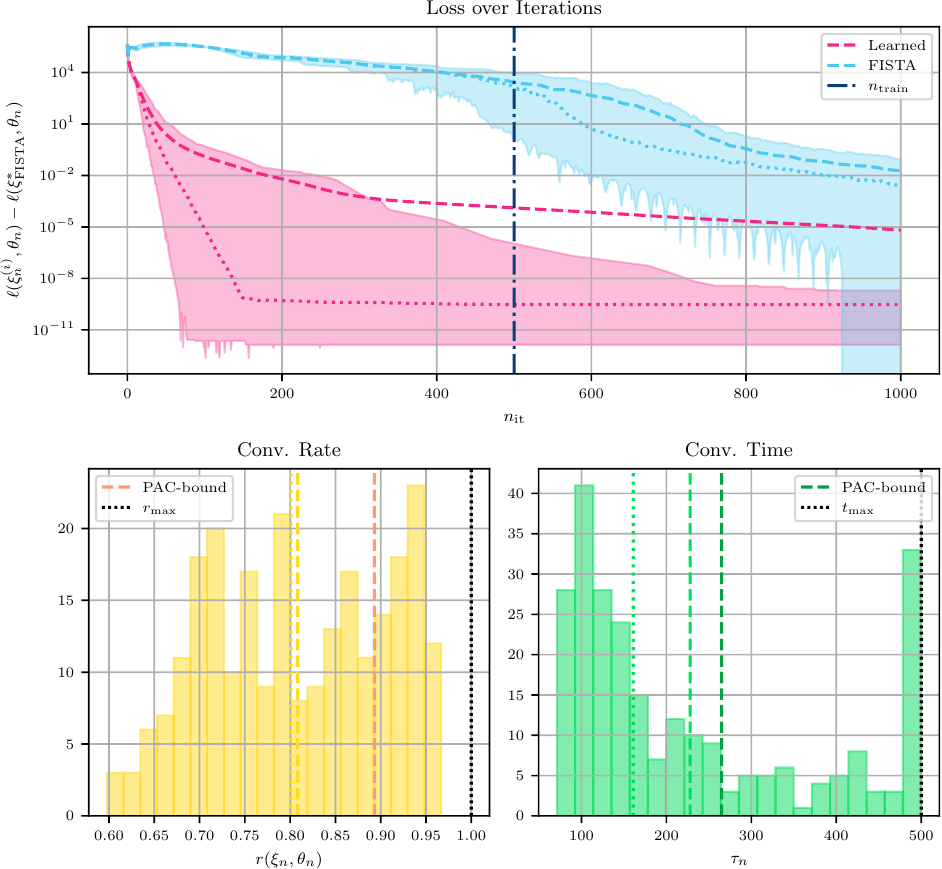}};

      \draw (A.north west) ++ (0,-1.5) coordinate (A1);
      \draw (A1) ++ (0.6,-4.4) coordinate (A2);
      \draw[white,fill=white] (A1) rectangle (A2);
      \draw (A1) node [below left, rotate=90,font=\small] {$\ell(\iter{\vx}{t}, \realPar)- \ell(\vx^*_{\text{FISTA}},\realPar)$};

      \draw (A) ++ (+0.3,0.5) coordinate (B1);
      \draw (B1) ++ (0.6,-0.6) coordinate (B2);
      \draw[white,fill=white] (B1) rectangle (B2);
      \draw (B1) node [below right,font=\small] {$t$};

      \draw (A) ++ (-4.5,+0.15) coordinate (C1);
      \draw (C1) ++ (3,-0.6) coordinate (C2);
      \draw[white, fill=white] (C1) rectangle (C2);
      \draw (C1) node [below right,font=\small] {Contraction factor};

      \draw (A) ++ (+2,+0.15) coordinate (D1);
      \draw (D1) ++ (4,-0.6) coordinate (D2);
      \draw[white, fill=white] (D1) rectangle (D2);
      \draw (D1) node [below right,font=\small] {Truncated stopping time};

      \draw (A) ++ (-3.5,-6.3) coordinate (E1);
      \draw (E1) ++ (1.3,-0.6) coordinate (E2);
      \draw[white,fill=white] (E1) rectangle (E2);
      \draw (E1) node [below right,font=\small] {$\contrate_{\contrate_{\max}}$};

      \draw (A) ++ (+3.5,-6.3) coordinate (F1);
      \draw (F1) ++ (1.3,-0.6) coordinate (F2);
      \draw[white,fill=white] (F1) rectangle (F2);
      \draw (F1) node [below right,font=\small] {$\tau_{t_{\max}}$};

      \draw (A) ++ (-5.3,-0.9) coordinate (F1);
      \draw (F1) ++ (1.3,-0.35) coordinate (F2);
      \draw[white,fill=white] (F1) rectangle (F2);
      \draw (F1) node [below right,font=\scriptsize] {$\contrate_{\max}$};
    }
    
    \caption{
    LASSO problems with $\ell_1$ regularization. Top: evolution of the objective value over previously unseen test problems. The shaded region contains $95\%$ of the observed trajectories, while dashed and dotted curves indicate the empirical mean and median, respectively. The discrepancy between mean and median again highlights the statistical variability of optimization performance across problem instances. We use FISTA \cite{Beck_Teboulle_2009} as a baseline. left: empirical distribution of contraction factors together with the corresponding PAC-Bayesian estimate. Bottom right: empirical distribution of stopping times together with the corresponding PAC-Bayesian estimate. Both estimates remain reasonably accurate, illustrating the applicability of the proposed framework to non-smooth optimization problems.}
    \label{fig:lasso}
\end{figure}

Figure~\ref{fig:lasso} shows the resulting stopping-time and contraction-factor distributions on previously unseen test problems.  As in the previous experiments, the stopping-time distribution exhibits considerable variability across problem instances. In particular, the mean stopping time is noticeably larger than the median, indicating that a relatively small number of difficult instances contributes disproportionately to the expected performance.

The stopping-time distribution reveals that most optimization problems satisfy the solution criterion within approximately $200$ iterations. Moreover, both the stopping-time and contraction-factor PAC-Bayesian estimates remain reasonably accurate. 

More importantly, the experiment illustrates that the proposed framework is not tied to smooth optimization problems or gradient-based optimality measures. By formulating performance through solution criteria and performance functionals, the same statistical framework applies equally to non-smooth optimization settings.

\subsection{Neural-Network Training}
\label{Subsec:training_nn}

We now move beyond convex optimization and consider the problem of training a neural network on a regression task, a non-convex optimization problem. A distribution of regression problems is generated by sampling random polynomials of degree five and observing noisy measurements $y=g(x)+\varepsilon$ with $\varepsilon\sim\mathcal N(0,1)$. We construct polynomials $g_i$, $i=1,...,N$, of degree $d=5$ by sampling points $\{x_{i,j}\}_{j=1}^K$ (here: $K = 50$) uniformly in $[-2, 2]$ and the coefficients $(c_{i, 0}, ..., c_{i, 5})$ of $g_i$ uniformly in $[-5, 5]$. For every function $g_i: \R \to \R$ the neural network is trained on the data set $\{X_i, Y_i\}$ with $X_i = (x_{i,1}, ..., x_{i,K}) \in \R^K$ and $Y_i = (y_{i,1}, ..., y_{i,K}) \in \R^K$. Hence, the problem parameter $\realPar$ is identified as one of these data sets $(X_i,Y_i)$. The network $\mathtt N$ is a two-layer ReLU network with $\beta\in\R^{351}$ parameters, which is the optimization variable in our framework. 

The objective function is the mean squared prediction error
\[
\ell(\beta,\realPar) = \frac1K \sum_{j=1}^{K} \bigl( \mathtt N(\beta,x_j)-y_j \bigr)^2 \,,
\]
and the solution criterion is given by
\[
\|\nabla \ell(\beta,\realPar)\|<0.75, \qquad \ell(\beta,\realPar)<0.75 \,.
\]
The learned algorithm simply performs the update $\iter{x}{t+1} = \iter{x}{t} + \iter{d}{t}$, where $\iter{d}{t}$ is predicted by a neural network. For more details on the architecture, we refer to Appendix~\ref{Appendix:architecture_nn}. 

\begin{figure}[p]
    \centering
        \tikz{
      \node (A) {\includegraphics[width=0.95\textwidth]{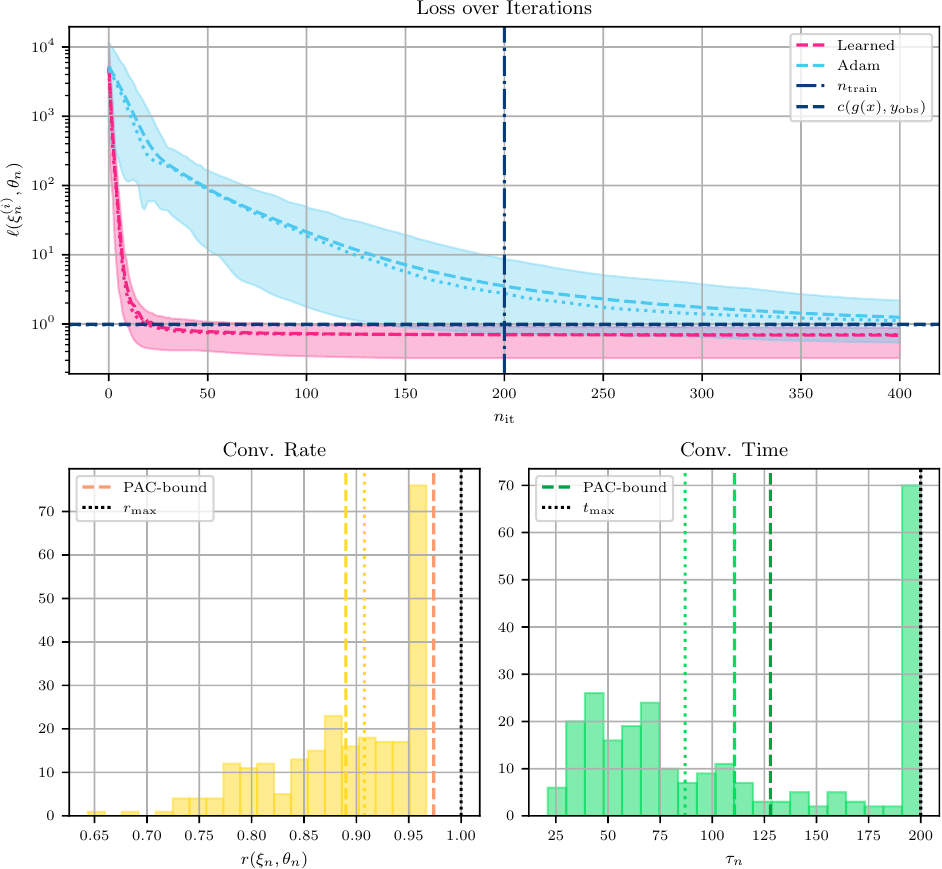}};

      \draw (A.north west) ++ (0,-2.5) coordinate (A1);
      \draw (A1) ++ (0.6,-2.4) coordinate (A2);
      \draw[white,fill=white] (A1) rectangle (A2);
      \draw (A1) node [below left, rotate=90,font=\small] {$\ell(\iter{\vx}{t}, \realPar)$};

      \draw (A) ++ (+0.3,0.5) coordinate (B1);
      \draw (B1) ++ (0.6,-0.6) coordinate (B2);
      \draw[white,fill=white] (B1) rectangle (B2);
      \draw (B1) node [below right,font=\small] {$t$};

      \draw (A) ++ (-4.5,+0.15) coordinate (C1);
      \draw (C1) ++ (3,-0.6) coordinate (C2);
      \draw[white, fill=white] (C1) rectangle (C2);
      \draw (C1) node [below right,font=\small] {Contraction factor};

      \draw (A) ++ (+2,+0.15) coordinate (D1);
      \draw (D1) ++ (4,-0.6) coordinate (D2);
      \draw[white, fill=white] (D1) rectangle (D2);
      \draw (D1) node [below right,font=\small] {Truncated stopping time};

      \draw (A) ++ (-3.55,-6.3) coordinate (E1);
      \draw (E1) ++ (1.3,-0.6) coordinate (E2);
      \draw[white,fill=white] (E1) rectangle (E2);
      \draw (E1) node [below right,font=\small] {$\contrate_{\contrate_{\max}}$};

      \draw (A) ++ (+3.5,-6.3) coordinate (F1);
      \draw (F1) ++ (1.3,-0.6) coordinate (F2);
      \draw[white,fill=white] (F1) rectangle (F2);
      \draw (F1) node [below right,font=\small] {$\tau_{t_{\max}}$};

      \draw (A) ++ (-5.3,-0.9) coordinate (F1);
      \draw (F1) ++ (1.3,-0.35) coordinate (F2);
      \draw[white,fill=white] (F1) rectangle (F2);
      \draw (F1) node [below right,font=\scriptsize] {$\contrate_{\max}$};
    }
   
    \caption{Neural-network training problems. Top: evolution of the objective value over previously unseen test tasks. The shaded region contains $95\%$ of the observed trajectories, while dashed and dotted curves indicate the empirical mean and median, respectively. The pronounced spread across trajectories reflects the substantial variability of optimization performance in the nonconvex setting. 
    As baseline we use a full batch version of Adam \citep{Kingma_Ba_2015} as it is implemented in PyTorch \citep{PyTorch_2019}, which is a widely used optimization algorithm for training neural networks. 
    Bottom left: empirical distribution of contraction factors together with the corresponding PAC-Bayesian estimate. Compared to the convex experiments, the distribution exhibits considerably larger variability. Bottom right: empirical distribution of stopping times together with the corresponding PAC-Bayesian estimate. A substantial fraction of problem instances does not satisfy the solution criterion within the prescribed computational budget, resulting in a heavy-tailed stopping-time distribution.}
    \label{fig:nn_training}
\end{figure}

Figure~\ref{fig:nn_training} displays several statistical summaries of optimization performance on previously unseen neural-network training tasks. Compared to the convex experiments, the variability across problem instances is substantially larger. While a large fraction of tasks satisfies the solution criterion quickly, there also exists a significant subset of instances for which the criterion is not reached within the prescribed computational budget.

This behavior is reflected both in the stopping-time distribution and in the contraction factors. The latter show a considerably larger spread than in the previous experiments, illustrating the increased difficulty and heterogeneity of the underlying optimization problems. Nevertheless, the PAC-Bayesian estimates remain reasonably informative for both performance measures.

More broadly, the experiment demonstrates that the proposed framework is not restricted to convex optimization settings. The same notions of solution criteria, performance functionals, statistical performance measures, and PAC-Bayesian estimates remain applicable in a genuinely nonconvex learning scenario.

\subsection{Stochastic Empirical Risk Minimization}

Finally, we consider the same neural-network training problem as in Subsection~\ref{Subsec:training_nn}, but replace full-batch gradients by mini-batch gradients of size $m=5$. The resulting optimization problem is stochastic, nonconvex, and non-smooth. Consequently, the optimization trajectory becomes a genuinely random object, providing a natural setting for the probabilistic framework developed in this work. 

Since full gradients are no longer available, the solution criterion is defined through the objective value, 
\[ 
    \ell(\beta,\realPar)<0.75 \,.
\] 
As learned algorithm, we employ a preconditioned variant of Adam with additional learnable parameters. Further architectural details are provided in Appendix~\ref{Appendix:architecture_stochastic_emp_risk}. 

\begin{figure}[t!]
    \centering
    
        \tikz{
      \node (A) {\includegraphics[width=0.95\textwidth]{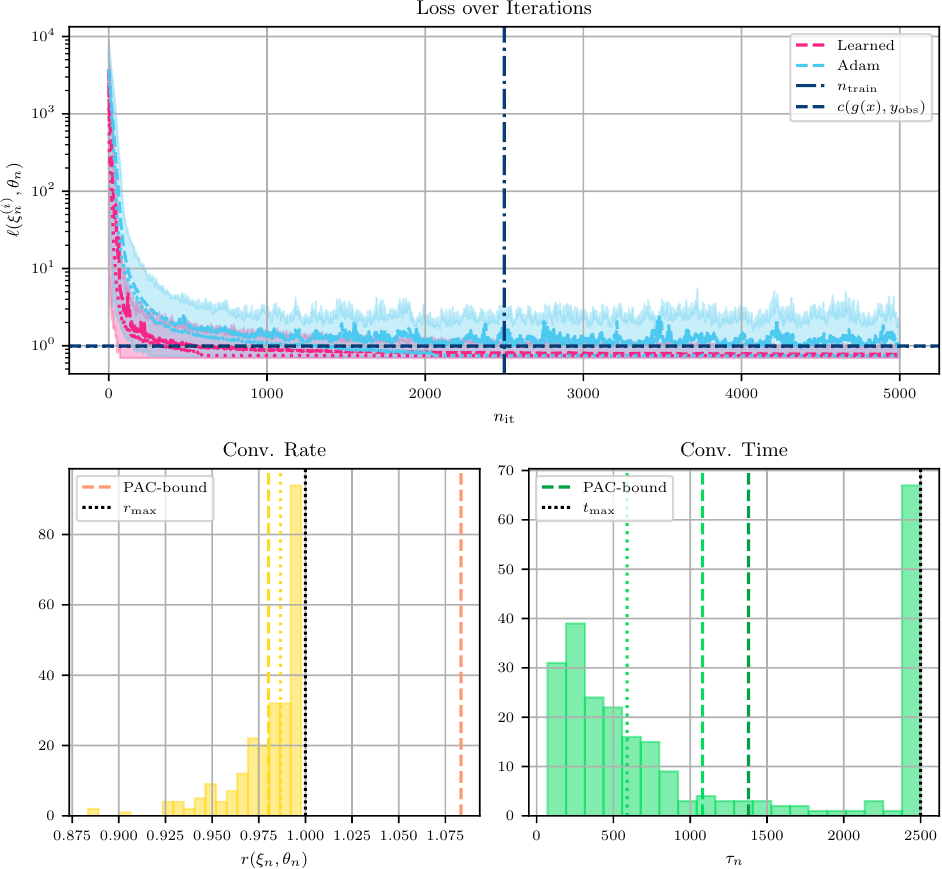}};

      \draw (A.north west) ++ (0,-2.5) coordinate (A1);
      \draw (A1) ++ (0.6,-2.4) coordinate (A2);
      \draw[white,fill=white] (A1) rectangle (A2);
      \draw (A1) node [below left, rotate=90,font=\small] {$\ell(\iter{\vx}{t}, \realPar)$};

      \draw (A) ++ (+0.3,0.5) coordinate (B1);
      \draw (B1) ++ (0.6,-0.6) coordinate (B2);
      \draw[white,fill=white] (B1) rectangle (B2);
      \draw (B1) node [below right,font=\small] {$t$};

      \draw (A) ++ (-4.5,+0.15) coordinate (C1);
      \draw (C1) ++ (3,-0.6) coordinate (C2);
      \draw[white, fill=white] (C1) rectangle (C2);
      \draw (C1) node [below right,font=\small] {Contraction factor};

      \draw (A) ++ (+2,+0.15) coordinate (D1);
      \draw (D1) ++ (4,-0.6) coordinate (D2);
      \draw[white, fill=white] (D1) rectangle (D2);
      \draw (D1) node [below right,font=\small] {Truncated stopping time};

      \draw (A) ++ (-3.55,-6.3) coordinate (E1);
      \draw (E1) ++ (1.3,-0.6) coordinate (E2);
      \draw[white,fill=white] (E1) rectangle (E2);
      \draw (E1) node [below right,font=\small] {$\contrate_{\contrate_{\max}}$};

      \draw (A) ++ (+3.5,-6.3) coordinate (F1);
      \draw (F1) ++ (1.3,-0.6) coordinate (F2);
      \draw[white,fill=white] (F1) rectangle (F2);
      \draw (F1) node [below right,font=\small] {$\tau_{t_{\max}}$};

      \draw (A) ++ (-5.3,-0.9) coordinate (F1);
      \draw (F1) ++ (1.3,-0.35) coordinate (F2);
      \draw[white,fill=white] (F1) rectangle (F2);
      \draw (F1) node [below right,font=\scriptsize] {$\contrate_{\max}$};
    }
  
    \caption{
    Stochastic empirical risk minimization. Top: evolution of the empirical risk over previously unseen test problems. The shaded region contains $95\%$ of the observed trajectories, while dashed and dotted curves indicate the empirical mean and median, respectively. The increased spread across trajectories reflects the combined variability induced by the problem distribution and the stochastic optimization algorithm. Bottom left: empirical distribution of contraction factors together with the corresponding PAC-Bayesian estimate. As in the image-restoration experiment, the estimate becomes vacuous. Bottom right: empirical distribution of stopping times together with the corresponding PAC-Bayesian estimate. Despite the increased variability, the estimate remains reasonably informative and captures the correct scale of the expected computational effort.}
    \label{fig:stochastic_emp_risk}
\end{figure}

Figure~\ref{fig:stochastic_emp_risk} displays several statistical summaries of optimization performance on previously unseen stochastic optimization problems. Compared to the deterministic neural-network training experiment, the stopping-time distribution becomes broader and exhibits substantially heavier tails. In particular, the difference between mean and median stopping times indicates that a relatively small number of difficult optimization tasks dominates the expected computational effort. 

The stopping-time PAC-Bayesian estimate remains informative and provides a reasonable upper bound on the expected computational budget. In contrast, the corresponding contraction-factor estimate becomes vacuous. As already observed in the image-restoration experiment, this illustrates that different performance measures may exhibit substantially different generalization behavior, even when evaluated on the same class of optimization problems. 

Most importantly, this experiment illustrates one of the principal motivations for introducing trajectory distributions. Unlike the preceding deterministic experiments, randomness now enters both through the optimization problems and through the optimization algorithm itself. Nevertheless, the same notions of solution criteria, performance functionals, statistical performance measures, and PAC-Bayesian estimates remain applicable without modification.

\section{Conclusion}

We have proposed a statistical-learning framework for optimization algorithms that treats optimization performance as a random quantity induced by a distribution of optimization problems. The central object of the framework is the probability distribution on optimization trajectories, which allows optimization-specific notions of performance to be expressed through measurable performance functionals. This provides a unified perspective on stopping times, contraction factors, solution probabilities, and more general trajectory properties.

Building on this probabilistic foundation, we formulated the learning of optimization algorithms as a statistical learning problem and introduced corresponding notions of empirical and population performance. In contrast to classical worst-case analyses, the proposed framework focuses on the behavior of optimization algorithms over a distribution of problems and naturally accommodates average-case, high-probability, and risk-sensitive notions of performance. As a result, the framework separates the definition of optimization performance from the subsequent learning and generalization analysis.

Within this setting, PAC-Bayesian theory yields generalization guarantees for a broad class of optimization-specific performance criteria. Rather than being tied to a particular loss function, the resulting guarantees apply uniformly to bounded performance functionals and therefore provide a common theoretical treatment of stopping times, convergence behavior, and trajectory-level properties.

More broadly, the present work argues for a shift in perspective. Optimization algorithms are traditionally designed and analyzed through deterministic worst-case guarantees. In contrast, modern learnable optimization algorithms are increasingly trained from data and deployed on distributions of problems. This suggests that optimization algorithms should be studied not only as dynamical systems but also as statistical objects whose performance can be learned, estimated, and generalized. We believe that the proposed framework provides a first step toward such a statistical theory of optimization algorithms.

Several directions remain open. On the theoretical side, it would be interesting to develop PAC-Bayesian guarantees beyond bounded performance functionals, to study problem-adaptive optimization algorithms that explicitly depend on the observed problem instance, and to investigate alternative notions of statistical performance. On the practical side, the framework opens the possibility of learning optimization algorithms that are tailored to specific performance objectives rather than generic convergence criteria. We hope that this viewpoint contributes to a closer integration of optimization, probability, and statistical learning theory in the design and analysis of future optimization algorithms.

\newpage

%\tableofcontents
\appendix

\section{Probabilistic Foundations} 

The main text develops a statistical-learning framework for optimization algorithms and discusses its conceptual motivation and implications. The purpose of this appendix is to provide a self-contained mathematical treatment of the framework. In particular, it contains the probabilistic foundations, examples of learnable optimization algorithms, examples of solution criteria and performance functionals, the construction of statistical performance measures, and the proofs of the PAC-Bayesian generalization results. The material is organized according to the hierarchy 
\[ 
\begin{aligned} 
&\;\text{probabilistic class of optimization problems} \\
&\;\quad \longrightarrow\; \text{trajectory distributions}\\ 
&\;\quad\quad\longrightarrow\; \text{performance functionals}\\
&\;\quad\quad\quad\longrightarrow\; \text{statistical performance measures}\\
&\;\quad\quad\quad\quad\longrightarrow\; \text{learning from data}\\
&\;\quad\quad\quad\quad\quad\longrightarrow\; \text{generalization guarantees}, 
\end{aligned} 
\] which underlies the development of the main text. Throughout the appendix, the focus is on the mathematical construction and analysis of the framework. The motivation, interpretation, and implications of the individual concepts are discussed in the corresponding sections of the main text.

\subsection{Preliminaries and Notation}

We will endow every topological space $X$ with its Borel-$\sigma$-algebra $\mathcal{B}(X)$, and we assume it to be a Polish space, that is, it is separable and admits a complete metrization. Given two spaces $X$ and $Y$, we write their product as $X \times Y$, and, for products with a generic number of terms, we use $\prod_{n=1}^N X_n$. Similarly, the product-$\sigma$-algebra of $\mathcal{B}(X)$ and $\mathcal{B}(Y)$ on $X \times Y$ is denoted by $\mathcal{B}(X) \otimes \mathcal{B}(Y)$. Further, for a generic number of terms, we use $\bigotimes_{n=1}^N \mathcal{B}(X_n)$ to denote the product-$\sigma$-algebra, and, if $X_n \equiv X$ for all $n = 1, ..., N$, also $X^N$ and $\mathcal{B}(X)^{\otimes N}$. 
\begin{Rem}
    Countable products of Polish spaces are again Polish, and we have the equality $\mathcal{B}\left( \prod_{n \in \N} X_n \right) = \bigotimes_{n \in \N} \mathcal{B}(X_n)$ \citep[Lemma 1.2, p.11]{Kallenberg_2021}. Therefore, as we consider a discrete-time algorithm, all resulting product spaces in this work will be Polish spaces endowed with the Borel-$\sigma$-algebra, which coincides with the product-$\sigma$-algebra.
\end{Rem}
We use the same notation for measures: Given two measures $\nu_1$ and $\nu_2$ on $X$ and $Y$, the corresponding product measure on $X \times Y$ is denoted by $\nu_1 \otimes \nu_2$. If the product involves a generic number of measures, this is denoted by $\bigotimes_{n=1}^N \nu_n$, and, if $\nu_n \equiv \nu$ for all $n = 1, ..., N$, we also use $\nu^{\otimes N}$. Further, if necessary, we will write the integral of a function $f: X \to \R$ w.r.t. $\nu$ in the operator notation, that is, $\nu[f] := \int_X \nu(dx) \ f(x) := \int_X f(x) \ \nu(dx)$. 
Especially, this applies when having multiple iterated integrals at once, in which case this has to be read from right to left instead from inside to outside. For example, if we integrate $f: X \times Y \times Z \to \R$ w.r.t. $\nu \otimes \mu \otimes \lambda$, by Fubini's theorem we have: 
\begin{align*}
    \int_{X \times Y \times Z} f(x,y,z) \ (\nu \otimes \mu \otimes \lambda)(dx, dy, dz) 
    &= \int_X  \int_Y \int_Z  f(x,y,z) \ \lambda(dz) \ \mu(dy) \ \nu(dx)\\
    &= \int_X \nu(dx) \ \int_Y \mu(dy) \ \int_Z \lambda(dz) \ f(x,y,z) \,.
\end{align*}
In doing so, the integrand is closer to its \say{next} integrator and one can avoid many brackets. Especially, this applies to kernels, which are of fundamental importance for this work:
\begin{Def}
    Let $(X, \mathcal{X})$ and $(Y, \mathcal{Y})$ be measurable spaces. A \emph{kernel} $\mu$ from $X$ to $Y$ is a mapping 
    $$
    \mu: X \times \mathcal{Y} \to [0, \infty] \,,
    $$
    such that $x \mapsto \mu(x, \setA)$ is measurable for every fixed $\setA \in \mathcal{Y}$, and $\setA \mapsto \mu(x, \setA)$ is a measure for every $x \in X$. Moreover, $\mu$ is called a \emph{probability kernel}, if $\mu(x, Y) = 1$ for every $x \in X$.
\end{Def}
\begin{Not}
    We adopt the standard (abuse of) notation for kernels, that is, $\mu: X \times \mathcal{Y} \to [0, \infty]$ is abbreviated as $\mu: X \to Y$.
\end{Not}
For two (probability) kernels $\mu: X \to Y$ and $\nu: X \times Y \to Z$, the product of $\mu$ and $\nu$ is defined as the kernel $\mu \otimes \nu: X \to Y \times Z$ given by:
$$
    (\mu \otimes \nu)(x)[f] = \int_Y \mu(x, dy) \ \int_Z \nu((x,y),dz) \ f(y,z)\,,
$$
where $f: Y \times Z \to \R$ is a measurable function. That is, for every $x \in X$, $(\mu \otimes \nu)(x, \cdot)$ is a measure on $Y \times Z$, such that for a set $\setA \times \setB \in \mathcal{B}(Y \times Z)$ it holds:
$$
    (\mu \otimes \nu)(x, \setA \times \setB) = \int_\setA \nu((x,y), \setB) \ \mu(x, dy)\,. 
$$
Similarly, the composition of $\mu$ and $\nu$ is given by the kernel $\mu \cdot \nu: X \to Z$ defined through:
$$
    (\mu \cdot \nu)(x)[g] = \int_Y \mu(x,dy) \ \int_Z \nu((x,y), dz) \ g(z) \,,
$$
where $g: Z \to \R$ is a measurable function. That is, for a measurable set $\setB \subset Z$ it holds:
$$
    (\mu \cdot \nu)(x, \setB) = \int_Y \nu((x,y), \setB) \ \mu(x,dy) \,.
$$
Hence, it holds that $(\mu \cdot \nu)(x)[g] = (\mu \otimes \nu)(x)[\mathds{1}_Y \otimes g]$, that is, $(\mu \cdot \nu)(x, \cdot)$ is the marginal of $(\mu \otimes \nu)(x, \cdot)$ on $Z$. Here, $\mathds{1}_\setA$ denotes the indicator function of a set $\setA$, which is equal to one for $x \in \setA$ and zero else, that is, we have $\mathds{1}_\setA(x) = \delta_x(\setA)$, where $\delta_x$ is the Dirac-measure.
\begin{Exa}
    Having two random variables on the same probability space taking values in Polish spaces $X$ and $Y$, say $\rv{X}: \pspace \to X$ and $\rv{Y}: \pspace \to Y$, there exists a regular version of the conditional distribution $\prob_{\rv{Y} \vert \rv{X}}$ of $\rv{Y}$, given $\rv{X}$, such that their joint distribution can be factorized into the marginal and the conditional distribution:
    $$
        \prob_{(\rv{X}, \rv{Y})} = \prob_\rv{X} \otimes \prob_{\rv{Y} \vert \rv{X}}\,,
    $$
    where $(x, \setA) \mapsto \prob_{\rv{Y} \vert \rv{X} = x} \{ \setA \}$ is a probability kernel from $X$ to $Y$.
    Similarly, the marginal of $\rv{Y}$ is given by:
    \begin{align*}
        \prob_\rv{Y} \left\{\setB \right\} &= \prob \{ \rv{Y} \in \setB \} = \prob \{\rv{X} \in X, \ \rv{Y} \in \setB \} = \prob_{(\rv{X}, \rv{Y})} \left\{ X \times \setB \right\}  \\
        &= \left(\prob_\rv{X} \otimes \prob_{\rv{Y} \vert \rv{X}} \right) \{X \times \setB \}
        = \int_X \prob_{\rv{Y} \vert \rv{X} = x} \{ \setB \} \ \prob_\rv{X}(dx)  
        = \left(\prob_\rv{X} \cdot \prob_{\rv{Y} \vert \rv{X}} \right) \left \{ \setB \right\} \,.
    \end{align*}
\end{Exa}
For notational simplicity, when we have $N$ elements $x_1, ..., x_N$ in some space $X$, and we refer to them all at once, that is, to the vector $(x_1, ..., x_N) \in X^N$, we indicate this by $x_{[N]} = (x_1, ..., x_N)$. For example, when integrating w.r.t. a product measure $\prob_\rv{X}^{\otimes N}$ this allows for the more compact notation:
$$
    \prob_\rv{X}^{\otimes N} [f] = \int_{X^N} \prob_\rv{X}^{\otimes N}(dx_{[N]}) \ f(x_{[N]}) := \int_{X^N} f(x_1, ..., x_N) \ \prob_\rv{X}^{\otimes N}(dx_1, ..., dx_N) \,.
$$
Similarly, given such a vector $x_{[N]}$, we refer to its components as $x_1, ..., x_N$, and the correspondence will be clear from the context. Furthermore, as this will turn up in the generalization bounds, the space of measures on an underlying space $X$ is denoted by $\mathcal{M}(X)$, and all probability measures that are absolutely continuous w.r.t. a reference measure $\mu \in \mathcal{M}(X)$ are denote by $\mathcal{M}_{\mathrm{AC}}(\mu) := \{ \nu \in \mathcal{M}(X) \ : \ \text{$\nu \ll \mu$ and $\nu[X] = 1$} \}$. In this context, the Kullback-Leiber divergence between two measures $\mu$ and $\nu$ is defined as:
$$
\divergence{\rm{KL}}{\nu}{\mu} = \begin{cases}
    \nu[\log(f)] = \int_\space{X} \log(f(x)) \ \nu(dx), &\text{$\nu \ll \mu$ with density $f$}\,, \\ +\infty, &\text{otherwise}\,.
\end{cases}
$$ 
Here, we have the following variational formulation due to \citet{DV1975}:
\begin{Lem}[Variational Formulation by Donsker and Varadhan] \label{Lem:Donsker_Varadhan}
    For any measurable and bounded function $h: X \to \R$, it holds that:
    $$
        \log \left( \int_X \exp(h(x)) \ \mu(dx) \right) = \sup_{\nu \ll \mu } \left\{\int_X h(x) \ \nu(dx) - \divergence{\rm{KL}}{\nu}{\mu} \right\} \,.
    $$
\end{Lem}
Finally, the following definitions are used in the so-called \emph{monotone-class argument}, which is needed to derive the distribution of the trajectory of the algorithm:
\begin{Def}
    Let $(X, \mathcal{X})$ be a measurable space. A class $\mathcal{C} \subset \mathcal{X}$ is called a \emph{$\pi$-system}, if it is closed under finite intersection, that is, $\setA, \setB \in \mathcal{C}$ implies $\setA \cap \setB \in \mathcal{C}$. Furthermore, a class $\mathcal{D} \subset \mathcal{X}$ is called a \emph{$\lambda$-system}, if it contains $X$ and it is closed under proper differences and increasing limits. That is, we require $X \in \mathcal{D}$, that $\setA, \setB \in \mathcal{D}$ with $\setA \supset \setB$ implies $\setA \setminus \setB \in \mathcal{D}$, and that $\setA_1, \setA_2, ... \in \mathcal{D}$ with $\setA_n \uparrow \setA$ implies $\setA \in \mathcal{D}$. 
\end{Def}
%The following theorem can be found in \citet[Thm. 1.1, p.10]{Kallenberg_2021}.
\begin{Thm}[Monotone Classes, {\citet[Thm. 1.1, p.10]{Kallenberg_2021}}]\label{Thm:monotone_class}
    For any $\pi$-system $\mathcal{C}$ and $\lambda$-system $\mathcal{D}$ in a measurable space $X$, it holds that:
    $$
        \mathcal{C} \subset \mathcal{D} \ \implies \ \sigma(\mathcal{C}) \subset \mathcal{D},
    $$
    where $\sigma(\mathcal{C})$ is the $\sigma$-algebra generated by $\mathcal{C}$.
\end{Thm}

\subsection{Standing Assumptions}\label{sec:prob-foundations}

\begin{Ass}\label{Ass:polish_spaces}
    The 
    problem parameter space $(\spacePar, \balg{\spacePar})$, 
    algorithmic parameter space $(\spaceHyp, \balg{\spaceHyp})$, 
    state space $(\spaceState, \balg{\spaceState})$, 
    and randomization space $(\spaceRand, \balg{\spaceRand})$ are Polish measurable spaces. Moreover, $\distPar$, $\distHyp$, $\distInit$, and $\distRand$ are probability measures on their respective spaces. 
    Finally, $\map{\proj{\spaceState}}{\spaceState}{\spX}$ denotes the canonical projection onto the optimization space.
\end{Ass}

\begin{Ass}\label{Ass:algorithm_measurable}
    We are given a measurable function $\map{\ell}{\spX\times\spacePar}{[0, \infty]}$, the \emph{loss function}, and a measurable map $\map{\LOA}{\spaceHyp\times\spacePar\times\spaceState\times\spaceRand}{\spaceState}$, the update of the \emph{learnable optimization algorithm (LOA)}.
\end{Ass}
\begin{Ass}\label{Ass:convergence_set_measurable}
    The solution criterion $\solutionCrit\subset \spX\times\spacePar$ is measurable.
\end{Ass}

\section{From Iterative Algorithms to Distributions of Trajectories} \label{appdx:dist-of-traj}

We now construct the probability distribution on optimization trajectories induced by a learnable optimization algorithm. This construction is the probabilistic foundation underlying the statistical
performance measures introduced in the main text.\\

Define the measurable space $(\Omega_{\rm{pre}}, \sigTraj_{\rm{pre}})$ through:
\begin{align*}
    \Omega_{\rm{pre}} := \spaceHyp\times\spacePar\times\spaceState \times \spaceRand^\N, \quad 
    \sigTraj_{\rm{pre}} := \mathcal{B}(\Omega_{\rm{pre}})\,,
\end{align*}
and endow it with the probability measure 
$$
\prob_{\rm{pre}} := \distHyp \otimes \distPar \otimes \distInit \otimes \distRand^{\otimes \N} \,. 
$$
By definition of $(\Omega_{\rm{pre}}, \sigTraj_{\rm{pre}}, \prob_{\rm{pre}})$ and Fubini's theorem, for any set $\setB \in \mathcal{B}(\spaceState)$ we have the following factorization:
\begin{align*}
    &\prob_{\rm{pre}} \left\{ \LOA(\rvHyp, \rvPar, \rvInit, \iter{\rvRand}{1}) \in \setB \right\} \\
    &= \int_{\spaceHyp\times\spacePar\times\spaceState} \prob_\rv{R} \left\{ \LOA(\realHyp, \realPar, \realState, \cdot) \in \setB \right\} \ \left(\distHyp \otimes \distPar \otimes \distInit \right) (\df\realHyp, \df\realPar, \df\realState) \,,
\end{align*}
which motivates the following definition:
\begin{Def}\label{Def:transition_kernel}
    The \emph{transition kernel} of $\LOA$ is given through 
    $$
    \map{\gamma}{\spaceHyp\times\spacePar\times\spaceState}{\spaceState}\,,\quad 
    \left( (\realHyp, \realPar, \realState), \setB \right) \mapsto \distRand \left\{ \LOA(\realHyp, \realPar, \realState, \cdot) \in \setB \right\} \,.
    $$
\end{Def}
%%\begin{Rem}
%%    As will be shown below, for every $(\realHyp, \realPar) \in\spaceHyp \times\spacePar$, the distribution of the process $\xi = \left( \iter{\xi}{t}\right)_{t \in \N_0}$ generated by $\LOA(\realHyp, \realPar, \cdot, \cdot)$ is uniquely defined by the transition kernel $\gamma(\realHyp, \realPar, \cdot): S \to S$ and the initial distribution $\distInit$. This is the probabilistic \emph{generalization} of the fact that the trajectory of a deterministic algorithm is uniquely defined by the initialization and its update-step. 
%%\end{Rem}
%See Example~\ref{ex:intro-transition-kernel} for illustration.
\begin{Lem}\label{lem:trans-kernel-is-prob-kernel}
    Suppose that $\LOA$ satisfies Assumption~\ref{Ass:algorithm_measurable}. Then the transition kernel $\gamma$ is a probability kernel from $\spaceHyp\times\spacePar\times\spaceState$ to $\spaceState$.
\end{Lem}
\begin{proof}
    By definition, it holds that:
    $$
        \gamma((\realHyp, \realPar, \realState), \setB) = \left(\prob_\rv{R} \circ \LOA(\realHyp, \realPar, \realState, \cdot)^{-1} \right) \{ \setB \} \,.
    $$
    Thus, since $\map{\LOA}{\spaceHyp\times\spacePar\times\spaceState \times \spaceRand}{\spaceTraj}$ is measurable, and $\distRand$ can be seen as the constant kernel from $\spaceHyp\times\spacePar\times\spaceState \to \spaceRand$, we get from \citet[Lemma 3.2 (ii), p.56]{Kallenberg_2021} that $\gamma$ is a probability kernel from $\spaceHyp\times\spacePar\times\spaceState$ to $\spaceTraj$. 
\end{proof}    
    
\noindent
The $t$-fold recursive application of the algorithmic map induces a recursive application of the transition kernel. 
\begin{Def}
    The \emph{transition semi-group} $(\gamma^t)_{t \in \N_0}$ is defined recursively by:
    $$
        \gamma^0(\realHyp, \realPar, \realState) := \delta_\realState\,, \quad \gamma^t(\realHyp, \realPar, \realState) := \gamma^{t-1}(\realHyp, \realPar, \realState)  \cdot \gamma(\realHyp, \realPar, \cdot)\,, \quad t \in \N \,.
    $$
\end{Def}
%%
%%\begin{Rem}
%%    \begin{itemize}
%%        \item[(i)] $\gamma^t$ models the t-fold application of the algorithm: applying $\gamma^t$ to the distribution of $\iter{\realTraj}{t_0}$ yields the distribution of $\iter{\realTraj}{t_0 + t}$. Thus, the resulting process is \emph{time-homogeneous}. Nevertheless, one can still model algorithms that have a parameter $\iter{T}{t}$, which itself models the progression of time, as, for example, in Nesterov's accelerated gradient descent \citep{Nesterov_1983}. The prerequisite for this is that the update of this time-parameter can be written in closed-form, that is, $\iter{T}{t+1} = f(\iter{T}{t})$ for some measurable function $f$, such that $\iter{T}{t}$ can be included into the state $\iter{\realTraj}{t}$. 
%%        \item[(ii)] For every $(\realHyp, \realPar) \in\spaceHyp \times\spacePar$, the process $\realTraj$ generated by $\LOA(\realHyp, \realPar, \cdot, \cdot)$ is a time-homogeneous Markov process (w.r.t. the natural filtration generated by the iterates) with initial distribution $\distInit$ and transition semi-group $\left(\gamma^t(\realHyp, \realPar, \cdot)\right)_{t \in \N_0}$. This can be seen by noting that, if $\realHyp$ and $\realPar$ are \emph{fixed}, the equation
%%        $$
%%            \iter{\realTraj}{t+1} = \LOA(\realHyp, \realPar, \iter{\realTraj}{t}, \iter{\eta}{t+1})
%%        $$
%%        corresponds to the so-called \emph{functional representation} of a Markov process.
%%    \end{itemize}
%%\end{Rem}

\paragraph{Key construction principle.} By \citet[Corollary 14.44, p.299]{Klenke_2013}, for every $\realHyp \in\spaceHyp$ and $\realPar \in\spacePar$ there exist a unique probability measure $\psi_{\realHyp, \realPar}$ on $\spaceTraj$, such that for any natural numbers $0 = t_0 < t_1 < ... < t_K$, it holds that:
\begin{equation}\label{Eq:finite_dim_psi}
    \psi_{\realHyp, \realPar} \circ \rvState_J^{-1} = \distInit \otimes \bigotimes_{k=0}^{K-1} \gamma^{t_{k+1} - t_k}(\realHyp, \realPar, \cdot) \,,
\end{equation}
where $J := \{t_0 ,..., t_K\}$ and $\map{\rvState_J}{\spaceTraj}{\spaceState^{K+1}}$ denotes the corresponding coordinate projection. This means that the finite-dimensional distribution of $\psi_{\realHyp, \realPar}$ corresponding to the time-points $t_0, ..., t_K$ is given by the distribution of the iterates $\left( \iter{\realState}{t_0}, ..., \iter{\realState}{t_K}\right)$ generated by $\LOA(\realHyp, \realPar, \cdot, \cdot)$ (with initial distribution $\distInit$). Since the distribution of a stochastic process is uniquely determined by its finite-dimensional distributions \citep[Prop. 4.2, p.84]{Kallenberg_2021}, this implies that, for every $(\realHyp, \realPar) \in\spaceHyp \times\spacePar$, there is exactly one distribution on the space of trajectories $\spaceTraj$, namely $\psi_{\realHyp, \realPar}$, that describes the iterates corresponding to $\LOA(\realHyp, \realPar, \cdot, \cdot)$. 
%%\begin{figure}[t!]
%%    \centering
%%    \includegraphics[width=\textwidth]{Images/kernel_properties.pdf}
%%    \caption{Visualization of the (joint) transition kernel: The upper row shows how the kernel $\gamma(\realHyp, \realPar, \cdot)$ acts on the initial distribution: The iterative concatenation (upper left) transforms the initial distribution of $\iter{\realTraj}{0}$ on $S$ (dark blue) into the distributions of $\iter{\realTraj}{t_1}$ (light blue), $\iter{\realTraj}{t_2}$ (purple), $\iter{\realTraj}{t_3}$ (pink), and $\iter{\realTraj}{t_4}$ (light pink). Similarly, the iterative product (upper middle) transforms the initial distribution on $S$ into a distribution on $S^5$, namely the joint distribution of $(\iter{\realTraj}{0}, \iter{\realTraj}{t_1}, ..., \iter{\realTraj}{t_4})$. Then, this yields the unique distribution $\psi(\realHyp, \realPar, \cdot)$ on $\spaceTraj$ (upper right) for the whole trajectory (orange lines). The lower row shows the same thing on the space $S^N$, just that the initial distribution now is given by $\bigotimes_{n=1}^N \distInit$ and the corresponding kernel is $\Gamma(\realHyp, \realPar_{[N]}, \cdot)$, which acts on all problem instances $\realPar_1, ..., \realPar_N$ at once.}
%%    \label{fig:kernel_properties}
%%\end{figure}

\begin{Thm}\label{Thm:kernel_distr_markov_process}
    Suppose that Assumptions~\ref{Ass:polish_spaces} and \ref{Ass:algorithm_measurable} hold. Then, the map 
    $$
    \map{\psi}{(\spaceHyp\times\spacePar) \times \mathcal{B}(\spaceTraj )}{[0,1]}\,,\quad \left((\realHyp, \realPar), \setB\right) \mapsto \psi_{\realHyp, \realPar}(\setB)
    $$ is the unique probability kernel from $\spaceHyp\times\spacePar$ to $\spaceTraj$, such that \eqref{Eq:finite_dim_psi} holds. 
\end{Thm}
\begin{proof}
    By construction, we have that $\psi(\realHyp, \realPar, \cdot)$ is a probability measure on $\spaceTraj$ for each $(\realHyp, \realPar) \in\spaceHyp \times\spacePar$. Thus, we only have to show the measurability. Again, this follows by a monotone-class argument. For this, define the following two classes of sets:
    \begin{align*}
        &\mathcal{D} := \{\setA \in \balg{\spaceTraj} \ : \ \text{$(\realHyp, \realPar) \mapsto \psi_{\realHyp, \realPar}(\setA)$ is measurable} \} \\
        &\mathcal{C} := \{ \setA^{0} \times ... \times \setA^{K} \times \prod_{k > K} \spaceState \ : \ K \in \N_0, \quad \setA^{0}, ..., \setA^K \in \balg{\spaceState}\} \,.
    \end{align*}
    The first step of the proof is to show that $\mathcal{C} \subset \mathcal{D}$:
    \begin{align*}
        &\psi_{\realHyp, \realPar}\left\{\setA^{0} \times ... \times \setA^{K} \times \prod_{k>K} \spaceState \right\} 
        = \left(\psi_{\realHyp, \realPar} \circ \rvState_{\{0, ..., K\}}^{-1} \right)\left\{\setA^{0} \times ... \times \setA^{K} \right\} \\
        &= \left(\distInit \otimes \bigotimes_{k=0}^{K-1} \gamma^{1}(\realHyp, \realPar, \cdot) \right) \{ \setA^{0} \times ... \times \setA^{K} \} 
        = \left(\distInit \otimes \bigotimes_{k=0}^{K-1} \gamma(\realHyp, \realPar, \cdot) \right) \{ \setA^{0} \times ... \times \setA^{K} \} \,.
    \end{align*}
    Since $\gamma$ is a probability kernel from $\spaceHyp\times \spacePar \times \spaceState$ to $\spaceState$, by \citet[Lemma 3.3 (i), p.58]{Kallenberg_2021} it holds that $\distInit \otimes \bigotimes_{k=0}^{K-1} \gamma(\realHyp, \realPar, \cdot)$ is a probability kernel from $\spaceHyp\times \spacePar$ to $\spaceState^{K+1}$. Thus, the map $(\realHyp, \realPar) \mapsto \distInit \otimes \bigotimes_{k=0}^{K-1} \gamma(\realHyp, \realPar, \cdot)$ is measurable. 
    Therefore, the map $(\realHyp, \realPar) \mapsto \psi\left((\realHyp, \realPar), \setA^{0} \times ... \times \setA^{K} \times \prod_{k>K} \spaceState \right)$ is measurable. Hence, we get that$\mathcal{C} \subset \mathcal{D}$. Further, since $\mathcal{C}$ is the class of cylinder sets, it is clearly $\cap$-stable, that is, it is a $\pi$-system.
    
    The second step of the proof is to show that $\mathcal{D}$ is a $\lambda$-system. Since $\spaceTraj \in \mathcal{C} \subset \mathcal{D}$, we have $\spaceTraj \in \mathcal{D}$. Hence, take $\setA, \setB \in \mathcal{D}$ with $\setA \supset \setB$. By definition of $\mathcal{D}$, we have that both $(\realHyp, \realPar) \mapsto \psi_{\realHyp, \realPar}(\setA)$ and $(\realHyp, \realPar) \mapsto \psi_{\realHyp, \realPar}(\setB)$ are measurable. However, this implies that the map $(\realHyp, \realPar) \mapsto \psi_{\realHyp, \realPar}(\setA \setminus \setB) = \psi_{\realHyp, \realPar}(\setA) - \psi_{\realHyp, \realPar}(\setB)$ is measurable, where the (point-wise) equality follows from $\psi_{\realHyp, \realPar}$ being a probability measure for each $(\realHyp, \realPar) \in \spaceHyp\times P$. Therefore, we have that $\setA \setminus \setB \in \mathcal{D}$, and $\mathcal{D}$ is closed under proper differences. Finally, take $\setA_1, \setA_2, ... \in \mathcal{D}$ with $\setA_n \uparrow \setA$. Then, since $\psi_{\realHyp, \realPar}$ is a measure for each $(\realHyp, \realPar) \in \spaceHyp\times P$, we have the equality:
    $$
        \psi(\realHyp, \realPar, \setA) = \psi_{\realHyp, \realPar}(\setA) = \lim_{n \to \infty} \psi_{\realHyp, \realPar}(\setA_n) = \lim_{n \to \infty} \psi(\realHyp, \realPar, \setA_n) \,.
    $$
    By definition of $\mathcal{D}$, we have that $(\realHyp, \realPar) \mapsto \psi(\realHyp, \realPar, \setA_n)$ is measurable for each $n \in \N$. Thus, the map $(\realHyp, \realPar) \mapsto \psi(\realHyp, \realPar, \setA)$ is the pointwise limit of measurable functions, and therefore measurable itself. Hence, we have $\setA \in \mathcal{D}$, such that $\mathcal{D}$ is a $\lambda$-system. Then, by Theorem~\ref{Thm:monotone_class} we get that
    $
        \sigma(\mathcal{C}) \subset \mathcal{D} 
    $.
    However, since $\mathcal{C}$ is a $\pi$-system, we have that $\sigma(\mathcal{C}) = \balg{\spaceTraj}$. Thus, the map $(\realHyp, \realPar) \mapsto \psi_{\realHyp, \realPar}(\setA)$ is measurable for each $\setA \in \balg{\spaceTraj}$. Consequently $\psi$ is a probability kernel from $\spaceHyp\times \spacePar$ to $\spaceTraj$ and uniqueness follows from that of $\psi_{\realHyp, \realPar}$. 
\end{proof}
\begin{Rem}
    Assumption~\ref{Ass:algorithm_measurable} is needed for measurability, while Assumptions~\ref{Ass:polish_spaces} is needed for the existence of $\psi_{\realHyp, \realPar}$.
\end{Rem}
\noindent
Theorem~\ref{Thm:kernel_distr_markov_process} states that the distribution of the trajectory on $\spaceTraj$ depends \emph{measurably} on the parameters of the problem and the hyperparameters of the algorithm. This allows to define a new probability space, which describes this directly.

\subsection{Definition of the Probability Space of Trajectories}

In the following, we will only be interested in the distribution of the trajectory on $\spaceTraj$ and how it evolves with $\realHyp$ and $\realPar$. Hence, we define the measurable space $(\Omega, \sigTraj)$ as:
\begin{align*}
    \Omega := \spaceHyp\times\spacePar \times \spaceTraj\,, \quad 
    \sigTraj := \mathcal{B}(\Omega)\,,
\end{align*}
and endow it with the probability measure $\left( \distHyp \otimes \distPar \right) \otimes \psi$.
%Further, as before, we denote the coordinate projections by $\rv{X} := (\rvHyp, \rvPar_{[N]}, \realTraj_{[N]})$, that is, $\rvHyp \sim \distHyp$, $\rvPar_1, ..., \rvPar_N \overset{iid}{\sim} \distPar$, and $\realTraj_{[N]} := (\realTraj_1, ..., \realTraj_N) \sim (\distHyp \otimes \distPar^{\otimes N}) \cdot \Psi$. 
%\begin{Rem}
%    Since we define $\prob$ through the probability kernel $\Psi$, it is assumed implicitly that Assumptions~\ref{Ass:polish_spaces} and \ref{Ass:algorithm_measurable} do hold all the time, as they were needed for its construction. 
%\end{Rem}
\begin{Thm}\label{Thm:reg_cond_distr}
    %It holds that:
    \begin{itemize}
        \item[(i)] $\psi$ is a regular version of the conditional distribution of $\rvTraj$, given $\rvHyp$ and $\rvPar$, that is: 
        $$
        \prob_{\rvTraj \vert \rvHyp, \rvPar} \{ \setA \} = \psi(\rvHyp, \rvPar, \setA)\,, \qquad \prob_{(\rvHyp, \rvPar)}\rm{-a.s.}
        $$
        \item[(ii)] $\distPar \otimes \psi$ is a regular version of the conditional distribution of $(\rvPar, \rvTraj)$, given $\rvHyp$, that is:
        \begin{align*}
            \prob_{(\rvPar, \rvTraj) \vert \rvHyp} \{ \setA \} 
            &= \left(\prob_{\rvPar} \otimes \psi(\rvHyp,\rvPar, \cdot) \right) \{\setA\} 
             \,,\quad \prob_{\rvHyp}\rm{-a.s.}\,.
        \end{align*}
    \end{itemize}
\end{Thm}
\begin{proof}
    Basically, these statements are a direct consequence of the definition of the probability space and the properties of $\psi$. 
    %\todo[inline]{There was proof detail, but I believe they can be omitted. Notation is not yet adjusted.}
\end{proof}
\begin{Rem}
    \begin{itemize}
        \item[(i)] In the following, we will use solely these regular versions of the conditional distributions and omit the \say{$\rm{a.s.}$}. 
        \item[(iii)] We will use the following notation interchangeably depending on which seems to be easier to read:
        \begin{align*}
            \prob_{(\rvPar, \rvTraj) \condto \rvHyp=\realHyp, \rvPar=\realPar} \{ \setA \} 
            &= \prob \{ (\rvPar, \rvTraj) \in \setA  \condto  \rvHyp=\realHyp, \rvPar=\realPar\} \\
            &= \prob \{ (\rvPar, \rvTraj) \in \setA \condto  \rvHyp, \rvPar\}(\omega) \,.
        \end{align*}
        Similarly for expected values.
    \end{itemize}
\end{Rem}

\section{Examples of Learnable Optimization Algorithms} \label{appdx:examples-LOA}

Given an optimization problem represented by $\realPar\in\spacePar$ and algorithmic parameters $\realHyp\in\spaceHyp$, the iterative algorithm $\LOA$, starting at some random initialization $\realInit$ from $\distInit$, generates a sequence $(\iter{\realState}{t})_{t\in\N}$ following the recursive definition
\[
    \iter{\realState}{t+1} = \LOA_{\realHyp}(\realPar,\iter{\realState}{t},\iter{\realRand}{t+1}) := \LOA(\realHyp, \realPar,\iter{\realState}{t},\iter{\realRand}{t+1})\,,
\]
where $\iter{\realRand}{t+1}$ is sampled from $\distRand$. The corresponding random trajectories induce a (time-discrete) $\spaceState$-valued stochastic process $(\iter{\rvState}{t})_{t\in\N}$ in the state space $\spaceState$. Beyond the simple example of a learnable Gradient-Descent-like update as in the running Example~\ref{ex:running-ex-toy-alg}, we mention the following example, the illustrate three important modeling features: stochasticity through the randomization space, non-smooth optimization through proximal updates, and internal memory through an enlarged state space. While all these features could be combined, for the sake of a clean presentation, we split them. Consequently, the framework naturally accommodates many classical and modern optimization algorithms, including momentum methods, quasi-Newton methods, adaptive gradient methods, proximal algorithms, and learned optimization algorithms.
\begin{Exa}\label{Exa:examples_algorithm}
    \begin{enumerate}[(i)]
        \item Consider the finite-sum problem $\ell(x, \realPar) := \frac{1}{m} \sum_{i=1}^m f_i(x, \realPar)$ with some problem distribution $\distPar$. For a fixed $\realPar\in\spacePar$, we devise stochastic gradient descent iterates on the optimization space $\spaceState=\spX=\R^n$ by sampling an index $j$ uniformly in $\spaceRand:=\{1, \ldots, m\}$ and updating:
        $$
            \iter{\vx}{t+1} = \iter{\vx}{t} - \realHyp \nabla f_j(\iter{\vx}{t}, \realPar)\,,
        $$
        where $\realHyp\in\spaceHyp=(0,\infty)$ is a step-size. This can be summarized into a single mapping $\LOA$ as:
        $$
            \LOA(\realHyp, \realPar, \iter{\vx}{t}, \iter{\eta}{t+1}) := \iter{\vx}{t} - \realHyp \sum_{i=1}^m \mathds{1}_{\{i\}}(\iter{\eta}{t+1}) \nabla f_i(\iter{\vx}{t}, \realPar) \,,
        $$
        where $\iter{\eta}{t+1} \sim \mathcal{U} \{1, ..., m\}$. 
        \item Since the only crucial assumption in our framework is measurability, we also encompass learnable non-smooth algorithms, for example, using proximal steps or even subgradient evaluations (whenever analytically suitable). For example, for structured optimization problems of the type $\loss(\vx,\realPar) = f(\vx,\realPar) + g(\vx,\realPar)$, where $f$ is smooth and $g$ is non-smooth but prox-friendly (i.e. the associated proximal mapping $\prox{\alpha}{g}(\vx):=\argmin_{\hat{\vx}\in\spX} g(\hat{\vx}) + \frac{1}{2\alpha}\vnorm{\hat{\vx}-\vx}^2$ is easy to evaluate), the learnable algorithm may jointly exploit the structure of the optimization problem as well as the statistics of the distribution of problems $\distPar$. Proximal gradient descent can be modeled in our framework using the update
        \[
            \LOA(\realHyp, \realPar, \iter{\vx}{t}, \iter{\eta}{t+1}) := \prox{\alpha}{g(\cdot,\realPar)}\Big(
                \iter{\vx}{t} - \alpha \nabla f(\iter{\vx}{t},\realPar) 
            \Big)\,,
        \]
        where $\realHyp=\alpha$ is the learnable parameter. In this example, there is no randomization and the state-space matches the optimization space $\spaceState=\spX$. Modern, deep learning based variants could use a neural network $\mathcal N_{d}(\nabla f(\iter{\vx}{t},\realPar)$ to predict the update direction based on the gradient information of $f$. 
        \item To illustrate the role of the state space, let $\spX=\R^n$ and $\spaceState=\R^n\times\R^m$. Consider an update of the state variable $\iter{\realState}{t} = (\iter{y}{t}, \iter{x}{t}) \in \R^{m+n}$ as follows:
        $$
            \iter{\realState}{t+1} := \begin{pmatrix}
                \iter{y}{t+1} \\
                \iter{x}{t+1}
            \end{pmatrix} := 
            \begin{pmatrix}
                \mathcal{N}_1(\realHyp_1, \realPar, \iter{\realState}{t}, \iter{\eta}{t+1}) \\
                \iter{x}{t} - \mathcal{N}_2(\realHyp_2, \realPar, \iter{\realState}{t}, \iter{\eta}{t+1})
            \end{pmatrix} \,,
        $$
        where, additionally to updating the iterates $\iter{x}{t} \in \R^n$ with a neural network $\mathcal{N}_2$, one updates a hidden state $\iter{y}{t} \in \R^m$ with another neural network $\mathcal{N}_1$. Here, the optimization variable is recovered by the projection $\proj{\R^n}(\iter{\realState}{t}) = \iter{x}{t}$ from $\R^{m+n}$ onto $\R^n$. The algorithmic (learnable) parameters $\realHyp$ are given the parameters of the two networks. There is no randomness built into the update.
    \end{enumerate}
\end{Exa}

\subsection{Examples of Transition Kernels} \label{appdx:examples-LOA-kernels}

Having established the abstract trajectory-space construction, we now consider several examples of transition kernels arising from classical and learnable optimization algorithms. The purpose of these examples is not to develop new theory, but to demonstrate how common optimization methods fit naturally into the probabilistic framework.
\begin{Exa} \label{ex:intro-transition-kernel}
    \begin{enumerate}[(i)]
        \item The transition kernel of stochastic gradient descent from Example~\ref{Exa:examples_algorithm} is given by:
        $$
            \distRand \left\{ \LOA(\realHyp, \realPar, x, \cdot) \in \setB \right\}
            := \mathcal{U}_{\{1, ..., m\}} \left\{ x - \realHyp \sum_{i=1}^m \mathds{1}_{\{ i\}} (\cdot) \nabla f_i (x, \realPar) \in \setB \right\} \,.
        $$
        Basically, this is the transition kernel used by \citet{Bianchi_Hachem_Schechtman_2022}, and our definition is a direct generalization of it.
        \item Deterministic optimization algorithms $\LOA(\realHyp,\realPar,\realState)$ arise as a special case of the framework. If the update does not depend on a randomization variable, then \[ \gamma(\realHyp,\realPar,\realState,\setB) = \delta_{\LOA(\realHyp,\realPar,\realState)}(\setB) \] for every measurable set $\setB\subset\spaceState$. Hence, \[ \int_{\spaceState} \varphi(\hat{\realState}) \,\gamma(\realHyp,\realPar,\realState,d\hat{\realState}) = \varphi\!\left( \LOA(\realHyp,\realPar,\realState) \right), \] showing that the transition kernel reduces to the classical deterministic update rule.
%        \item The transition kernel is a direct generalization of the usual algorithmic update: Consider a deterministic algorithm. Then, it holds $\gamma(\realHyp, \realPar, \realState) = \delta_{\LOA(\realHyp, \realPar, \realState)}$, and we get:
%        $$
%            \distRand \left\{ \LOA(\realHyp, \realPar, \realState, \cdot) \in \setB \right\} = \delta_{\LOA(\realHyp, \realPar, \realState,\cdot)}[\setB] = \mathds{1}_\setB(\LOA(\realHyp, \realPar, \realState)) \,.
%        $$
%        Thus, integrating w.r.t. $\gamma$ just yields the new iterate. 
        Taking this approach, we recover the average-case setting of \citet{Pedregosa_Scieur_2020, Scieur_Pedregosa_2020}.
        \item Since the updates in Example~\ref{Exa:examples_algorithm}(ii) and (iii) are deterministic, they are easily expressed using the Dirac measure as transition kernel. 
    \end{enumerate}
\end{Exa}

\section{Examples of Solution Criteria}\label{appdx:examples-solution-crit}

The framework accommodates a broad range of optimization-specific notions of successful termination. Typical examples include:
\begin{Exa}
    \begin{itemize}
        \item[(i)] One could use $\solutionCrit := \{ (\realPar, s) \in \spX\times\spacePar \setsep \ell(\vx,\realPar) \le \eps \}$ for convergence in terms of the loss function, which is measurable, since $\ell$ is measurable.
        \item[(ii)] If $\ell(\vx,\realPar)$ has a unique minimizer $\vx_\realPar^*$, convergence in terms of the iterates could be written as $\solutionCrit := \{ (\vx,\realPar) \in \spX\times\spacePar \setsep \mathrm{dist}\left(\vx, \vx_\realPar^* \right) \le \eps \}$. If $\realPar \mapsto \vx_\realPar^*$ is measurable, $\solutionCrit$ is measurable, because the distance $\mathrm{dist}$ is continuous.
        \item[(iii)] One could use $\solutionCrit := \{ (\vx,\realPar) \in \spX\times\spacePar \ : \ \norm{\grad_\vx \ell(\vx, \realPar)}{} \le \eps \}$ for convergence to a stationary point. For example, if $\grad_\vx \ell$ is (jointly) continuous, $\solutionCrit$ is measurable.
    \end{itemize}
\end{Exa}

\section{Performance Functionals} \label{appdx:perform-func}

In order for the performance functional listed in Section~\ref{sec:perf-measures} to be well-defined, their measurability is established in the following subsections. As a preliminary result, we show the measurability of several examples of optimality-gap functions. Recall, that as usual, we $\vx$ and $\realState$ are connected via $\vx=\proj{\spaceState}(\realState)$.
\begin{enumerate}[(i)]
    \item For $\lyap(\realState,\realPar) = \ell(\vx,\realPar) - \min_{\vx\in\spX}\ell(\vx,\realPar)$, the measurability is induced by that of the loss function (see~Assumption~\ref{Ass:algorithm_measurable}) and the canonical projection $\proj{\spaceState}$.
    \item If $\ell(\vx,\realPar)$ has a unique minimizer $\vx_\realPar^*$ for any $\realPar$, the optimality-gap of the iterates $\lyap(\vx,\realPar)= \mathrm{dist}\left(\vx, \vx_\realPar^* \right)$ is measurable whenever $\realPar \mapsto \vx_\realPar^*$ is measurable, since $\mathrm{dist}$ is continuous.
    \item If $\ell(\cdot,\realPar)$ is differentiable with respect to $\vx$, and $\nabla_{\vx} \ell(\vx,\realPar)$ is (jointly) continuous in $(\vx,\realPar)$, the optimality-gap function $\lyap(\realState,\realPar)= \vnorm{\nabla_{\vx} \ell(\vx,\realPar)}$ is measurable by continuity. 
\end{enumerate}

\subsection{$k$-step optimality gap}

This performance functional in \eqref{eq:perf-func-k-step-gap} is trivial by the measurability assumption of $\lyap$. 

\subsection{Stopping Time}

We rewrite the stopping time in \eqref{eq:def-stopping-time-func} as follows:
\[
    \tau(\realTraj,\realPar) 
    = \inf{t\in\N_0\setsep (\realPar,\iter{\vx}{t}) \in \solutionCrit }
    = \inf_{t\in\N_0} t\cdot(1+\iota_\solutionCrit(\realPar,\iter{\vx}{t}))\,,
\]
with $0$-or-$\infty$-indicator function $\iota_{\solutionCrit}$ of the solution criterion $\solutionCrit$. By measurability of $\solutionCrit$ by Assumption~\ref{Ass:convergence_set_measurable}, for fixed $t\in\N_0$, the map $(\realPar,\realTraj)\mapsto t\cdot(1+\iota_\solutionCrit(\realPar,\iter{\vx}{t}))$ is measurable. Therefore, the stopping time is measurable as the infimum of countably many measurable functions. 

Clearly, the truncated stopping time $\tau_{t_{\max}}= \min(\tau, t_{\max})$, which we use in the PAC-Bayesian setting, for some finite $t_{\max}\in\N_0$ is also measurable. \\

For justifying the name \emph{stopping time} (optional time) in the sense of probability theory, we provide the following lemma. Intuitively, this requires to show that at time $t\in\N_0$, the information collected in the filtration $\iter{\mathcal{F}}{t}$ at time $t$ is indeed enough to decide whether the algorithm did reach the solution criterion. 
\begin{Pro}
    Suppose that Assumption~\ref{Ass:convergence_set_measurable} holds. Then, $\tau$ is a stopping time w.r.t. the filtration $\mathcal{F} = ( \iter{\mathcal{F}}{t})_{t \in \N_0}$, where $\iter{\mathcal{F}}{t}_n := \sigma ( \rvPar, \iter{\rvState}{0}, ..., \iter{\rvState}{t})$.
\end{Pro}
\begin{proof}
    By definition, it holds that $\tau$ is a so-called \emph{hitting time} for the process $\rv{Y}_n = ( \iter{\rv{Y}}{t}_n )_{t \in \N_0}$ given by $\iter{\rv{Y}}{t}_n := (\rv{P}_n, \iter{\rvState}{t})$, that is, it is of the form $ \tau_{\rm{conv}, n} = \inf \left\{ t \in \N_0 \setsep \iter{\rv{Y}}{t}_n \in \setC \right\} $. Then, by \citet[Lemma 9.6 (i)]{Kallenberg_2021}, $\tau$ is weakly $\Tilde{\mathcal{F}}$-optional for every filtration $\Tilde{\mathcal{F}}$ such that $\rv{Y}$ is adapted to $\Tilde{\mathcal{F}}$. In particular, this holds for $\mathcal{F}$, because:
    \begin{align*}
        \iter{\mathcal{F}}{t} 
        &= \sigma ( \rvPar, \iter{\rvState}{0}, ..., \iter{\rvState}{t}) 
        = \sigma ( (\rvPar, \iter{\rvState}{0} ), ..., (\rvPar,\iter{\rvState}{t})) 
        = \sigma ( \iter{\rv{Y}}{0}, ..., \iter{\rv{Y}}{t}) \,.
    \end{align*}
    Since in discrete time there is no distinction between weakly optional times and optional times, the conclusion follows.
\end{proof}

\subsection{Convergence Rate}

Consider \eqref{eq:def-convergence-rate-perf}. For a fixed $t\in\N_0$, assuming the the denominator is strictly positive, we observe that $\lyap(\iter{\realState}{t+1},\realPar)$ composes the measurable optimality-gap function $\lyap$ with the evaluation of the trajectory at $t+1\in\N_0$, which is measurable, hence $\frac{\lyap(\iter{\realState}{t+1},\realPar)}{\lyap(\iter{\realState}{t},\realPar)}$ is measurable. The convergence rate $\contrate$ is the supremum of a countable index set $\N_0$, hence measurable.

\subsection{Contraction Factor}

For showing the measurability of \eqref{eq:def-contraction-factor}, we first observe that the state-mapping $(\realPar,\realTraj)\mapsto \iter{\realState}{\tau}$ with stopping time $\tau$ (as above) is measurable. This is seen by writing $\iter{\realState}{\tau}=\sum_{t\in\N_0} \iter{\realState}{t}\mathds{1}_{\setA_t}(\realPar,\traj)$, where $\setA_t:=\set{(\realPar,\traj)\setsep \tau(\realPar,\traj)=t}$ is measurable, since the stopping time is measurable. Assuming a strictly positive denominator, by measurability of $(a,t)\mapsto a^{1/t}$ on $[0,\infty]\times \N$, we obtain that $\contrate_\tau$ is measurable on $\set{\tau\geq1}$. On $\set{\tau=0}$, $\contrate_\tau$ takes the value $0$, which makes $\contrate_\tau$ a measurable function.

\subsection{Indicators of Trajectory Properties}

The property indicator \eqref{eq:def-convergence-probability-perf} is trivially measurable, since $\setA$ is assumed to be measurable.

\subsection{Oracle Complexity Measures}

A natural extension of stopping-time performance measures is obtained by counting oracle evaluations rather than iterations. Let $c_t(\realPar,\traj)$ denote the number of oracle calls performed during iteration $t$. The corresponding oracle complexity up to a stopping time $\tau(\realPar,\traj)$ is then given by 
\[ 
    C(\realPar,\traj) = \sum_{t=0}^{\tau(\realPar,\traj)-1} c_t(\realPar,\traj)\,. 
\] In the special case where every iteration requires exactly one oracle evaluation, the oracle complexity coincides with the stopping time. More generally, oracle complexity measures the total computational effort required to satisfy the solution criterion. The measurability follows from standard arguments as above.

\section{Learning from Datasets} \label{appdx:learning-from-data}

Following Subsection~\ref{sec:learning-problem}, we aim to learn the best algorithm $\LOA$ from a dataset $\realPar_{[\sizeData]}=(\realPar_1,\ldots,\realPar_{\sizeData})$ according to some performance functional. Furthermore, in order to provide generalization guarantees to the full population $\distPar$, we employ the PAC-Bayesian framework, which requires to consider the dataset as a random variable $\rvPar_{[\sizeData]}=(\rvPar_1,\ldots,\rvPar_\sizeData)$ and to model the joint distribution of all trajectories generated for the dataset. 
\begin{Ass}
    The dataset is independent and identically distributed according to $\distPar$.
\end{Ass}

\subsection{Extension of the Trajectory Distribution to $N$ Problems}

Due to the i.i.d. assumption of the dataset, the construction of the joint trajectory distribution is naturally induced by its factorization to individual problems. Nevertheless, we outline the rigorous details here.

By extending the definition of $(\Omega_{\rm{pre}}, \LOA_{\rm{pre}}, \prob_{\rm{pre}})$ to a dataset and Fubini's theorem, for any cylinder set $\setB_1 \times ... \times \setB_N \in \mathcal{B}(S)^{\otimes N}$ we have the following factorization:
\begin{align*}
    &\prob_{\rm{pre}} \left\{ \left( \LOA(\rvHyp, \rvPar_1, \rvInit_1, \iter{\rvRand}{1}_1), ..., \LOA(\rvHyp, \rvPar_N, \rvInit_N, \iter{\rvRand}{1}_N) \right) \in \setB_1 \times ... \times \setB_N\right\} \\
    &= \int_{\spaceHyp\times\spacePar^N \times \spaceState^N} \prod_{n=1}^N \distRand \left\{ \LOA(\realHyp, \realPar_n, \realState_n, \cdot) \in \setB_n \right\} \ \left(\distHyp \otimes \distPar^{\otimes N} \otimes \distInit^{\otimes N} \right) (d\realHyp, d\realPar_{[N]}, d\realState_{[N]}) \,,
\end{align*}
which motivates the following definition:
\begin{Def}
    We define the joint transition kernel $\Gamma(\realHyp, \realPar_{[N]}, \realState_{[N]}) = \bigotimes_{n=1}^N \gamma(\realHyp, \realPar_n, \realState_n)$.
\end{Def}
\begin{Lem}\label{lem:trans-kernel-is-prob-kernel-N}
    Suppose that $\LOA$ satisfies Assumption~\ref{Ass:algorithm_measurable}. Then the joint transition kernel $\Gamma$ is a probability kernel from $\spaceHyp\times\spacePar^N \times \spaceState^N$ to $\spaceState^N$.
\end{Lem}
\begin{proof}
    On extending Lemma~\ref{lem:trans-kernel-is-prob-kernel}, by definition of $\Gamma$, if $(\realHyp, \realPar_{[N]}, \realState_{[N]})$ is given, we also get that $\Gamma(\realHyp, \realPar_{[N]}, \realState_{[N]})$ is a measure on $\spaceState^N$. Therefore, it remains to show measurability of $\Gamma(\cdot, \setA)$ for fixed $\setA \in \balg{\spaceState}^{\otimes N}$. We do this by a monotone-class argument. For this, define the classes of sets:
    \begin{align*}
        &\mathcal{D} := \{\setA \in \balg{\spaceState}^{\otimes N} \ : \ \text{$(\realHyp, \realPar_{[N]}, \realState_{[N]}) \mapsto \Gamma(\realHyp, \realPar_{[N]}, \realState_{[N]}, \setA)$ is measurable} \} \\
        &\mathcal{C} := \{ \setA^{0} \times ... \times \setA^{N}\ : \ \setA^{0}, ..., \setA^N \in \balg{\spaceState}\} \,.
    \end{align*}
    $\mathcal{C}$ is the class of cylinder sets, which, by definition, is a $\cap$-stable generator of the product-$\sigma$-field. Thus, $\mathcal{C}$ is a $\pi$-system with $\sigma(\mathcal{C}) = \balg{\spaceState}^{\otimes N}$. Furthermore, for any $\setB_1 \times ... \times \setB_N \in \mathcal{C}$, it holds that: 
    \begin{align*}
        \Gamma\left((\realHyp, \realPar_{[N]}, \realState_{[N]}), \setB_1 \times ... \times \setB_N \right) 
        &= \prod_{n=1}^N \gamma(\realHyp, \realPar_n, \realState_n, \setB_n) \,.
    \end{align*}
    Since $\gamma$ is a probability kernel, that is, measurable for any fixed $\setB_n$, 
%    fixed $(\realHyp, \realPar_n, \realState_n)$\todo[inline]{This should be $\setA$?}, 
    it follows that $\mathcal{C} \subset \mathcal{D}$. Thus, it remains to show that $\mathcal{D}$ is a $\lambda$-system. Clearly, it holds that $\spaceState^N \in \mathcal{C} \subset \mathcal{D}$. Thus, take $\setA, \setB \in \mathcal{D}$ with $\setA \supset \setB$. Then, since $\Gamma\left(\realHyp, \realPar_{[N]}, \realState_{[N]} \right)$ was already shown to be a probability measure for each fixed $(\realHyp, \realPar_{[N]}, \realState_{[N]})$, we have the following pointwise equality:
    $$
        \Gamma\left((\realHyp, \realPar_{[N]}, \realState_{[N]}), \setA \setminus \setB \right) =  \Gamma\left((\realHyp, \realPar_{[N]}, \realState_{[N]}), \setA\right) - \Gamma\left((\realHyp, \realPar_{[N]}, \realState_{[N]}), \setB \right) \,.
    $$
    Therefore, since $\setA, \setB \in \mathcal{D}$, we have that the right-hand side is measurable, which in turn implies $\setA \setminus \setB \in \mathcal{D}$. Finally, for $\setA_1, \setA_2, ... \in \mathcal{D}$ with $\setA_n \uparrow \setA$, by continuity of a measure, we have the pointwise equality:
    $$
        \Gamma\left((\realHyp, \realPar_{[N]}, \realState_{[N]}), \setA \right) = \lim_{n \to \infty} \Gamma\left((\realHyp, \realPar_{[N]}, \realState_{[N]}), \setA_n \right) \,.
    $$
    Since limits of measurable functions are measurable, it follows that $\setA \in \mathcal{D}$. Therefore, $\mathcal{D}$ is a $\lambda$-system, and Theorem~\ref{Thm:monotone_class} yields $\balg{\spaceState}^{\otimes N} \subset \mathcal{D}$. Thus, $\Gamma$ is a probability kernel from $\spaceHyp\times\spacePar^N \times \spaceState^N$ to $\spaceState^N$.
\end{proof}
Using a similar reasoning as for a single trajectory, we introduce the joint transition semi-group and obtain existence and uniqueness of the joint probability kernel $\Psi$.
\begin{Def}
    The \emph{joint transition semi-group} is defined by:
    $$
        \Gamma^0(\realHyp, \realPar_{[N]}, \realState_{[N]}) := \delta_{\realState_{[N]}}, \quad 
        \Gamma^t(\realHyp, \realPar_{[N]}, \realState_{[N]}) := \Gamma^{t-1}(\realHyp, \realPar_{[N]}, \realState_{[N]}) \cdot \Gamma(\realHyp, \realPar_{[N]}, \cdot), \quad t \in \N \,.
    $$
\end{Def}

There exists a unique probability measure $\Psi_{\realHyp, \realPar_{[N]}}$ on $\left(\spaceState^N\right)^{\N_0}$, such that:
\begin{equation}\label{Eq:finite_dim_Psi}
    \Psi_{\realHyp, \realPar_{[N]}} \circ \rvState_{J, [N]}^{-1} = \distInit^{\otimes N} \otimes \bigotimes_{k=0}^{K-1} \Gamma^{t_{k+1} - t_k}(\realHyp, \realPar_{[N]}, \cdot) \,,
\end{equation}
where $\rvState_{J, [N]}: \left(\spaceState^N\right)^{\N_0} \to \left(\spaceState^N\right)^{K+1}$ denotes the corresponding coordinate projections on $\left( \spaceState^N \right)^{\N_0}$. As before, $\Psi_{\realHyp, \realPar_{[N]}}$ is the unique measure on $\left(\spaceState^N\right)^{\N_0}$ that describes the distribution of the trajectory of $\traj_{[N]} = (\traj_1, ..., \traj_N)$ in $\spaceState^N$ which is generated by the collection $\left(\LOA(\realHyp, \realPar_1, \cdot, \cdot), ..., \LOA(\realHyp, \realPar_N, \cdot, \cdot) \right)$. Intuitively, and as will be shown later on, this is (up to reordering) the same as considering $N$ individual processes $\traj_1, ..., \traj_N$ on $\spaceTraj$ generated by $\LOA(\realHyp, \cdot, \cdot, \cdot)$ on the problem instances $\realPar_1, ..., \realPar_N$. 
\begin{Thm}\label{Thm:kernel_distr_markov_process-N}
    Suppose that Assumptions~\ref{Ass:polish_spaces} and \ref{Ass:algorithm_measurable} hold. Then, the map 
    $$
    \Psi: (\spaceHyp\times\spacePar^N) \times \mathcal{B}\left( \spaceState^N  \right)^{\otimes \N_0} \to [0,1], \ \left((\realHyp, \realPar_{[N]}), \setB\right) \mapsto \Psi_{\realHyp, \realPar_{[N]}}(\setB)
    $$ is the unique probability kernel from $\spaceHyp\times\spacePar^N$ to $\left( \spaceState^N \right)^{\N_0}$, such that \eqref{Eq:finite_dim_Psi} holds.
\end{Thm}
\begin{proof}
   The proof follows the same arguments as in Theorem~\ref{Thm:kernel_distr_markov_process}. 
\end{proof}
Before we can show the conditional independence in Lemma~\ref{Lem:conditional_independence}, we need another technical lemma.
\begin{Lem}\label{Lem:factorization_joint_semi_group}
    For every $(\realHyp, \realPar_{[N]}, \realState_{[N]}) \in \spaceHyp\times\spacePar^N \times \spaceState^N$, $K \in \N$, $t_0 < t_1 < ... < t_K \in \N_0$, and any set $\left( \setB^{t_0}_1 \times ... \times \setB^{t_0}_N \right) \times ... \times \left( \setB^{t_K}_1 \times ... \times \setB^{t_K}_N \right)$ with $\setB^{t_k}_j \in \mathcal{B}(\spaceState)$, it holds that:
    \begin{align*}
        &\left(\delta_{\realState_{[N]}} \otimes \bigotimes_{k=0}^{K-1} \Gamma^{t_{k+1} - t_k}(\realHyp, \realPar_{[N]}, \cdot) \right) \left\{ \left( \setB^{t_0}_1 \times ... \times \setB^{t_0}_N \right) \times ... \times \left( \setB^{t_K}_1 \times ... \times \setB^{t_K}_N \right) \right\} \\
        &= \prod_{n=1}^N \left(\delta_{\realState_n} \otimes \bigotimes_{k=0}^{K-1} \gamma^{t_{k+1} - t_k}(\realHyp, \realPar_n, \cdot) \right) \left\{ \setB^{t_0}_n \times ... \times \setB^{t_K}_n \right\}
    \end{align*}
\end{Lem}
\begin{proof}
    Inserting $B^{t_k}_j = \spaceState$ if necessary, w.l.o.g. we can restrict to the case $t_k = k$, that is, $t_{k+1} - t_k = 1$. Then, we prove the result by induction. Thus, first, consider the case $K = 1$:
    \begin{align*}
        &\left(\delta_{\realState_{[N]}} \otimes \Gamma(\realHyp, \realPar_{[N]}, \cdot) \right) \left\{ \left( \setB^0_1 \times ... \times \setB^0_N \right) \times \left( \setB^1_1 \times ... \times \setB^1_N \right) \right\} \\
        &= \delta_{\realState_{[N]}} \left[ \mathds{1}_{\setB^0_1 \times ... \times \setB^0_N } \Gamma(\realHyp, \realPar_{[N]}, \cdot, \setB^1_1 \times ... \times \setB^1_N) \right] \,.
        \intertext{By the factorization of $\Gamma$ and Fubini's theorem this can be written as:}
        &= \delta_{\realState_{[N]}} \left[ \prod_{n=1}^N \mathds{1}_{\setB^0_n} \gamma(\realHyp, \realPar_n, \cdot, \setB^1_n) \right] 
        = \prod_{n=1}^N \delta_{\realState_n} \left[  \mathds{1}_{\setB^0_n} \gamma(\realHyp, \realPar_n, \cdot, \setB^1_n) \right] \\
        &= \prod_{n=1}^N \left(\delta_{\realState_n} \otimes \gamma(\realHyp, \realPar_n, \cdot) \left\{ \setB^0_n \times \setB^1_n \right\} \right)\,.
    \end{align*}
    Thus, let the statement be true for $K \in \N$ and consider the case $K+1$. Since we have the recursive definition $\bigotimes_{k=0}^{K-1} \Gamma(\realHyp, \realPar_{[N]}, \cdot) = \left(\bigotimes_{k=0}^{K-2} \Gamma(\realHyp, \realPar_{[N]}, \cdot) \right) \otimes \Gamma(\realHyp, \realPar_{[N]}, \cdot)$,
    the statement follows directly from the factorization of $\Gamma$, the induction hypothesis, and Fubini's theorem:
    \begin{align*}
        &\left(\delta_{\realState_{[N]}} \otimes \bigotimes_{k=0}^{K} \Gamma(\realHyp, \realPar_{[N]}, \cdot)\right) \left\{ \left( \setB^{0}_1 \times ... \times \setB^{0}_N \right) \times ... \times \left( \setB^{K+1}_1 \times ... \times \setB^{K+1}_N \right)  \right\} \\
        &= \left(\delta_{\realState_{[N]}} \otimes \bigotimes_{k=0}^{K-1} \Gamma(\realHyp, \realPar_{[N]}, \cdot)\right) \left[ \mathds{1}_{\setB^{0}_1 \times ... \times \setB^{K}_N}  \Gamma(\realHyp, \realPar_{[N]}, \cdot, \setB^{K+1}_1 \times ... \times \setB^{K+1}_N)  \right] \\
        &= \left(\delta_{\realState_{[N]}} \otimes \bigotimes_{k=0}^{K-1} \Gamma(\realHyp, \realPar_{[N]}, \cdot)\right) \left[ \prod_{n=1}^N \mathds{1}_{\setB^{0}_n \times ... \times \setB^{K}_n} \gamma(\realHyp, \realPar_{[N]}, \cdot, \setB^{K+1}_n)  \right] \\
        &= \prod_{n=1}^N \left(\delta_{\realState_n} \otimes \bigotimes_{k=0}^{K-1} \gamma(\realHyp, \realPar_n, \cdot)\right) \left[  \mathds{1}_{\setB^{0}_n \times ... \times \setB^{K}_n}  \gamma(\realHyp, \realPar_{[N]}, \cdot, \setB^{K+1}_n)  \right] \\
        &= \prod_{n=1}^N \left(\delta_{\realState_n} \otimes \bigotimes_{k=0}^{K} \gamma(\realHyp, \realPar_n, \cdot)\right) \left \{  \setB^{0}_n \times ... \times \setB^{K+1}_n \right \} \,. %\qedhere
    \end{align*}
\end{proof}
The generated processes are conditionally independent.
\begin{Lem}\label{Lem:conditional_independence}
    Suppose that Assumptions~\ref{Ass:polish_spaces} and \ref{Ass:algorithm_measurable} hold. Then, for any $(\realHyp, \realPar_{[N]}) \in\spaceHyp \times\spacePar^N$, we have 
    $$
        \Psi(\realHyp, \realPar_{[N]}) \cong \bigotimes_{n=1}^N \psi(\realHyp, \realPar_n) \,.
    $$
\end{Lem} 
\begin{proof}
    For the proof, we use the following notation: We write cylinder sets $\setA \in \mathcal{B}\left(\spaceState^N\right)^{\otimes \N_0}$ with their corresponding coordinates $\setA = \setA_1 \times ... \times \setA_N$, where each $\setA_i \in \mathcal{B}\left(\spaceTraj\right)$. This is justified by the fact that $\left(\spaceState^N\right)^{\N_0} \cong \left( \spaceTraj\right)^N$ and $ \mathcal{B}\left(\spaceState^N\right)^{\otimes \N_0} \cong  \mathcal{B}\left(\spaceTraj\right)^{\otimes N}$. Further, $\mathcal{B}\left(\spaceState^N\right)^{\otimes \N_0}$ is generated by cylinder sets of the form $\left(\setB^{0}_1 \times ... \times \setB^0_N\right) \times ... \times \left(\setB^{K}_1 \times ... \times \setB^K_N\right) \times \prod_{k > K} \spaceState^N$, while $\mathcal{B}\left(\spaceTraj\right)^{\otimes N}$ is generated by cylinder sets of the form $\left(\setB^0_1 \times ... \times \setB^K_1 \times \prod_{k > K} \spaceState \right) \times ... \times \left(\setB^0_N \times ... \times \setB^K_N \times \prod_{k > K} \spaceState \right)$, which are just a reordering of each other.

    $\Psi(\realHyp, \realPar_{[N]})$ is uniquely defined by its values on a $\cap$-stable generator of $\mathcal{B}\left(\spaceState^N \right)^{\otimes \N_0}$. As stated above, such a generator is given by cylinder sets of the form:
    $$
    \underbrace{\left(\setB^{0}_1 \times ... \times \setB^0_N\right)}_{=: \setC^0} \times ... \times \underbrace{\left(\setB^{K}_1 \times ... \times \setB^K_N\right)}_{=: \setC^K} \times \prod_{k > K} \spaceState^N \,.
    $$
    Thus, denoting $J := \{0, ..., K\}$, we get:
    \begin{align*}
        &\Psi(\realHyp, \realPar_{[N]}, \setA_1 \times ... \times \setA_N) 
        = \Psi_{\realHyp, \realPar_{[N]}} \left \{ \setA_1 \times ... \times \setA_N\right\} \\
        &= \left(\Psi_{\realHyp, \realPar_{[N]}} \circ \rvState_{J, [N]}^{-1} \right) \left \{  \setC^0 \times ... \times \setC^K \right\} 
        = \left( \distInit^{\otimes N} \otimes \bigotimes_{k=0}^{K-1} \Gamma(\realHyp, \realPar_{[N]}, \cdot) \right) \left \{ \setC^0 \times ... \times \setC^K \right\} \\
        &= \int_{\spaceState^N} \distInit^{\otimes N} (d\realState_{[N]}) \ \left( \delta_{\realState_{[N]}} \otimes \bigotimes_{k=0}^{K-1} \Gamma(\realHyp, \realPar_{[N]}, \cdot) \right) \left \{ \setC^0 \times ... \times \setC^K  \right\} \,.
        \intertext{By Lemma~\ref{Lem:factorization_joint_semi_group}, this is the same as:}
        &= \int_{\spaceState^N} \distInit^{\otimes N} (d\realState_{[N]}) \ \prod_{i=1}^N \left( \delta_{\realState_n} \otimes \bigotimes_{k=0}^{K-1} \gamma(\realHyp, \realPar_n, \cdot) \right) \left \{ \setB_n^0 \times ... \times \setB_n^K \right\} \,. \\
        \intertext{By Fubini's theorem, this is the same as:}
        &= \prod_{i=1}^N \int_{\spaceState} \distInit(d\realState_n) \ \left( \delta_{\realState_n} \otimes \bigotimes_{k=0}^{K-1} \gamma(\realHyp, \realPar_n, \cdot) \right) \left \{ \setB_n^0 \times ... \times \setB_n^K \right\}  \\
        &= \prod_{i=1}^N \left( \distInit \otimes \bigotimes_{k=0}^{K-1} \gamma(\realHyp, \realPar_n, \cdot) \right) \left\{ \setB_n^0 \times ... \times \setB_n^K \right\} \\
        &= \prod_{i=1}^N \left( \psi_{\realHyp, \realPar_n} \circ \rvState_J^{-1} \right) \left\{ \setB_n^0 \times ... \times \setB_n^K \right\} 
        = \prod_{i=1}^N \psi_{\realHyp, \realPar_n}  \left\{ \setB_n^0 \times ... \times \setB_n^K \times \prod_{k > K} \spaceState \right\} \,.
    \end{align*}
    Since $\setB_n^0 \times ... \times \setB_n^K \times \prod_{k > K} \spaceState$ is the n-th \say{coordinate} of the set $\setC^{0} \times .... \times \setC^{K} \times \prod_{k > K} \spaceState^N$, this shows that all finite-dimensional marginals of $\Psi(\realHyp, \realPar_{[N]})$ coincide with the corresponding ones of $\otimes_{n=1}^N \psi(\realHyp, \realPar_n)$. Thus, by the Kolmogorov Extension Theorem \citep[Thm. 14.36, p.295]{Klenke_2013} (uniqueness of the projective limit), we get the statement.
    %%$$
    %%    \Psi(\realHyp, \realPar_{[N]}) \cong \bigotimes_{n=1}^N \psi(\realHyp, \realPar_n) \,,
    %%$$
    %%that is, up to reordering, $\Psi(\realHyp, \realPar_{[N]})$ is just the product of $\psi(\realHyp, \realPar_n)$, $n=1, ..., N$.
\end{proof}
\begin{Rem}
    Based on Lemma~\ref{Lem:conditional_independence}, to simplify the argument, we can (and often will) identify the measure $\Psi(\realHyp, \realPar_{[N]})$ on $\left(\spaceState^N\right)^{\N_0}$ with the measure $\bigotimes_{n=1}^N \psi(\realHyp, \realPar_n)$ on $\left(\spaceTraj\right)^N$, and correspondingly $\left(\left(\spaceState^N\right)^{\N_0}, \mathcal{B}\left(\spaceState^N \right)^{\otimes \N_0}\right)$ with $\left(\left(\spaceTraj\right)^N, \mathcal{B}\left(\spaceTraj \right)^{\otimes N}\right)$. 
\end{Rem}

\subsection{Definition of the Probability Space}

We define the measurable space $(\Omega, \sigTrajN)$ as:
\begin{align*}
    \Omega := \spaceHyp \times \spacePar^N \times \left( \spaceState^N \right)^{\N_0}, \quad 
    \sigTrajN := \balg{\Omega}\,,
\end{align*}
and endow it with the probability measure $\left(\distHyp\times\distPar^{\otimes N} \right) \otimes \Psi$. 
\begin{Thm}\label{Thm:reg_cond_distr-N}
    \begin{itemize}
        \item[(i)] $\Psi$ is a regular version of the conditional distribution of $\rvTraj_{[N]}$, given $\rvHyp$ and $\rvPar_{[N]}$, that is: 
        $$
        \prob_{\rvTraj_{[N]} \vert \rvHyp, \rvPar_{[N]}} \{\setA\} = \Psi(\rvHyp, \rvPar_{[N]}, \setA), \qquad \prob_{(\rvHyp, \rvPar_{[N]})}\rm{-a.s.}
        $$ 
        \item[(ii)] $\prob_{\rvPar_{[N]}} \otimes \Psi$ is a regular version of the conditional distribution of $(\rvPar_{[N]}, \rvTraj_{[N]})$, given $\rvHyp$, that is: 
        \begin{align*}
            \prob_{(\rvPar_{[N]}, \rvTraj_{[N]})  \vert \rvHyp} \{ \setA \} 
            &= \left(\prob_{\rvPar_{[N]}} \otimes \Psi(\rvHyp, \rvPar_{[N]},\cdot)\right) \{\setA\}, \qquad \prob_{\rvHyp}\rm{-a.s.} 
        \end{align*}
    \end{itemize}
\end{Thm}
\begin{proof}
    Basically, these statements are a direct consequence of the definition of the probability space and the properties of $\psi$. 
\end{proof}
\begin{Cor}\label{Cor:conditional_independence}
    Given any $(\realHyp, \realPar_{[N]}) \in\spaceHyp \times\spacePar^N$, the processes $\rvTraj_{[N]} = (\rvTraj_1, ..., \rvTraj_N)$ are independent, and $\rvTraj_i$ is independent of $\realPar_j$ for $i \neq j$. It holds that
    \[
        \prob_{\rvTraj_{[N]} \vert \rvHyp, \rvPar_{[N]}} = \bigotimes_{n=1}^N \prob_{\rvTraj_n \vert \rvHyp, \rvPar_n}\,.
    \]
\end{Cor} 
\begin{proof}
    This follows directly from Lemma~\ref{Lem:conditional_independence} and Theorem~\ref{Thm:reg_cond_distr-N}.
\end{proof}
\begin{Cor}\label{Cor:sum_and_cond_expectation}
    Let $f_1, ..., f_N: \spaceTraj \to \R_{\ge 0}$ be measurable functions. Then it holds that:
    $$
        \expectation \left\{ \sum_{n=1}^N f_n(\rvTraj_n) \ \vert \ \rvHyp, \rvPar_{[N]} \right\} =  \sum_{n=1}^N \expectation \left\{ f_n(\rvTraj_n) \ \vert \ \rvHyp, \rvPar_n \right\} \,.
    $$
\end{Cor}
\begin{proof}
    By Theorem~\ref{Thm:reg_cond_distr-N}, it holds that:
    \begin{align*}
        \expectation \left\{ \sum_{n=1}^N f_n(\rvTraj_n) \ \vert \ \rvHyp, \rvPar_{[N]} \right\}
        &= \int_{(\spaceTraj)^N}  \sum_{n=1}^N f_n(z_n) \ \Psi(\rvHyp, \rvPar_{[N]}, dz_{[N]}) \\
        &= \sum_{n=1}^N \int_{(\spaceTraj)^N}  f_n(z_n) \ \Psi(\rvHyp, \rvPar_{[N]}, dz_{[N]}) \,.
        \intertext{By Lemma~\ref{Lem:conditional_independence} and Fubini's theorem, this is the same as:}
        &= \sum_{n=1}^N \int_{\spaceTraj} f_n(z_n) \ \psi(\rvHyp, \rvPar_n, dz_n) \cdot \prod_{i\neq n}^N \underbrace{\int_{\spaceTraj} \ \psi(\rvHyp, \rvPar_i, dz_i)}_{=1} 
        \\
        &= \sum_{n=1}^N \int_{\spaceTraj} f_n(z_n) \ \psi(\rvHyp, \rvPar_n, dz_n) 
        = \sum_{n=1}^N \expectation \left\{ f_n(\rvTraj_n) \ \vert \ \rvHyp, \rvPar_n \right\} \,.
    \end{align*}
    where the last step follows from applying Theorem~\ref{Thm:reg_cond_distr-N} again.
\end{proof}

\section{PAC-Bayesian Generalization}

In this section, we prove the main PAC-Bayesian generalization Theorem~\ref{Thm:generalization_general_bounded_function} in Subsection~\ref{proof:Thm:generalization_general_bounded_function}, as well as the special version for Theorem~\ref{Thm:gen_property_trajectory} in Section~\ref{Proof:Thm:gen_property_trajectory}. The corollaries in the main text that specialize the performance functional are immediate consequences. 

\subsection{Preparation of the proof of Theorem~\ref{Thm:generalization_general_bounded_function}} 
\begin{Lem}\label{Lem:bounded_exponential}
    Let $\map{\performFunc}{\spacePar\times\spaceTraj}{[0, \infty)}$ be a measurable function that is bounded from above by $\performFunc_{\rm{max}} \in \R$. Then, for every $\lambda \in \R$ it holds that:
    $$
        \expectation \left \{ \exp \left( \lambda \left( \frac{1}{N} \sum_{n=1}^N \expectation \left\{\performFunc(\rvPar_n, \rvTraj_n) \ \vert \ \rvHyp, \rvPar_n \right\} - \expectation_{(\rvPar, \rvTraj) \vert \rvHyp} \left\{ \performFunc  \right\} \right) - \frac{\lambda^2}{2N} \performFunc_{\rm{max}}^2 \right) \right\} \le 1 \,.
    $$
\end{Lem}
\begin{proof}
    This follows from standard arguments like Jensen's inequality, Fubini's theorem, Hoeffding's inequality, and the i.i.d. assumption. For brevity, we define $\Bar{f} := \expectation_{(\rvPar, \rvTraj) \vert \rvHyp} \{ \performFunc\}$.

    By Lemma~\ref{Lem:conditional_independence} and Corollary~\ref{Cor:sum_and_cond_expectation} it holds that:
    \begin{align*}
        &\expectation \left \{ \exp \left( \lambda \left( \frac{1}{N} \sum_{n=1}^N \expectation \left\{\performFunc(\rvPar_n, \rvTraj_n) \ \vert \ \rvHyp, \rvPar_n \right\} - \Bar{f} \right) - \frac{\lambda^2}{2N} \performFunc_{\rm{max}}^2 \right) \right\} \\
        &= \expectation \left \{ \exp \left( \lambda \left( \expectation \left\{ \frac{1}{N} \sum_{n=1}^N \performFunc(\rvPar_n, \rvTraj_n) \ \Bigl \vert \  \rvHyp, \rvPar_{[N]} \right\} - \Bar{f} \right) - \frac{\lambda^2}{2N} \performFunc_{\rm{max}}^2 \right) \right\} \,.
        \intertext{Since the other terms are constant w.r.t. $\Psi$, this is the same as:}
        &= \expectation \left \{ \exp \left( \expectation \left\{ \lambda \left(  \frac{1}{N} \sum_{n=1}^N \performFunc(\rvPar_n, \rvTraj_n) - \Bar{f} \right) - \frac{\lambda^2}{2N} \performFunc_{\rm{max}}^2 \ \Bigl \vert \ \rvHyp, \rvPar_{[N]} \right\} \right) \right\} \,.
        \intertext{By Jensen's inequality, this can be bounded by:}
        &\le \expectation \left \{ \expectation \left\{ \exp \left(  \lambda \left(  \frac{1}{N} \sum_{n=1}^N \performFunc(\rvPar_n, \rvTraj_n) - \right) - \frac{\lambda^2}{2N} \performFunc_{\rm{max}}^2  \right) \ \Bigl \vert \ \rvHyp, \rvPar_{[N]}  \right\} \right\} \,.
        \intertext{Since we take the (total) expectation on the outside, this is the same as:}
        &= \expectation \left \{ \expectation  \left\{ \exp \left( \lambda \left( \frac{1}{N} \sum_{n=1}^N \performFunc(\rvPar_n, \rvTraj_n) - \Bar{f} \right) - \frac{\lambda^2}{2N} \performFunc_{\rm{max}}^2 \right) \ \Bigl \vert \ \rvHyp \right\} \right\} \,. 
        \intertext{Since $\performFunc_{\rm{max}}$ is a constant, and by the properties of the exponential function, this can be written as:}
        &= \expectation \left \{ \exp\left( - \frac{\lambda^2}{2N} \performFunc_{\rm{max}}^2 \right) \expectation \left\{ \prod_{n=1}^N \exp\left( \frac{\lambda}{N} \left(\performFunc(\rvPar_n, \rvTraj_n) - \Bar{f} \right) \right) \ \Bigl \vert \ \rvHyp \right\} \right\} \,.
        \intertext{By Lemma~\ref{Lem:conditional_independence} and Fubini's theorem, this is the same as:}
        &= \expectation \left \{ \exp\left( - \frac{\lambda^2}{2N} \performFunc_{\rm{max}}^2 \right) \prod_{n=1}^N \expectation \left\{ \exp\left( \frac{\lambda}{N} \left(\performFunc(\rvPar_n, \rvTraj_n) - \Bar{f}\right) \right) \ \Bigl \vert \ \rvHyp \right\} \right\} \\
        &= \expectation \left \{ \exp\left( - \frac{\lambda^2}{2N} \performFunc_{\rm{max}}^2 \right) \prod_{n=1}^N \expectation_{(\rvPar_n, \rvTraj_n) \vert \rvHyp} \left\{ \exp\left( \frac{\lambda}{N} \left(\performFunc - \Bar{f}\right) \right)  \right\} \right\} \,.
        \intertext{Since the parameters are i.i.d., this is the same as:}
        &= \expectation \left \{ \exp\left( - \frac{\lambda^2}{2N} \performFunc_{\rm{max}}^2 \right) \prod_{n=1}^N \expectation_{(\rvPar, \rvTraj) \vert \rvHyp} \left\{ \exp\left( \frac{\lambda}{N} \left(\performFunc - \Bar{f}\right) \right)  \right\} \right\} \,.
        \intertext{By Hoeffding's lemma, this can be bounded by:}
        &\le \expectation \left \{ \exp\left( - \frac{\lambda^2}{2N} \performFunc_{\rm{max}}^2 \right) \prod_{n=1}^N \exp\left( \frac{\lambda^2}{8N^2} (2 \performFunc_{\rm{max}})^2 \right) \right\} \le 1 \,.
    \end{align*}
    This concludes the proof.
\end{proof}

\subsection{Proof of Theorem~\ref{Thm:generalization_general_bounded_function}} \label{proof:Thm:generalization_general_bounded_function}

\begin{proof}
    Abbreviate $\Bar{f} := \expectation_{(\rvPar, \rvTraj) \vert \rvHyp} \{ \performFunc\}$ and $\hat{f} := \frac{1}{N} \sum_{n=1}^N \expectation_{(\rvPar_n, \rvTraj_n) \vert \rvHyp, \rvPar_n} \left\{\performFunc\right\}$. Then, since Lemma~\ref{Lem:bounded_exponential} holds for any $\lambda \in \R$, we get that:
    $$
        \expectation_{(\rvPar_{[N]}, \rvHyp)} \left \{ \exp \left( \lambda \left( \Bar{f} - \hat{f} \right) - \frac{\lambda^2}{2N} f_{\rm{max}}^2 \right) \right\} \le 1 \,.
    $$
    By Fubini's theorem applied to $\prob_{(\rvPar_{[N]}, \rvHyp)} = \prob_{\rvPar_{[N]}} \otimes \prob_\rvHyp$, the variational formulation of Donsker--Varadhan, and the linearity of the integral, this is the same as:
    \begin{align*}
        \expectation_{\rvPar_{[N]}} \left \{  \exp \left( 
        \sup_{\rho\ll \distHyp}  \lambda \left( \rho [ \Bar{f} ] - \rho [ \hat{f} ] \right) 
        - \divergence{\rm{KL}}{\rho}{\prob_\rvHyp} - \frac{\lambda^2}{2N} f_{\rm{max}}^2 \right) \right\} \le 1 \,.
    \end{align*}
    Then, for any $s \in \R$ we get from Markov's inequality:
    $$
        \prob_{\rvPar_{[N]}} \left \{  \sup_{\rho\ll \rvHyp}  \lambda \left( \rho [\Bar{f} ] - \rho [\hat{f} ] \right) - \divergence{\rm{KL}}{\rho}{\prob_\rvHyp} - \frac{\lambda^2}{2N} f_{\rm{max}}^2  \ge s \right\} \le \exp(-s) \,.
    $$
    Using $s = \log(1/\varepsilon)$ yields:
    $$
        \prob_{\rvPar_{[N]}} \left \{  \sup_{\rho\ll\distHyp}  \lambda \left( \rho [\Bar{f} ] - \rho [\hat{f} ] \right) - \divergence{\rm{KL}}{\rho}{\prob_\rvHyp}  - \frac{\lambda^2}{2N} f_{\rm{max}}^2  \ge \log \left( \frac{1}{\varepsilon}\right) \right\} \le \varepsilon \,.
    $$
    Restricting to $\lambda > 0$, reformulating, and taking the complementary event yields the result.
    %\hfill \qedsymbol
\end{proof} 
\begin{Rem}
    Under some mild assumptions, by a covering argument, the provided bound could also be made uniform in $\lambda$. 
\end{Rem}

\subsection{Preparation of the proof of Theorem~\ref{Thm:gen_property_trajectory}} 
\begin{Lem}\label{Lem:prob_traj_property}
    Let $\setA \subset \spacePar\times \spaceTraj$ be measurable. Then, for any $\lambda \in \R$, it holds that:
    $$
        \expectation \left\{ \exp\left( -\frac{\lambda}{N} \sum_{n=1}^N \mathds{1}_\setA(\rvPar_n, \rvTraj_n)\right)\right\}
        = \expectation \left\{ \left(1 - \left[ 1 - \exp\left( -\frac{\lambda}{N} \right) \right] \prob_{(\rvPar, \rvTraj) \vert \rvHyp} \left\{ \setA \right\}\right)^N \right\} \,.
    $$
\end{Lem}
\begin{proof}
    We get from Theorem~\ref{Thm:reg_cond_distr-N}:
    \begin{align*}
        \expectation \left\{ \exp\left( -\frac{\lambda}{N} \sum_{n=1}^N \mathds{1}_\setA(\rvPar_n, \rvTraj_n)\right)\right\}
        &= \expectation \left\{ \left(\expectation_{(\rvPar, \rvTraj) \vert \rvHyp} \left\{ \exp\left( -\frac{\lambda}{N} \mathds{1}_\setA \right) \right\} \right)^N \right\}  \,.
    \end{align*}
    Then, for any $\realHyp \in\spaceHyp$, the inner integral is given by:
    \begin{align*}
        \expectation_{(\rvPar, \rvTraj) \vert \rvHyp=\realHyp} \left\{ \exp\left( -\frac{\lambda}{N} \mathds{1}_\setA \right) \right\}
        &= \prob_{(\rvPar, \rvTraj) \vert \rvHyp=\realHyp} \left\{ \setA^c \right\} + \exp\left( -\frac{\lambda}{N} \right) \prob_{(\rvPar, \rvTraj) \vert \rvHyp=\realHyp} \left\{ \setA \right\} \\
        &= 1 - \prob_{(\rvPar, \rvTraj) \vert \rvHyp=\realHyp} \left\{ \setA \right\} + \exp\left( -\frac{\lambda}{N} \right) \prob_{(\rvPar, \rvTraj) \vert \rvHyp=\realHyp} \left\{ \setA \right\}  \\
        &= 1 - \left[1 - \exp\left( -\frac{\lambda}{N} \right)\right] \prob_{(\rvPar, \rvTraj) \vert \rvHyp=\realHyp} \left\{ \setA \right\} \,,
    \end{align*} 
    which concludes the proof.
\end{proof}
\noindent
The term inside the power is the Laplace transform of a Bernoulli random variable with parameter $\prob_{(\rvPar, \rvTraj) \vert \rvHyp} \left\{ \setA \right\}$. Generalization bounds for Bernoulli-random variables were already presented, for example, by \citet{Catoni_2007}, whose approach will be applied in the following. 

\begin{Lem}[\citet{Catoni_2007}]\label{Lem:properties_Phi}
    For the function 
    $$
    \Phi_a(p) := - \frac{1}{a} \log \left( 1 - \left[ 1 - \exp(-a) \right] p \right) 
    $$
    it holds that:
    \begin{itemize}
        \item[(i)] $\Phi_a$ is an increasing one-to-one mapping of the unit-interval onto itself,
        \item[(ii)] $\Phi_a$ is strictly convex for $a > 0$ and strictly concave for $a < 0$,
        \item[(iii)] $\Phi_a^{-1}$ is given by:
        $
            \Phi_a^{-1}(q) := \frac{1-\exp(-aq)}{1-\exp(-a)} \,.
        $
    \end{itemize}
\end{Lem}
\begin{proof}
    That $\Phi$ is one-to-one can be observed by just plugging in the formula for $\Phi_a^{-1}$, which then also shows (iii). The first derivative of $\Phi_a$ is given by:
    $$
        \frac{\partial}{\partial p} \Phi_a(p) = \frac{1}{a} \frac{1 - \exp(-a)}{1 - [1 - \exp(-a)] p }\,.
    $$
    The only way that this term could be negative is given when $1 - \exp(-a)$ is negative. In this case, the whole second term has to be negative, as the denominator is positive. However, since $1 - \exp(-a)$ can only be negative, when $a$ is, this shows that $\frac{\partial}{\partial p} \Phi_a(p)$ is positive. Thus, $\Phi_a$ is increasing. Similarly, the second derivative of $\Phi_a$ is given by:
    $$
        \frac{\partial^2}{\partial p^2} \Phi_a(p) = \frac{1}{a} \left(\frac{1 - \exp(-a)}{1 - [1-\exp(-a)] p}\right)^2 \,.
    $$
    This term is strictly negative/positive, when $a$ is, and therefore shows (ii).
\end{proof}

\noindent
This yields the following corollary, which is needed to apply the PAC-Bayesian argument:
\begin{Cor}\label{Cor:bound_on_expectation_property_trajectory}
    Let $\setA \subset \spacePar\times \spaceTraj$ be measurable. Then, for any $\lambda \in \R$ it holds that:
    \begin{align*}
        &\expectation \left\{ \exp\left( \lambda \left[ \Phi_{\frac{\lambda}{N}}\left(\prob_{(\rvPar, \rvTraj) \vert \rvHyp} \left\{ \setA \right\}\right) - \frac{1}{N} 
        \sum_{n=1}^N \mathds{1}_\setA(\rvPar_n, \rvTraj_{n}) \right] \right) \right\}  
        = 1 \,.
    \end{align*}
    In particular, it holds that:
    $$
        \expectation_{\rvPar_{[N]}} \left \{ 
        \expectation_\rvHyp \left\{ 
        \exp \left( \lambda \left[ \Phi_{\frac{\lambda}{N}}\left( \prob_{(\rvPar, \rvTraj) \vert \rvHyp} \left\{ \setA \right\} \right) 
        - \frac{1}{N} \sum_{n=1}^N \prob_{(\rvPar_n, \rvTraj_n) \vert \rvHyp, \rvPar_n } \left\{ \setA \right\} \right] 
        \right) \right\} \right\} \le 1 \,.
    $$
\end{Cor}
\begin{proof}
    The first statement follows directly by combining Lemma~\ref{Lem:prob_traj_property} and Lemma~\ref{Lem:properties_Phi}. The second statement then follows from Fubini's theorem and Jensen's inequality. In detail, the proof is as follows.

    By Lemma~\ref{Lem:prob_traj_property} and Lemma~\ref{Lem:properties_Phi}, for every $\lambda \in \R$ we have the equality:
    $$
        \expectation \left\{ \exp\left( \frac{\lambda}{N} \sum_{n=1}^N \one{\setA}(\rvPar_n, \rvTraj_{n}) \right) \right\} = 
        \expectation\left\{\exp\left(-\lambda \Phi_{\frac{\lambda}{N}}\left(\prob_{(\rvPar, \rvTraj) \vert \rvHyp} \left\{ \setA \right\}\right)\right) \right\}\,,
    $$
    This directly yields the first statement. Further, one gets from Fubini's theorem:
    \begin{align*}
        &\expectation_{\rvPar_{[N]}} \left \{ 
        \expectation_\rvHyp \left\{ 
        \exp \left( \lambda \left[ \Phi_{\frac{\lambda}{N}}\left( \prob_{(\rvPar, \rvTraj) \vert \rvHyp} \left\{ \setA \right\} \right) 
        - \frac{1}{N} \sum_{n=1}^N \prob_{(\rvPar_n, \rvTraj_n) \vert \rvHyp, \rvPar_n } \left\{ \setA \right\} \right] 
        \right) \right\} \right\} \\
        &=\expectation \left\{ 
        \exp \left( \lambda \left[ \Phi_{\frac{\lambda}{N}}\left( \prob_{(\rvPar, \rvTraj) \vert \rvHyp} \left\{ \setA \right\} \right) 
        - \frac{1}{N} \sum_{n=1}^N \prob_{(\rvPar_n, \rvTraj_n) \vert \rvHyp, \rvPar_n } \left\{ \setA \right\} \right] 
        \right) \right\}  \\
        &= \expectation \left\{ 
        \exp \left( \lambda \left[ \Phi_{\frac{\lambda}{N}}\left( \prob_{(\rvPar, \rvTraj) \vert \rvHyp} \left\{ \setA \right\} \right) 
        - \frac{1}{N} \sum_{n=1}^N \expectation \left\{ \mathds{1}_\setA(\rvPar_n, \rvTraj_n) \ \vert \  \rvHyp, \rvPar_{n} \right\} \right] 
        \right) \right\}  \,.
        \intertext{By Lemma~\ref{Cor:sum_and_cond_expectation}, this is the same as:}
        &= \expectation \left\{ 
        \exp \left( \lambda \left[ \Phi_{\frac{\lambda}{N}}\left( \prob_{(\rvPar, \rvTraj) \vert \rvHyp} \left\{ \setA \right\} \right) 
        - \expectation \left\{\frac{1}{N} \sum_{n=1}^N  \mathds{1}_\setA(\rvPar_n, \rvTraj_n) \ \vert \  \rvHyp, \rvPar_{[N]}\right\} \right] 
        \right) \right\} \\
        &= \expectation \left\{ 
        \exp \left( \expectation \left\{\lambda \left[ \Phi_{\frac{\lambda}{N}}\left( \prob_{(\rvPar, \rvTraj) \vert \rvHyp} \left\{ \setA \right\} \right) 
        - \frac{1}{N} \sum_{n=1}^N  \mathds{1}_\setA(\rvPar_n, \rvTraj_n) \ \vert \  \rvHyp, \rvPar_{[N]} \right] \right\}
        \right) \right\} \,.
        \intertext{By Jensen's inequality, this can be bounded by:}
        &\le \expectation \left\{  \expectation \left\{
        \exp \left( \lambda \left[ \Phi_{\frac{\lambda}{N}}\left( \prob_{(\rvPar, \rvTraj) \vert \rvHyp} \left\{ \setA \right\} \right) 
        - \frac{1}{N} \sum_{n=1}^N  \mathds{1}_\setA(\rvPar_n, \rvTraj_n) \right] 
        \right) \ \Bigl \vert \  \rvHyp, \rvPar_{[N]}  \right\} \right\} \\
        &= \expectation\left \{ 
        \exp \left( \lambda \left[ \Phi_{\frac{\lambda}{N}}\left( \prob_{(\rvPar, \rvTraj) \vert \rvHyp} \left\{ \setA \right\} \right) 
        - \frac{1}{N} \sum_{n=1}^N  \mathds{1}_\setA(\rvPar_n, \rvTraj_n) \right] 
        \right) \right\} = 1 \,.
    \end{align*}

\end{proof}

\subsection{Proof of Theorem~\ref{Thm:gen_property_trajectory}} \label{Proof:Thm:gen_property_trajectory}
\begin{proof}
    Abbreviate $\Bar{p} := \prob_{(\rvPar, \rvTraj) \vert \rvHyp} \left\{ \setA \right\}$ and $\hat{p} := \frac{1}{N} \sum_{n=1}^N \prob_{(\rvPar_n, \rvTraj_n) \vert \rvHyp, \rvPar_n } \left\{ \setA \right\}$.
    Then, by Corollay~\ref{Cor:bound_on_expectation_property_trajectory} it holds that:
    $$
        \expectation_{\rvPar_{[N]}} \left \{ 
        \expectation_\rvHyp \left\{ 
        \exp \left( \lambda \left[ \Phi_{\frac{\lambda}{N}}\left(  \Bar{p} \right) 
        -  \hat{p}\right] 
        \right) \right\} \right\} \le 1 \,.
    $$
    Therefore, by the Donsker--Varadhan variational formulation, we have:
    \begin{align*}
        1 &\ge \expectation_{\rvPar_{[N]}} \left \{ 
        \exp \left( \sup_{\rho\ll\distHyp} \ \lambda \rho \left[ \Phi_{\frac{\lambda}{N}} \circ \Bar{p}  
        - \hat{p} \right] - \divergence{\rm{KL}}{\rho}{\distHyp}
        \right) \right\} \\
        &= \expectation_{\rvPar_{[N]}} \left \{ 
        \exp \left( \sup_{\rho\ll\distHyp} \ \lambda \left(\rho \left[ \Phi_{\frac{\lambda}{N}} \circ \Bar{p} \right] 
        - \rho \left[\hat{p} \right] \right) - \divergence{\rm{KL}}{\rho}{\distHyp} \right) \right\} \,.
        \intertext{Since $\lambda > 0$, $\Phi$ is convex by Lemma~\ref{Lem:properties_Phi}. Thus, applying Jensen's inequality yields:}
        &\ge \expectation_{\rvPar_{[N]}} \left \{ 
        \exp \left( \sup_{\rho\ll\distHyp} \ \lambda \left(\Phi_{\frac{\lambda}{N}} \left(\rho \left[ \Bar{p} \right] \right)
        - \rho \left[\hat{p} \right] \right) - \divergence{\rm{KL}}{\rho}{\distHyp} \right) \right\}
    \end{align*}
    Then, Markov's inequality yields:
    \begin{align*}
        \prob_{\rvPar_{[N]}} \left \{ \sup_{\rho\ll\distHyp} \ \lambda \left(\Phi_{\frac{\lambda}{N}} \left(\rho \left[ \Bar{p} \right] \right)
        - \rho \left[\hat{p} \right] \right) - \divergence{\rm{KL}}{\rho}{\distHyp} \ge \log\left(\frac{1}{\varepsilon} \right) \right\}
        \le \varepsilon \,,
    \end{align*}
    from which the conclusion follows.
\end{proof}

\acks{M. Sucker and P. Ochs acknowledge funding by the German Research Foundation under Germany’s Excellence Strategy – EXC number 2064/1 – 390727645. }

\newpage
\bibliography{bibliography.bib}

\newpage
\section{Architecture for the Experiment on Quadratic Functions} \label{Appendix:architecture_quad}

\definecolor{test_1}{RGB}{241, 241, 242} 
\definecolor{test_2}{RGB}{25, 149, 173} 
\definecolor{test_3}{RGB}{161, 214, 226} 
\definecolor{test_4}{RGB}{188, 186, 190} 

\begin{figure}[h!]
    \centering
    \begin{tikzpicture}
        \tikzset{conv/.style={black,draw=black,fill=test_1,rectangle,minimum height=0.5cm}}
        \tikzset{linear/.style={black,draw=black,fill=test_2,rectangle,minimum height=0.5cm}}
        \tikzset{relu/.style={black,draw=black,fill=test_3,rectangle,minimum height=0.25cm}}

        % Directions
        \node (d1) at (0, 1.0) {\tiny{$\iter{d}{t}_1$}};
        \node (d2) at (0, 0.0) {\tiny{$\iter{d}{t}_2$}};
        \node (d3) at (-0.4, -1.0) {\tiny{$\iter{d}{t}_1 \odot \iter{d}{t}_2$}};

        % Update Block
        \node[conv,rotate=90,minimum width=2.75cm] (conv1) at (2,0) {\tiny{\texttt{Conv2d(3,10,1,bias=F)}}};
        \node[conv,rotate=90,minimum width=2.75cm] (conv2) at (2.5,0) {\tiny{\texttt{Conv2d(10,10,1,bias=F)}}};
        \node[relu,rotate=90,minimum width=2.75cm] (relu1) at (3.0,0) {\tiny{\texttt{ReLU}}};
        
        \node[conv,rotate=90,minimum width=2.75cm] (conv3) at (3.5,0) {\tiny{\texttt{Conv2d(10,10,1,bias=F)}}};
        \node[relu,rotate=90,minimum width=2.75cm] (relu2) at (4,0) {\tiny{\texttt{ReLU}}};

        \node[conv,rotate=90,minimum width=2.75cm] (conv4) at (4.5,0) {\tiny{\texttt{Conv2d(10,10,1,bias=F)}}};
        \node[relu,rotate=90,minimum width=2.75cm] (relu3) at (5.,0) {\tiny{\texttt{ReLU}}};

        \node[conv,rotate=90,minimum width=2.75cm] (conv5) at (5.5,0) {\tiny{\texttt{Conv2d(10,10,1,bias=F)}}};
        \node[relu,rotate=90,minimum width=2.75cm] (relu4) at (6.,0) {\tiny{\texttt{ReLU}}};

        \node[conv,rotate=90,minimum width=2.75cm] (conv6) at (6.5,0) {\tiny{\texttt{Conv2d(10,10,1,bias=F)}}};
        \node[conv,rotate=90,minimum width=2.75cm] (conv7) at (7,0) {\tiny{\texttt{Conv2d(10,1,1,bias=F)}}};

        \node (d) at (8, 0) {\tiny{$\iter{d}{t}$}};

        \draw[->] (d1) to [out=0, in=180] (conv1.north);
        \draw[->] (d2) -- (conv1);
        \draw[->] (d3) to [out=0, in=180] (conv1.north);

        \draw[-] (conv1) -- (conv2) -- (relu1) -- (conv3) -- (relu2) -- (conv4) -- (relu3) -- (conv5) -- (relu4) -- (conv6) -- (conv7);
        \draw[->] (conv7) -- (d);

        % Norms
        \node (n1) at (0, -1.8) {\tiny{$\iter{n}{t}_1$}};
        \node (n2) at (0, -2.3) {\tiny{$\iter{n}{t}_2$}};
        \node (n3) at (0, -2.8) {\tiny{$\iter{n}{t}_3$}};
        \node (n4) at (0, -3.3) {\tiny{$\iter{n}{t}_4$}};

        % Step-size block
        \node[linear,rotate=90,minimum width=2.3cm] (lin1) at (2, -2.6) {\tiny{\texttt{Linear(4,8,bias=F)}}};
        \node[relu,rotate=90,minimum width=2.3cm] (relu2_1) at (2.5, -2.6) {\tiny{\texttt{ReLU}}};

        \node[linear,rotate=90,minimum width=2.3cm] (lin2) at (3, -2.6) {\tiny{\texttt{Linear(8,8,bias=F)}}};
        \node[relu,rotate=90,minimum width=2.3cm] (relu2_2) at (3.5, -2.6) {\tiny{\texttt{ReLU}}};

        \node[linear,rotate=90,minimum width=2.3cm] (lin3) at (4., -2.6) {\tiny{\texttt{Linear(8,8,bias=F)}}};
        \node[relu,rotate=90,minimum width=2.3cm] (relu2_3) at (4.5, -2.6) {\tiny{\texttt{ReLU}}};

        \node[linear,rotate=90,minimum width=2.3cm] (lin4) at (5., -2.6) {\tiny{\texttt{Linear(8,8,bias=F)}}};
        \node[linear,rotate=90,minimum width=2.3cm] (lin5) at (5.5, -2.6) {\tiny{\texttt{Linear(8,1,bias=F)}}};

        \node (s) at (7, -2.6) {\tiny{$\iter{\beta}{t}$}};

        \draw[->] (n1) to [out=0, in=180] (lin1.north);
        \draw[->] (n2) to [out=0, in=180] (lin1.north);
        \draw[->] (n3) to [out=0, in=180] (lin1.north);
        \draw[->] (n4) to [out=0, in=180] (lin1.north);

        \draw[-] (lin1) -- (relu2_1) -- (lin2)  -- (relu2_2) -- (lin3) -- (relu2_3) -- (lin4);
        \draw[->] (lin5) -- (s);

        % Combine to new point
        \node (xk) at (10, 0) {\tiny{$\iter{x}{t}$}};
        \node (new) at (10, -1.3) {\tiny{$\iter{x}{t+1} := \iter{x}{t} + \iter{\beta}{t} \cdot \iter{d}{t}$}};
        \draw[->] (d) to [out=0, in=90] (new.north);
        \draw[->] (xk) -- (new);
        \draw[->] (s) to [out=0, in=270] (new.south);
        
    \end{tikzpicture}
    \caption{Update step of $\mathcal{A}$ for quadratic problems: The directions $\iter{d}{t}_1$, $\iter{d}{t}_2$ and $\iter{d}{t}_1 \odot \iter{d}{t}_2$ are inserted as different channels into the \texttt{Conv2d}-block, which performs $1 \times 1$ \say{convolutions}, that is, the algorithm acts coordinate-wise on the input. The scales $\iter{n}{t}_1$, ..., $\iter{n}{t}_4$ get transformed separately by the fully-connected block.}
\end{figure}
\noindent
Here, the step-size $\iter{\beta}{t}$ is computed by a fully-connected block (without bias) and ReLU activation functions based on the the inputs $\iter{n}{t}_1 = \log(1 + \norm{\nabla \ell(\iter{x}{t}, \realPar)}{})$, $\iter{n}{t}_2 = \log(1 + \norm{\iter{x}{t} - \iter{x}{t-1}}{})$, $\iter{n}{t}_3 = \log(1 + \ell(\iter{x}{t}, \realPar))$ and $\iter{n}{t}_4 = \log(1 + \ell(\iter{x}{t-1}, \realPar))$. On the other hand, the direction $\iter{d}{t}$ is computed by a $1 \times 1$-convolutional block, where the input-channels are given by the normalized gradient $d_1 := \frac{\nabla \ell(\iter{x}{t}, \realPar)}{\norm{\nabla \ell(\iter{x}{t}, \realPar)}{}}$, the normalized momentum term $d_2 := \frac{\iter{x}{t} - \iter{x}{t-1}}{\norm{\iter{x}{t} - \iter{x}{t-1}}{}}$, and their pointwise product $d_1 \odot d_2$, and we use 10 channels in each hidden layer.

\newpage
\section{Architecture for the Image Processing Experiment} \label{Appendix:architecture_img}

\definecolor{test_1}{RGB}{241, 241, 242} 
\definecolor{test_2}{RGB}{25, 149, 173} 
\definecolor{test_3}{RGB}{161, 214, 226} 
\definecolor{test_4}{RGB}{188, 186, 190} 

\begin{figure}[h!]
    \centering
    \begin{tikzpicture}
        \tikzset{conv/.style={black,draw=black,fill=test_1,rectangle,minimum height=0.5cm}}
        \tikzset{linear/.style={black,draw=black,fill=test_2,rectangle,minimum height=0.5cm}}
        \tikzset{relu/.style={black,draw=black,fill=test_3,rectangle,minimum height=0.25cm}}

        % Norms
        \node (reg) at (-3.95, 1.75) {\tiny{$\lambda$}};
        \node (n_1) at (-4.075, 1.4) {\tiny{$\iter{n}{t}_{1}$}};
        \node (n_2) at (-4.05, 1.05) {\tiny{$\iter{n}{t}_{2}$}};
        \node (ld) at (-4.1, 0.7) {\tiny{$\iter{\Delta \ell}{t}$}};
        \node (ldd) at (-4.1, 0.35) {\tiny{$\iter{\Delta h}{t}$}};
        \node (ldr) at (-4.1, 0.0) {\tiny{$\iter{\Delta r}{t}$}};
        \node (mg) at (-4, -0.35) {\tiny{$\iter{g}{t}$}};
        \node (sp_1) at (-4, -0.7) {\tiny{$\iter{s}{t}_1$}};
        \node (sp_2) at (-4.05, -1.05) {\tiny{$\iter{s}{t}_2$}};
        \node (sp_3) at (-4.05, -1.4) {\tiny{$\iter{s}{t}_3$}};
        \node (sp_4) at (-4.05, -1.75) {\tiny{$\iter{s}{t}_4$}};

        % Step-size block
        \node[linear,rotate=90,minimum width=2.75cm] (lin1) at (-2.5, 0) {\tiny{\texttt{Linear(11,30,bias=F)}}};
        \node[relu,rotate=90,minimum width=2.75cm] (relu2_1) at (-2.0, 0) {\tiny{\texttt{ReLU}}};

        \node[linear,rotate=90,minimum width=2.75cm] (lin2) at (-1.5, 0) {\tiny{\texttt{Linear(30,20,bias=F)}}};
        \node[relu,rotate=90,minimum width=2.75cm] (relu2_2) at (-1.0, 0) {\tiny{\texttt{ReLU}}};
        
        \node[linear,rotate=90,minimum width=2.75cm] (lin3) at (-0.5, 0) {\tiny{\texttt{Linear(20,10,bias=F)}}};
        \node[relu,rotate=90,minimum width=2.75cm] (relu2_3) at (0, 0) {\tiny{\texttt{ReLU}}};

        \node[linear,rotate=90,minimum width=2.75cm] (lin4) at (0.5, 0) {\tiny{\texttt{Linear(10,4,bias=F)}}};

        \draw[->] (reg) to [out=0, in=180] (lin1.north);
        \draw[->] (n_1) to [out=0, in=180] (lin1.north);
        \draw[->] (n_2) to [out=0, in=180] (lin1.north);
        \draw[->] (ld) to [out=0, in=180] (lin1.north);
        \draw[->] (ldd) to [out=0, in=180] (lin1.north);
        \draw[->] (ldr) to [out=0, in=180] (lin1.north);
        \draw[->] (mg) to [out=0, in=180] (lin1.north);
        \draw[->] (sp_1) to [out=0, in=180] (lin1.north);
        \draw[->] (sp_2) to [out=0, in=180] (lin1.north);
        \draw[->] (sp_3) to [out=0, in=180] (lin1.north);
        \draw[->] (sp_4) to [out=0, in=180] (lin1.north);

        \node (s1) at (1.8, 1.25) {\tiny{$s_1$}};
        \node (s2) at (1.8, 0.325) {\tiny{$s_2$}};
        \node (s3) at (1.8, -0.325) {\tiny{$s_3$}};
        \node (s4) at (1.8, -1.25) {\tiny{$s_4$}};

        \draw[->] (lin4.south) to [out=0, in=180] (s1.west);
        \draw[->] (lin4.south) to [out=0, in=180] (s2.west);
        \draw[->] (lin4.south) to [out=0, in=180] (s3.west);
        \draw[->] (lin4.south) to [out=0, in=180] (s4.west);

        \draw[-] (lin1) -- (relu2_1) -- (lin2) -- (relu2_2) -- (lin3) -- (relu2_3) -- (lin4);

        % Directions
        \node (d1) at (3.4, 1.25) {\tiny{$s_1 \cdot \iter{d}{t}_1$}};
        \node (d2) at (3.4, 0.325) {\tiny{$s_2 \cdot \iter{d}{t}_2$}};
        \node (d3) at (3.4, -0.325) {\tiny{$s_3 \cdot \iter{d}{t}_3$}};
        \node (d4) at (3.4, -1.25) {\tiny{$s_4 \cdot \iter{d}{t}_4$}};

        \draw[->] (s1) to (d1.west);
        \draw[->] (s2) to (d2.west);
        \draw[->] (s3) to (d3.west);
        \draw[->] (s4) to (d4.west);

        % Update Block
        \node[conv,rotate=90,minimum width=2.7cm] (conv1) at (5.,0) {\tiny{\texttt{Conv2d(4,15,1,bias=F)}}};
        \node[relu,rotate=90,minimum width=2.7cm] (relu1) at (5.5,0) {\tiny{\texttt{ReLU}}};
        
        \node[conv,rotate=90,minimum width=2.7cm] (conv2) at (6.,0) {\tiny{\texttt{Conv2d(15,15,1,bias=F)}}};
        \node[relu,rotate=90,minimum width=2.7cm] (relu2) at (6.5,0) {\tiny{\texttt{ReLU}}};

        \node[conv,rotate=90,minimum width=2.7cm] (conv3) at (7.,0) {\tiny{\texttt{Conv2d(15,15,1,bias=F)}}};
        \node[relu,rotate=90,minimum width=2.7cm] (relu3) at (7.5,0) {\tiny{\texttt{ReLU}}};

        \node[conv,rotate=90,minimum width=2.7cm] (conv4) at (8.,0) {\tiny{\texttt{Conv2d(15,2,1,bias=F)}}};

        \node (d_out_1) at (9.2, 0.5) {\tiny{$\iter{d}{t}_{out,1}$}};
        \node (d_out_2) at (9.2, -0.5) {\tiny{$\iter{d}{t}_{out,2}$}};

        \node (xk) at (10.4, 2.) {\tiny{$\iter{x}{t}$}};
        
        \draw[->] (d1) to [out=0, in=180] (conv1.north);
        \draw[->] (d2) to [out=0, in=180] (conv1.north);
        \draw[->] (d3) to [out=0, in=180] (conv1.north);
        \draw[->] (d4) to [out=0, in=180] (conv1.north);

        \draw[-] (conv1) -- (relu1) -- (conv2) -- (relu2) -- (conv3) -- (relu3) -- (conv4);
        \draw[->] (conv4.south) to [out=0, in=180] (d_out_1.west);
        \draw[->] (conv4.south) to [out=0, in=180] (d_out_2.west);

        % Combine to new point
        \node (new) at (10.4, 0) {\tiny{$\iter{x}{t+1}$}};
        \draw[->] (xk) to [out=270, in=90] (new.north);
        \draw[->] (d_out_1) to [out=0, in=180] (new.west);
        \draw[->] (d_out_2) to [out=0, in=180] (new.west);
        
    \end{tikzpicture}
    \caption{Algorithmic update for the image processing problem: Based on the given eleven features, the first block computes four weights $s_1, ..., s_4$, which are used to perform a weighting of the different directions $\iter{d}{t}_1, ..., \iter{d}{t}_4$, which are used in the second block. This second block consists of a 1x1-convolutional blocks, which computes two update direction $\iter{d}{t}_{out, 1}$ and $\iter{d}{t}_{out, 1}$. Then, we update $\iter{x}{t+1} := \iter{x}{t} + \frac{\norm{\nabla \ell(\iter{x}{t}, \theta)}{}}{L} \iter{d}{t}_{out, 1} - \frac{1}{L} \nabla \ell(\iter{x}{t}, \theta) + \norm{\iter{x}{t} - \iter{x}{t-1}}{} \iter{d}{t}_{out, 2}$.}
\end{figure}
\noindent
The update of the learned algorithm reads:
$$
\iter{x}{t+1} := \iter{x}{t} + \frac{\norm{\nabla \ell(\iter{x}{t}, \realPar)}{}}{L} \iter{d}{t}_{out, 1} - \frac{1}{L} \nabla \ell(\iter{x}{t}, \realPar) + \norm{\iter{x}{t} - \iter{x}{t-1}}{} \iter{d}{t}_{out, 2} \,.
$$
Here, the directions $\iter{d}{t}_{out, 1}$ and $\iter{d}{t}_{out, 2}$ are predicted by a 1x1 convolutional block with ReLU-activation functions. These are predicted based on the reweighted directions $\iter{d}{t}_1, ..., \iter{d}{t}_4$, which are given as the normalized gradient, the normalized momentum, the normalized gradient of the data-fidelity term, and the normalized gradient of the regularization term. Here, the weights $s_1, ..., s_4$ for the reweighting are predicted by a fully-connected block with ReLU-activation functions based on the features $\iter{n}{t}_1 = \log(1 + \norm{\nabla \ell(\iter{x}{t}, \realPar)}{})$, $\iter{n}{t}_2 = \log(1 + \norm{\iter{x}{t} - \iter{x}{t-1}}{})$, $\Delta \iter{\ell}{t} := \ell(\iter{x}{t}, \realPar) - \ell(\iter{x}{t-1}, \realPar)$, $\Delta \iter{r}{t} := r(\iter{x}{t}, \realPar) - r(\iter{x}{t-1}, \realPar)$, where $r$ is the regularization term, $\Delta \iter{h}{t} := h(\iter{x}{t}, \realPar) - h(\iter{x}{t-1}, \realPar)$, where $h$ is the data-fidelity term, $\iter{g}{t} := \max_{i=1, ..., n} \abs{\nabla \ell(\iter{x}{t}, \realPar)}_i$, the scalarproducts $\iter{s}{t}_1, ..., \iter{s}{t}_4$ between the (normalized) gradient and the (normalized) momentum, between the (normalized) gradient of the regularization term and the (normalized) momentum, between the (normalized) gradient of the data-fidelity term and the (normalized) momentum, between the (normalized) gradient of the regularization term and the (normalized) gradient of the data-fidelity term, and, finally, the regularization parameter $\lambda$.

\newpage
\section{Architecture for the LASSO Experiment} \label{Appendix:architecture_lasso}

\definecolor{test_1}{RGB}{241, 241, 242} 
\definecolor{test_2}{RGB}{25, 149, 173} 
\definecolor{test_3}{RGB}{161, 214, 226} 
\definecolor{test_4}{RGB}{188, 186, 190} 

\begin{figure}[h!]
    \centering
    \begin{tikzpicture}
        \tikzset{conv/.style={black,draw=black,fill=test_1,rectangle,minimum height=0.5cm}}
        \tikzset{linear/.style={black,draw=black,fill=test_2,rectangle,minimum height=0.5cm}}
        \tikzset{relu/.style={black,draw=black,fill=test_3,rectangle,minimum height=0.25cm}}

        % Norms
        \node (gn_1) at (-4.05, 2.2) {\tiny{$\iter{n}{t}_{1,>}$}};
        \node (gn_2) at (-4, 1.8) {\tiny{$\iter{n}{t}_{1,0}$}};
        \node (mn_1) at (-4.05, 1.4) {\tiny{$\iter{n}{t}_{2,>}$}};
        \node (mn_2) at (-4, 1.0) {\tiny{$\iter{n}{t}_{2,0}$}};
        \node (sn_1) at (-4, 0.6) {\tiny{$\iter{n}{t}_{3,>}$}};
        \node (sn_2) at (-3.95, 0.2) {\tiny{$\iter{n}{t}_{3,0}$}};
        \node (ld) at (-4, -0.2) {\tiny{$\iter{\Delta \ell}{t}$}};
        \node (ldns) at (-4.05, -0.6) {\tiny{$\iter{\Delta g}{t}$}};
        \node (lds) at (-4.075, -1.0) {\tiny{$\iter{\Delta h}{t}$}};
        \node (sp_1) at (-3.975, -1.4) {\tiny{$\iter{s}{t}_>$}};
        \node (sp_2) at (-4, -1.8) {\tiny{$\iter{s}{t}_0$}};
        \node (reg) at (-3.875, -2.2) {\tiny{$\lambda$}};

        % Step-size block
        \node[linear,rotate=90, minimum width=2.75cm] (lin1) at (-2.5, 0) {\tiny{\texttt{Linear(12,30,bias=F)}}};
        \node[relu,rotate=90, minimum width=2.75cm] (relu2_1) at (-2.0, 0) {\tiny{\texttt{ReLU}}};

        \node[linear,rotate=90, minimum width=2.75cm] (lin2) at (-1.5, 0) {\tiny{\texttt{Linear(30,20,bias=F)}}};
        \node[relu,rotate=90, minimum width=2.75cm] (relu2_2) at (-1.0, 0) {\tiny{\texttt{ReLU}}};
        
        \node[linear,rotate=90, minimum width=2.75cm] (lin3) at (-0.5, 0) {\tiny{\texttt{Linear(20,10,bias=F)}}};
        \node[relu,rotate=90, minimum width=2.75cm] (relu2_3) at (0, 0) {\tiny{\texttt{ReLU}}};

        \node[linear,rotate=90,minimum width=2.75cm] (lin4) at (0.5, 0) {\tiny{\texttt{Linear(10,8,bias=F)}}};

        \draw[->] (gn_1) to [out=0, in=180] (lin1.north);
        \draw[->] (gn_2) to [out=0, in=180] (lin1.north);
        \draw[->] (mn_1) to [out=0, in=180] (lin1.north);
        \draw[->] (mn_2) to [out=0, in=180] (lin1.north);
        \draw[->] (sn_1) to [out=0, in=180] (lin1.north);
        \draw[->] (sn_2) to [out=0, in=180] (lin1.north);
        \draw[->] (ld) to [out=0, in=180] (lin1.north);
        \draw[->] (ldns) to [out=0, in=180] (lin1.north);
        \draw[->] (lds) to [out=0, in=180] (lin1.north);
        \draw[->] (sp_1) to [out=0, in=180] (lin1.north);
        \draw[->] (sp_2) to [out=0, in=180] (lin1.north);
        \draw[->] (reg) to [out=0, in=180] (lin1.north);
        
        \node (s1) at (1.8, 1.4) {\tiny{$s_1$}};
        \node (s2) at (1.8, 1.0) {\tiny{$s_2$}};
        \node (s3) at (1.8, 0.6) {\tiny{$s_3$}};
        \node (s4) at (1.8, 0.2) {\tiny{$s_4$}};
        \node (s5) at (1.8, -0.2) {\tiny{$s_5$}};
        \node (s6) at (1.8, -0.6) {\tiny{$s_6$}};
        \node (s7) at (1.8, -1.0) {\tiny{$s_7$}};
        \node (s8) at (1.8, -1.4) {\tiny{$s_8$}};

        \draw[->] (lin4.south) to [out=0, in=180] (s1.west);
        \draw[->] (lin4.south) to [out=0, in=180] (s2.west);
        \draw[->] (lin4.south) to [out=0, in=180] (s3.west);
        \draw[->] (lin4.south) to [out=0, in=180] (s4.west);
        \draw[->] (lin4.south) to [out=0, in=180] (s5.west);
        \draw[->] (lin4.south) to [out=0, in=180] (s6.west);
        \draw[->] (lin4.south) to [out=0, in=180] (s7.west);
        \draw[->] (lin4.south) to [out=0, in=180] (s8.west);

        \draw[-] (lin1) -- (relu2_1) -- (lin2) -- (relu2_2) -- (lin3) -- (relu2_3) -- (lin4);

        % Directions
        \node (d1) at (3.4, 1.4) {\tiny{$s_1 \cdot \iter{d}{t}_{1, >}$}};
        \node (d2) at (3.4, 1.0) {\tiny{$s_2 \cdot \iter{d}{t}_{1, 0}$}};
        \node (d3) at (3.4, 0.6) {\tiny{$s_3 \cdot \iter{d}{t}_{2, >}$}};
        \node (d4) at (3.4, 0.2) {\tiny{$s_4 \cdot \iter{d}{t}_{2, 0}$}};
        \node (d5) at (3.4, -0.2) {\tiny{$s_5 \cdot \iter{d}{t}_{3, >}$}};
        \node (d6) at (3.4, -0.6) {\tiny{$s_6 \cdot \iter{d}{t}_{3, 0}$}};
        \node (d7) at (3.4, -1.0) {\tiny{$s_7 \cdot \iter{d}{t}_{4, >}$}};
        \node (d8) at (3.4, -1.4) {\tiny{$s_8 \cdot \iter{d}{t}_{4, 0}$}};

        \draw[->] (s1) to (d1.west);
        \draw[->] (s2) to (d2.west);
        \draw[->] (s3) to (d3.west);
        \draw[->] (s4) to (d4.west);
        \draw[->] (s5) to (d5.west);
        \draw[->] (s6) to (d6.west);
        \draw[->] (s7) to (d7.west);
        \draw[->] (s8) to (d8.west);

        % Update Block
        \node[conv,rotate=90,minimum width=2.7cm] (conv1) at (5.25,0) {\tiny{\texttt{Conv2d(8,20,1,bias=F)}}};
        \node[relu,rotate=90,minimum width=2.7cm] (relu1) at (5.75,0) {\tiny{\texttt{ReLU}}};
        
        \node[conv,rotate=90,minimum width=2.7cm] (conv2) at (6.25,0) {\tiny{\texttt{Conv2d(20,20,1,bias=F)}}};
        \node[relu,rotate=90,minimum width=2.7cm] (relu2) at (6.75,0) {\tiny{\texttt{ReLU}}};

        \node[conv,rotate=90,minimum width=2.7cm] (conv3) at (7.25,0) {\tiny{\texttt{Conv2d(20,20,1,bias=F)}}};
        \node[relu,rotate=90,minimum width=2.7cm] (relu3) at (7.75,0) {\tiny{\texttt{ReLU}}};

        \node[conv,rotate=90,minimum width=2.7cm] (conv4) at (8.25,0) {\tiny{\texttt{Conv2d(20,2,1,bias=F)}}};

        \node (d_1) at (9.3, 0.5) {\tiny{$\iter{d}{t}_{out, 1}$}};
        \node (d_2) at (9.3, -0.5) {\tiny{$\iter{d}{t}_{out, 2}$}};
        
        \draw[->] (d1) to [out=0, in=180] (conv1.north);
        \draw[->] (d2) to [out=0, in=180] (conv1.north);
        \draw[->] (d3) to [out=0, in=180] (conv1.north);
        \draw[->] (d4) to [out=0, in=180] (conv1.north);
        \draw[->] (d5) to [out=0, in=180] (conv1.north);
        \draw[->] (d6) to [out=0, in=180] (conv1.north);
        \draw[->] (d7) to [out=0, in=180] (conv1.north);
        \draw[->] (d8) to [out=0, in=180] (conv1.north);

        \draw[-] (conv1) -- (relu1) -- (conv2) -- (relu2) -- (conv3) -- (relu3) -- (conv4);
        \draw[->] (conv4.south) to [out=0, in=180] (d_1.west);
        \draw[->] (conv4.south) to [out=0, in=180] (d_2.west);

        % Combine to new point
        \node (xk) at (10.6, 2.) {\tiny{$\iter{x}{t}$}};
        \node (new) at (10.6, 0) {\tiny{$\iter{x}{t+1}$}};
        \draw[->] (xk) to [out=270, in=90] (new.north);
        \draw[->] (d_1) to [out=0, in=180] (new.west);
        \draw[->] (d_2) to [out=0, in=180] (new.west);
        
    \end{tikzpicture}
    \caption{Algorithmic update for the LASSO problem: Based on the given twelve features, the first block computes eight weights, which are used to perform a weighting of the different directions, which are used in the second block. This second block then predicts two directions $d_{out, 1}, d_{out, 2}$, where $d_{out, 1}$ only acts on the non-zero entries, and $d_{out, 2}$ acts on the zero entries. These are used in the update $\iter{x}{t+1} := \mathrm{prox}_{\beta g}\left(\iter{x}{t} + \left(\iter{d}{t}_{out, 1, >} - \nabla \ell(\iter{x}{t}) + \norm{\iter{x}{t} - \iter{x}{t-1}}{} \cdot \iter{d}{t}_{out, 2, 0} \right) / L \right)$.}
\end{figure}
\noindent
Since in the LASSO problem the algorithm has to identify the support of the solution, that is, the coordinates which are non-zero, we also treat the zero and non-zero entries of $\iter{x}{t}$ (and derived quantities) separately. Here, we denote the non-zero entries by $\iter{x}{t}_{>}$ and the zero entries by $\iter{x}{t}_0$, and similarly for all other quantities. First, we compute weights $s_1, ..., s_8$ with a fully-connected block with ReLU-activation functions. The used features are $\iter{n}{t}_1 = \log(1 + \norm{\nabla \ell(\iter{x}{t}, \realPar)}{})$, $\iter{n}{t}_2 = \log(1 + \norm{\iter{x}{t} - \iter{x}{t-1}}{})$, $\iter{n}{t}_3 = \log(1 + \norm{\iter{p}{t}}{})$, where $\iter{p}{t} = \mathrm{prox}_{\beta g} \left( \iter{x}{t} - \beta \nabla \ell(\iter{x}{t}, \realPar) \right)$, $\Delta \iter{\ell}{t} := \ell(\iter{x}{t}, \realPar) - \ell(\iter{x}{t-1}, \realPar)$, $\Delta \iter{g}{t} := g(\iter{x}{t}) - g(\iter{x}{t-1})$, $\Delta \iter{h}{t} := h(\iter{x}{t}, \realPar) - h(\iter{x}{t-1}, \realPar)$, the scalar products $\iter{s}{t}_>$ and $\iter{s}{t}_0$ between the (normalized) gradient and (normalized) momentum, and the regularization parameter $\lambda$. Then, these weights are used to perform a reweighting of the given directions $\iter{d}{t}_1, ..., \iter{d}{t}_4$, given by the normalized gradient, the normalized momentum, the normalized residual $\iter{x}{t} - \iter{p}{t}$, and the coordinate-wise product between (normalized) gradient and (normalized) momentum. Afterwards, these reweighted directions get fed into a 1x1-convolutional block, which predicts the two directions $\iter{d}{t}_{out, 1}$ and $\iter{d}{t}_{out, 2}$, which are used to compute the final update with the proximal mapping, given by
$$
\iter{x}{t+1} := \mathrm{prox}_{\beta g}\left(\iter{x}{t} + \left(\iter{d}{t}_{out, 1, >} - \nabla \ell(\iter{x}{t}) + \norm{\iter{x}{t} - \iter{x}{t-1}}{} \cdot \iter{d}{t}_{out, 2, 0} \right) / L \right) \,.
$$

\newpage
\section{Architecture for Training the Neural Network} \label{Appendix:architecture_nn}

\definecolor{test_1}{RGB}{241, 241, 242} 
\definecolor{test_2}{RGB}{25, 149, 173} 
\definecolor{test_3}{RGB}{161, 214, 226} 
\definecolor{test_4}{RGB}{188, 186, 190} 

\begin{figure}[h!]
    \centering
    \begin{tikzpicture}
        \tikzset{conv/.style={black,draw=black,fill=test_1,rectangle,minimum height=0.5cm}}
        \tikzset{linear/.style={black,draw=black,fill=test_2,rectangle,minimum height=0.5cm}}
        \tikzset{relu/.style={black,draw=black,fill=test_3,rectangle,minimum height=0.25cm}}

        % Norms
        \node (n1) at (-4.05, 1.5) {\tiny{$\iter{n}{t}_1$}};
        \node (n2) at (-4.05, 0.9) {\tiny{$\iter{n}{t}_2$}};
        \node (l) at (-4.1, 0.3) {\tiny{$\Delta \iter{\ell}{t}$}};
        \node (mg) at (-4, -0.3) {\tiny{$\iter{g}{t}$}};
        \node (sp) at (-4.05, -0.9) {\tiny{$\iter{s}{t}$}};
        \node (t) at (-3.9, -1.5) {\tiny{$t$}};

        % Step-size block
        \node[linear,rotate=90,minimum width=2.75cm] (lin1) at (-2.5, 0) {\tiny{\texttt{Linear(6,30,bias=F)}}};
        \node[relu,rotate=90,minimum width=2.75cm] (relu2_1) at (-2.0, 0) {\tiny{\texttt{ReLU}}};

        \node[linear,rotate=90,minimum width=2.75cm] (lin2) at (-1.5, 0) {\tiny{\texttt{Linear(30,20,bias=F)}}};
        \node[relu,rotate=90,minimum width=2.75cm] (relu2_2) at (-1.0, 0) {\tiny{\texttt{ReLU}}};
        
        \node[linear,rotate=90,minimum width=2.75cm] (lin3) at (-0.5, 0) {\tiny{\texttt{Linear(20,10,bias=F)}}};
        \node[relu,rotate=90,minimum width=2.75cm] (relu2_3) at (0, 0) {\tiny{\texttt{ReLU}}};

        \node[linear,rotate=90,minimum width=2.75cm] (lin4) at (0.5, 0) {\tiny{\texttt{Linear(10,4,bias=F)}}};

        \draw[->] (n1) to [out=0, in=180] (lin1.north);
        \draw[->] (n2) to [out=0, in=180] (lin1.north);
        \draw[->] (l) to [out=0, in=180] (lin1.north);
        \draw[->] (mg) to [out=0, in=180] (lin1.north);
        \draw[->] (sp) to [out=0, in=180] (lin1.north);
        \draw[->] (t) to [out=0, in=180] (lin1.north);

        \node (s1) at (1.8, 1.25) {\tiny{$s_1$}};
        \node (s2) at (1.8, 0.325) {\tiny{$s_2$}};
        \node (s3) at (1.8, -0.325) {\tiny{$s_3$}};
        \node (s4) at (1.8, -1.25) {\tiny{$s_4$}};

        \draw[->] (lin4.south) to [out=0, in=180] (s1.west);
        \draw[->] (lin4.south) to [out=0, in=180] (s2.west);
        \draw[->] (lin4.south) to [out=0, in=180] (s3.west);
        \draw[->] (lin4.south) to [out=0, in=180] (s4.west);

        \draw[-] (lin1) -- (relu2_1) -- (lin2) -- (relu2_2) -- (lin3) -- (relu2_3) -- (lin4);

        % Directions
        \draw (1.9, 2.) circle [radius=0.18] node (g) {\tiny{$g$}};
        \draw (1.9, -2.) circle [radius=0.18] node (m) {\tiny{$m$}};
        \node (d1) at (3.4, 1.25) {\tiny{$s_1 \cdot g \odot \iter{d}{t}_1$}};
        \node (d2) at (3.7, 0.325) {\tiny{$s_2 \cdot \iter{d}{t}_1$}};
        \node (d3) at (3.7, -0.325) {\tiny{$s_3 \cdot \iter{d}{t}_2$}};
        \node (d4) at (3.4, -1.25) {\tiny{$s_4 \cdot m \odot \iter{d}{t}_2$}};

        \draw[->] (g) to [out=0,in=90] (d1.north);
        \draw[->] (s1) to (d1.west);
        \draw[->] (s2) to (d2.west);
        \draw[->] (s3) to (d3.west);
        \draw[->] (s4) to (d4.west);
        \draw[->] (m) to [out=0, in=270] (d4.south);

        % Update Block
        \node[conv,rotate=90,minimum width=2.7cm] (conv1) at (5.25,0) {\tiny{\texttt{Conv2d(4,20,1,bias=F)}}};
        \node[relu,rotate=90,minimum width=2.7cm] (relu1) at (5.75,0) {\tiny{\texttt{ReLU}}};
        
        \node[conv,rotate=90,minimum width=2.7cm] (conv2) at (6.25,0) {\tiny{\texttt{Conv2d(20,20,1,bias=F)}}};
        \node[relu,rotate=90,minimum width=2.7cm] (relu2) at (6.75,0) {\tiny{\texttt{ReLU}}};

        \node[conv,rotate=90,minimum width=2.7cm] (conv3) at (7.25,0) {\tiny{\texttt{Conv2d(20,20,1,bias=F)}}};
        \node[relu,rotate=90,minimum width=2.7cm] (relu3) at (7.75,0) {\tiny{\texttt{ReLU}}};

        \node[conv,rotate=90,minimum width=2.7cm] (conv4) at (8.25,0) {\tiny{\texttt{Conv2d(20,1,1,bias=F)}}};

        \node (d) at (9.2, 0) {\tiny{$\iter{d}{t}_{out}$}};

        \node (xk) at (10.4, 2.) {\tiny{$\iter{x}{t}$}};
        
        \draw[->] (d1) to [out=0, in=180] (conv1.north);
        \draw[->] (d2) to [out=0, in=180] (conv1.north);
        \draw[->] (d3) to [out=0, in=180] (conv1.north);
        \draw[->] (d4) to [out=0, in=180] (conv1.north);

        \draw[-] (conv1) -- (relu1) -- (conv2) -- (relu2) -- (conv3) -- (relu3) -- (conv4);
        \draw[->] (conv4) -- (d);

        % Combine to new point
        \node (new) at (10.4, 0) {\tiny{$\iter{x}{t+1}$}};
        \draw[->] (xk) to [out=270, in=90] (new.north);
        \draw[->] (d) -- (new);
        
    \end{tikzpicture}
    \caption{Algorithmic update for training the neural network: Based on the given six features, the first block computes four weights $s_1, ..., s_4$, which are used to perform a weighting of the different directions $g \odot \iter{d}{t}_1$, $\iter{d}{t}_1$, $\iter{d}{t}_2$, $m \odot \iter{d}{t}_2$, which are used in the second block. This second block consists of a 1x1-convolutional blocks, which compute an update direction $\iter{d}{t}_{out}$. Then, we update $\iter{x}{t+1} := \iter{x}{t} + \iter{d}{t}_{out}$.}
\end{figure}
\noindent
To compute the weights $s_1, ..., s_4$ with the first block, we use the features $\iter{n_1}{t} = \log(1 + \norm{\nabla \ell(\iter{x}{t}, \realPar)}{})$, $\iter{n_2}{t} = \log(1 + \norm{\iter{x}{t} - \iter{x}{t-1}}{})$, $\Delta \iter{\ell}{t} := \ell(\iter{x}{t}, \realPar) - \ell(\iter{x}{t-1}, \realPar)$, $\iter{g}{t} := \max_{i=1, ..., n}  \abs{\nabla \ell(\iter{x}{t}, \realPar)_i}$, the scalar product $\iter{s}{t}$ between the (normalized) gradient and the (normalized) momentum, and the iteration counter $t$. Then, these weights are used to perform a weighting of the directions $\iter{d}{t}_1, \iter{d}{t}_2, g \odot \iter{d}{t}_1$ and $m \odot \iter{d}{t}_2$, where $g$ and $m$ are additional learned vectors of size $n$, which we use as diagonal preconditioning. Here, $\iter{d}{t}_1$ is the normalized gradient and $\iter{d}{t}_2$ is the normalized momentum term. These directions get fed into a $1 \times 1$-convolutional block, which predicts the direction $\iter{d}{t}_{out}$ that is used to update the iterate as $\iter{x}{t+1} = \iter{x}{t} + \iter{d}{t}_{out}$.

\newpage
\section{Architecture for Stochastic Empirical Risk Minimization} \label{Appendix:architecture_stochastic_emp_risk}

\definecolor{test_1}{RGB}{241, 241, 242} 
\definecolor{test_2}{RGB}{25, 149, 173} 
\definecolor{test_3}{RGB}{161, 214, 226} 
\definecolor{test_4}{RGB}{188, 186, 190} 

\begin{figure}[h!]
    \centering
    \begin{tikzpicture}
        \tikzset{conv/.style={black,draw=black,fill=test_1,rectangle,minimum height=0.5cm}}
        \tikzset{linear/.style={black,draw=black,fill=test_2,rectangle,minimum height=0.5cm}}
        \tikzset{relu/.style={black,draw=black,fill=test_3,rectangle,minimum height=0.25cm}}

        % Norms
        \node (n1) at (-4, 0.75) {\tiny{$\iter{n}{t}_1$}};
        \node (n2) at (-4, 0.25) {\tiny{$\iter{n}{t}_2$}};
        \node (t) at (-3.8, -0.25) {\tiny{$t$}};
        \node (l) at (-4.1, -0.75) {\tiny{$\iter{\ell}{t}_{\mathrm{stoch}}$}};

        % Step-size block
        \node[linear,rotate=90,minimum width=2.75cm] (lin1) at (-2.5, 0) {\tiny{\texttt{Linear(3,15,bias=F)}}};
        \node[relu,rotate=90,minimum width=2.75cm] (relu2_1) at (-2.0, 0) {\tiny{\texttt{ReLU}}};

        \node[linear,rotate=90,minimum width=2.75cm] (lin2) at (-1.5, 0) {\tiny{\texttt{Linear(15,15,bias=F)}}};
        \node[relu,rotate=90,minimum width=2.75cm] (relu2_2) at (-1.0, 0) {\tiny{\texttt{ReLU}}};

        \node[linear,rotate=90,minimum width=2.75cm] (lin4) at (-0.5, 0) {\tiny{\texttt{Linear(15,4,bias=F)}}};

        \draw[->] (n1) to [out=0, in=180] (lin1.north);
        \draw[->] (n2) to [out=0, in=180] (lin1.north);
        \draw[->] (l) to [out=0, in=180] (lin1.north);
        \draw[->] (t) to [out=0, in=180] (lin1.north);

        \node (s1) at (1.3, 1.) {\tiny{$s_1$}};
        \node (s2) at (1.3, 0.) {\tiny{$s_2$}};
        \node (s3) at (1.3, -1.) {\tiny{$s_3$}};
        \node (s4) at (1.3, -3.0) {\tiny{$s_4$}};

        \draw[->] (lin4.south) to [out=0, in=180] (s1.west);
        \draw[->] (lin4.south) to [out=0, in=180] (s2.west);
        \draw[->] (lin4.south) to [out=0, in=180] (s3.west);
        \draw[->] (lin4.south) to [out=0, in=180] (s4.west);

        \draw[-] (lin1) -- (relu2_1) -- (lin2) -- (relu2_2) -- (lin3) -- (relu2_3) -- (lin4);

        % Directions
        \node (d1) at (3.0, 1.) {\tiny{$s_1 \cdot \iter{d}{t}_1$}};
        \node (d2) at (2.7, 0.) {\tiny{$s_2 \cdot \iter{d}{t}_1 \odot \iter{d}{t}_2$}};
        \node (d3) at (3.0, -1.) {\tiny{$s_3 \cdot \iter{d}{t}_2$}};

        \node (m) at (3.0, 2.) {\tiny{$\iter{d}{t}_1$}};
        \node (v) at (3.0, -2.) {\tiny{$\iter{d}{t}_2$}};

        \draw[->] (m.south) to [out=270, in=90] (d1.north);
        \draw[->] (v.north) to [out=90, in=270] (d3.south);

        \draw[->] (s1) to (d1.west);
        \draw[->] (s2) to (d2.west);
        \draw[->] (s3) to (d3.west);

        % Update Block
        \node[conv,rotate=90,minimum width=2.7cm] (conv1) at (5.,0) {\tiny{\texttt{Conv2d(3,15,1,bias=F)}}};
        \node[relu,rotate=90,minimum width=2.7cm] (relu1) at (5.5,0) {\tiny{\texttt{ReLU}}};
        
        \node[conv,rotate=90,minimum width=2.7cm] (conv2) at (6.,0) {\tiny{\texttt{Conv2d(15,15,1,bias=F)}}};
        \node[relu,rotate=90,minimum width=2.7cm] (relu2) at (6.5,0) {\tiny{\texttt{ReLU}}};

        \node[conv,rotate=90,minimum width=2.7cm] (conv3) at (7.,0) {\tiny{\texttt{Conv2d(15,2,1,bias=F)}}};

        \node (p1) at (8.5, 0.5) {\tiny{$\iter{p}{t}_{1}$}};
        \node (p2) at (8.5, -0.5) {\tiny{$\iter{p}{t}_{2}$}};
        
        \draw[->] (d1) to [out=0, in=180] (conv1.north);
        \draw[->] (d2) to [out=0, in=180] (conv1.north);
        \draw[->] (d3) to [out=0, in=180] (conv1.north);

        \draw[-] (conv1) -- (relu1) -- (conv2) -- (relu2) -- (conv3);
        \draw[->] (conv3.south) to [out=0, in=180] (p1.west);
        \draw[->] (conv3.south) to [out=0, in=180] (p2.west);

        % Combine to new point
        \node (new) at (10.5, 0) {\tiny{$\iter{x}{t+1}$}};
        
        \node (xk) at (10.5, 5.) {\tiny{$\iter{x}{t}$}};
        \draw (9.8, 5.) circle [radius=0.2] node (b1) {\tiny{$\beta_1$}};
        \draw (9.3, 5.) circle [radius=0.2] node (b2) {\tiny{$\beta_2$}};
        \draw (8.8, 5.) circle [radius=0.2] node (eps) {\tiny{$\varepsilon$}};
        \draw (8.3, 5.) circle [radius=0.2] node (kappa) {\tiny{$\kappa$}};
        
        \draw[->] (xk) to [out=270, in=90] (new.north);
        \draw[->] (b1) to [out=270, in=90] (new.north);
        \draw[->] (b2) to [out=270, in=90] (new.north);
        \draw[->] (eps) to [out=270, in=90] (new.north);
        \draw[->] (kappa) to [out=270, in=90] (new.north);
        \draw[->] (p1.east) to [out=0, in=180] (new.west);
        \draw[->] (p2.east) to [out=0, in=180] (new.west);
        \draw[->] (s4.east) to [out=0, in=270] (new.south);
        \draw[->] (v.east) to [out=0, in=270] (new.south);
        \draw[->] (m.east) to [out=0, in=90] (new.north);
        
    \end{tikzpicture}
    \caption{Algorithmic update for stochastic empirical risk minimization: Based on the given four features, the first block computes four weights $s_1, ..., s_4$, where $s_1, ..., s_3$ are used to perform a weighting of the different directions $\iter{d}{t}_1$, $\iter{d}{t}_2$, and $\iter{d}{t}_1 \odot \iter{d}{t}_2$, which get fed into the second block, and $s_4$ is used afterwards as a step-size. The second block consists of a 1x1-convolutional layers, that is, they act coordinate-wise, and computes two vectors $\iter{p}{t}_1$, $\iter{p}{t}_2$, which are used as diagonal \say{preconditioners} for the update of Adam.}
\end{figure}
\noindent
First, we compute the same features as Adam, that is, $\iter{m}{t} = \beta_1 \iter{m}{t-1} + (1-\beta_1) \hat{\nabla} \ell(\iter{x}{t}, \realPar)$, $\iter{v}{t} = \beta_2 \iter{v}{t-1} + (1-\beta_2) \hat{\nabla} \ell(\iter{x}{t}, \realPar) \odot \hat{\nabla} \ell(\iter{x}{t}, \realPar)$, where $\hat{\nabla} \ell(\iter{x}{t}, \realPar)$ denotes the stochastic gradient. Then, as for Adam, we set $\iter{\hat{m}}{t} = \frac{\iter{m}{t}}{1 - \beta_1^t}$ and $\iter{\hat{v}}{t} = \frac{\iter{v}{t}}{1 - \beta_2^t}$, and split $\iter{\hat{m}}{t}$ and $\iter{\hat{v}}{t}$ into the corresponding (logarithmically transformed) norms $\iter{n}{t}_1$, $\iter{n}{t}_2$ and unit-vectors $\iter{d}{t}_1$, and $\iter{d}{t}_2$, respectively. The norms, together with the current (stochastic) loss and the iteration counter $t$, get fed into a fully-connected block to compute four weights $s_1, ..., s_4$. Then, $s_4$ is used as a step-size in the final update, while $s_1, ..., s_3$ get used to weigh the vectors $\iter{d}{t}_1$, $\iter{d}{t}_2$, and $\iter{d}{t}_1 \odot \iter{d}{t}_2$ before they get fed into the $1\times1$-convolutional block, which outputs vectors $p_1$ and $p_2$. These, in turn, are used as a kind of diagonal preconditioner for the update of Adam, that is, the final output is given by the formula 
$\iter{x}{t} = \iter{x}{t-1} - (s_4 \cdot \kappa \cdot d_1 \odot \iter{\hat{m}}{t}) / (0.001 \cdot \abs{d_2} \odot \sqrt{\iter{\hat{v}}{t}} + \varepsilon)$. Here, the constant 0.001 is just there to stabilize training in the beginning, and for the other constants we use the default values in PyTorch, that is, we set $\kappa = 0.001$, $\beta_1 = 0.9$, $\beta_2 = 0.999$, and $\varepsilon = 1\cdot 10^{-8}$.

\end{document}